\documentclass[5p,twocolumn]{elsarticle}

\journal{Journal of \LaTeX\ Templates}

\usepackage[normalem]{ulem} 
\usepackage[utf8]{inputenc}
\usepackage{hyperref}
\usepackage{graphicx}
\usepackage{natbib}
\usepackage{lineno}
\usepackage{amsmath,amsthm,amssymb}
\usepackage{lipsum}
\usepackage{array}

\modulolinenumbers[5]

\biboptions{authoryear}

\protected\def\picometer{\ifmmode \,\operatorname{pm}\else $\operatorname{pm}$\fi}
\protected\def\nm{\ifmmode \,\operatorname{nm}\else $\operatorname{nm}$\fi}
\protected\def\micron{\ifmmode \,\operatorname{\mu m}\else $\operatorname{\mu m}$\fi}
\protected\def\mm{\ifmmode \,\operatorname{mm}\else $\operatorname{mm}$\fi}
\protected\def\meter{\ifmmode \,\operatorname{m}\else $\operatorname{m}$\fi}
\protected\def\km{\ifmmode \,\operatorname{km}\else $\operatorname{km}$\fi}
\protected\def\au{\ifmmode \,\operatorname{AU}\else $\operatorname{AU}$\fi}
\protected\def\pc{\ifmmode \,\operatorname{pc}\else $\operatorname{pc}$\fi}
\protected\def\kpc{\ifmmode \,\operatorname{kpc}\else $\operatorname{kpc}$\fi}
\protected\def\Mpc{\ifmmode \,\operatorname{Mpc}\else $\operatorname{Mpc}$\fi}
\protected\def\rsun{\ifmmode \,\operatorname{R_\odot}\else $\operatorname{R_\odot}$\fi}
\protected\def\Rsun{\ifmmode \,\operatorname{R_\odot}\else $\operatorname{R_\odot}$\fi}

\protected\def\second{\ifmmode \,\operatorname{sec}\else $\operatorname{sec}$\fi}
\protected\def\yr{\ifmmode \,\operatorname{yr}\else $\operatorname{yr}$\fi}
\protected\def\Gyr{\ifmmode \,\operatorname{Gyr}\else $\operatorname{Gyr}$\fi}

\protected\def\eV{\ifmmode \,\operatorname{eV}\else $\operatorname{eV}$\fi}
\protected\def\keV{\ifmmode \,\operatorname{keV}\else $\operatorname{keV}$\fi}
\protected\def\MeV{\ifmmode \,\operatorname{MeV}\else $\operatorname{MeV}$\fi}
\protected\def\GeV{\ifmmode \,\operatorname{GeV}\else $\operatorname{GeV}$\fi}
\protected\def\TeV{\ifmmode \,\operatorname{TeV}\else $\operatorname{TeV}$\fi}

\protected\def\Lsun{\ifmmode \,\operatorname{L_\odot}\else $\operatorname{L_\odot}$\fi}
\protected\def\lsun{\ifmmode \,\operatorname{L_\odot}\else $\operatorname{L_\odot}$\fi}
\protected\def\Watt{\ifmmode \,\operatorname{W}\else $\operatorname{W}$\fi}
\protected\def\nW{\ifmmode \,\operatorname{nW}\else $\operatorname{nW}$\fi}

\protected\def\kJy{\ifmmode \,\operatorname{kJy}\else $\operatorname{kJy}$\fi}
\protected\def\Jy{\ifmmode \,\operatorname{Jy}\else $\operatorname{Jy}$\fi}
\protected\def\mJy{\ifmmode \,\operatorname{mJy}\else $\operatorname{mJy}$\fi}
\protected\def\microJy{\ifmmode \,\operatorname{\mu Jy}\else $\operatorname{\mu Jy}$\fi}
\protected\def\nJy{\ifmmode \,\operatorname{nJy}\else $\operatorname{nJy}$\fi}

\protected\def\Mag{\ifmmode \,\operatorname{mag}\else $\operatorname{mag}$\fi}

\protected\def\deg{\ifmmode ^{\circ}\else $^{\circ}$\fi}
\protected\def\arcsec{\ifmmode ^{\prime\prime}\else $^{\prime\prime}$\fi}

\protected\def\arcsecT{\ifmmode \,\operatorname{arcsec}\else $\operatorname{arcsec}$\fi}
\protected\def\arcmin{\ifmmode ^{\prime}\else $^{\prime}$\fi}

\protected\def\arcminT{\ifmmode \,\operatorname{arcmin}\else $\operatorname{arcmin}$\fi}
\protected\def\sr{\ifmmode \,\operatorname{sr}\else $\operatorname{sr}$\fi}

\newcommand{\code}[1]{\texttt{#1}}

\protected\def\d{\ifmmode \operatorname{d}\else
    $\operatorname{d}$\fi}

\makeatletter
\def\blfootnote{\xdef\@thefnmark{}\@footnotetext}
\makeatother

\def\apjref#1;#2;#3;#4 {\par\pp#1, {#2}, #3, #4 \par}

\begin{document}

\begin{frontmatter}
\title{An Exploration of How Training Set Composition Bias in Machine Learning Affects Identifying Rare Objects}

\author[NAOCaddress]{\href{https://orcid.org/0000-0002-4528-7637}{S.~E.~Lake\corref{corauth}}}
\cortext[corauth]{S.~E.~Lake}
\ead{lake@nao.cas.cn}

\author[NAOCaddress]{\href{https://orcid.org/0000-0002-9390-9672}{C.-W.~Tsai}}
\ead{cwtsai@nao.cas.cn}

\address[NAOCaddress]{National Astronomical Observatories, Chinese Academy of Sciences, Beijing, 100012, People's Republic of China}

\begin{abstract}
When training a machine learning classifier on data where one of the classes is intrinsically rare, the classifier will often assign too few sources to the rare class. To address this, it is common to up-weight the examples of the rare class to ensure it is nott ignored. 
It is also a frequent practice to train on restricted data where the balance of source types is closer to equal for the same reason. 
Here we show that these practices can bias the model toward over-assigning sources to the rare class.
We also explore how to detect when training data bias has had a statistically significant impact on the trained model's predictions, and how to reduce the bias's impact.
While the magnitude of the impact of the techniques developed here will vary with the details of the application, for most cases it should be modest.
They are, however, universally applicable to every time a machine learning classification model is used, making them analogous to Bessel's correction to the sample variance.
\end{abstract}

\begin{keyword}
methods: data analysis, methods: statistical, Machine learning, Probability and statistics
\end{keyword}

\end{frontmatter}


\section{Introduction}\label{sec:intro}
\blfootnote{Abbreviations: PCP = prediction consistent prior.}
Machine learning is taking a more prominent role in astronomy research as the size of observational and simulated data sets swell beyond a human's capability of digesting them. 
This is due to the rapid expansion of computing resources and sensor technology in the last four decades that has driven equally rapid expansions in the quantity of data to analyze.
Astronomy, in particular, has seen a proliferation of large scale imaging and spectroscopic surveys that have billions of sources in them---surveys like: the Sloan Digital Sky Survey \citep[SDSS,][]{York:2000}, the 2-Micron All Sky Survey \citep[2MASS,][]{Skrutskie:2006}, the \textit{Wide-field Infrared Survey Explorer} \citep[\textit{WISE},][]{Wright:2010}, the \textit{Gaia} satellite's survey \citep{Gaia:2016}, the Panoramic Survey Telescope and Rapid Response System (Pan-STARRS) surveys \citep{PanStarrsPS1}, the Dark Energy Spectroscopic Instrument (DESI) surveys \citep{Dey:2019}, the UKIRT Infrared Deep Sky Surveys \citep[UKIDSS,][]{Lawrence:2007}, and the \textit{Galaxy Evolution Explorer} (GALEX) surveys \citep{Martin:2005}.
The challenge in the future is only expected to grow as surveys like the Legacy Survey of Space and Time \citep[LSST,][]{Ivezic:2019} and Square Kilometer Array \citep[SKA,][]{Dewdney:2009} come online.

One of the main justifications for producing these massive databases is that they are needed to find enough examples of rare sources to study them.
For example, the task of identifying the extremely luminous and dusty active galactic nuclei (AGNs), named hot dust obscured galaxies (Hot DOGs), in the \textit{WISE} sky survey \citep[as described in Section~4.2 of][]{Wright:2010}, which contains $^3/_4$ of billion stationary sources in its AllWISE release catalog \citep{Cutri:2013}, had many technical challenges and required intensive astronomy expertise, experience, and labor to overcome \citep[][for example]{Eisenhardt:2012}.
A necessary first step in that process, though, is to classify the sources so that we can prioritize which sources might be interesting, and which are examples of already known sources.
Because these sources are rare it is usually easier to use a supervised machine learning algorithm, one that is tuned using sources with known classifications, than it is to use an unsupervised one.
The reason should be obvious: subgroups of the common known source types are likely to outnumber the rare new ones, meaning a naive unsupervised machine learning algorithm could need a lot of complexity before it actually finds the rare class.

Supervised learning also has drawbacks when used for finding rare objects.
In general terms, a classifier's function is to draw decision boundaries between clumps of partially overlapping classes of data. 
A machine learning algorithm is both a method of describing such boundaries using parameters (a particular instance of such is a ``model" or ``classifier") and a procedure for tuning those parameters to optimize the shape of the boundaries by minimizing a function called the ``loss."
When the model has to describe more than one boundary, the model's parameters have to be divided among them.
If one of the classes is much rarer than the others then the loss of the model with respect to the data can have a larger change from assigning parameters to refining the decision boundaries between the numerous classes than to the boundaries around the rare ones.
Perhaps more importantly, though, is when the classifier assigns none of the feature space to the rare class, or even just too few examples \citep{Kubat:1998}.
These issues are part of why it is described as `imbalanced' if the number of data points of each type is not nearly equal in the training data set.
In those situations one commonly used strategy to ensure that the model describes a boundary around the rare classes is to up-weight the rare class(es) of data during training \citep[][is a review that covers other techniques]{He:2009}.
Most commonly, the weights are chosen to simulate training data with equal contributions from each class, as was done in \citet{Clarke:2020} and \citet{Cheng:2021a}.
The mechanism through which weighting data operates is that it simulates cloning the weighted data, even if the number of clones is not an integer.
If the machine learning algorithm correctly handles data with real clones in it, then weighting will move the decision boundaries between the classes in the same way: expanding the regions assigned to the rare ones to reflect their artificially increased abundance.

Anomaly detecting algorithms are another possible technique for finding rare objects. 
Their drawback is that they find everything unusual in a data set, when they work well. 
Most often, anomalies in data sets have mundane explanations \citep[e.g. optical ghosts, latents in the detector, cosmic ray hits, or a signal from a microwave someone opened at the wrong time, as in][]{Petroff:2015}.
So, while anomaly hunting is an essential part of looking for new phenomena, it will likely be less fruitful than the sort of more targeted search that supervised machine learning can produce when several examples of the target are already known.
Thus anomaly detection algorithms are complementary to other techniques.

Semi-supervised methods are more challenging to address, because having labels for some of the data allows for a much larger number of possible algorithms than unsupervised cluster finding algorithms.
Indeed, one of the results in this work is a demonstration that applying a model that calculates probabilities to an otherwise unlabeled data set can lead to a refinement in the model that improves its accuracy, which is a type of semi-supervised method.
Based on this example, we can say that if the method allows for soft classification in the form of probabilities, then the techniques described in this work are applicable.

The purpose of this work is to explore how weighting training data moves the decision boundaries by examining the question through the lens of measuring empirical classification probabilities.
We, therefore, focus on how weighting affects the structure of the data as a whole and, thereby, the predictions from the model.
Since weighting changes the effective balance of the data, altering the probabilities the model produces to not reflect the actual frequency of each class, we describe this change as a ``bias" throughout this paper.
Introducing statistical bias is bad practice, even if the goal of introducing the bias is to move the decision boundaries.
First, this work shows that moving the decision boundaries can be achieved by applying the desired weights to model's output probabilities, with higher performance in the cases tested.
Second, unbiased classification probabilities can be used to estimate the demographics of a population more accurately than the classification individual example.
This work shows that such assessments can be used to assess a model's performance on a target population, even in the absence of known true labels.
If weighting must be done during model training for other reasons, we also demonstrate here a simple method to compensate for this bias and that it works with minimal performance loss as long as the initial up-weighting is not too extreme.
We also show that this method is effective at compensating for bias that is present in the training data, as long as the bias meets certain conditions.

A similar method has been studied in the context of reducing biases induced by imbalanced data sets during ordinary logistic regression by \citet{King:2001} and \citet{Maalouf:2018}.
What sets their work apart from this one is that their chief concern was compensating for statistical bias in probabilities calculated from models fit using ordinary logistic regression, not the performance of a machine learning classifier working with imbalanced data.
Their approach is to use either weighted data fitting or ``prior correction" to remove the bias induced by discarding data to equalize the data sets.
That is, they discard common examples to balance the data sets, reducing the statistical bias expected from maximum likelihood logistic regression, and then compensate for the greater bias this induces.
While the basics of this process agree with the work presented here, the conclusion \citet{Maalouf:2018} on the relative merits of weighted fitting versus prior correction are opposite to what we observe in this work.

\citet{Bailer-Jones:2019} noted that when using Gaussian mixture models\footnote{For details, see section 6.3.1 of \citet{Ivezic:2019book}.} to construct a classifier that it implicitly assumes a flat Bayesian prior for the classification decision, and so used an estimate of the prior to adjust the classifier's probabilities.
They also made a point of noting how this prior effects the performance metrics of the classifier, specifically its purity and completeness.
\citet{Delchambre:2022} and \citet{GaiaExgal:2022} would go on to apply this technique in Gaia's third data release \citep{Gaia:2022}.
This is similar to the deweighting method described in this work, but only for the special case where the data were weighted to assume equal composition to the training set. 
Because deweighting, effectively, includes removal of the old prior, it applies to a broader set of situations.

In this work, we demonstrate the biases in supervised learning that result from weighting training data or training data with a biased mix of classes.
In Section~\ref{sec:theory} we present analytic arguments for why we would expect most machine learning algorithm classifiers to be subject to this weighting bias and describe the machine learning algorithm used to test these statements empirically.
In Section~\ref{sec:data} we describe the data sets used to empirically test the effectiveness of our compensation methods with both simulated and real observed data.
The purpose of using simulated data is that we can exactly calculate the true class probabilities that the classifier being tuned to approximate during training, and real data is used to ensure that the conclusions herein are robust to the harder to simulate complications that come with real data.
While the real data tasks we are using for demonstration are simple object classification based on photometry (for example, star--galaxy--AGN), they are general enough that they should prove universally applicable.
The results of applying a multi-layer perceptron machine learning algorithm to data are presented in Section~\ref{sec:res}.
In Section~\ref{sec:disc} we discuss the implications of this proposed deweighting scheme presented in this work.
Finally, this paper concludes with Section~\ref{sec:conc}.




\section{Analytic Derivation}\label{sec:theory}
This section begins with an exploration of how training set composition, and therefore weighting, affects the models produced by machine learning algorithms in Subsection~\ref{sec:theory:prior}.
Second, Subsection~\ref{sec:theory:overshoot} describes how the machine learning algorithm model's probabilities can be used to predict standard performance metrics.
The comparison between predicted and observed performance gives a way to measure how accurate the model's probabilities are, even if the true probabilities are unknown.
Finally, Subsection~\ref{sec:theory:reweight} defines the deweighting algorithm used to compensate for weighting bias. 
It also defines one way to compute the necessary deweighting factors when an unbiased training set composition is unknown.

\subsection{Data, Weights, and Prior Probability}\label{sec:theory:prior}
\begin{figure*}[htb]
	\begin{center}
	\includegraphics[width=\textwidth]{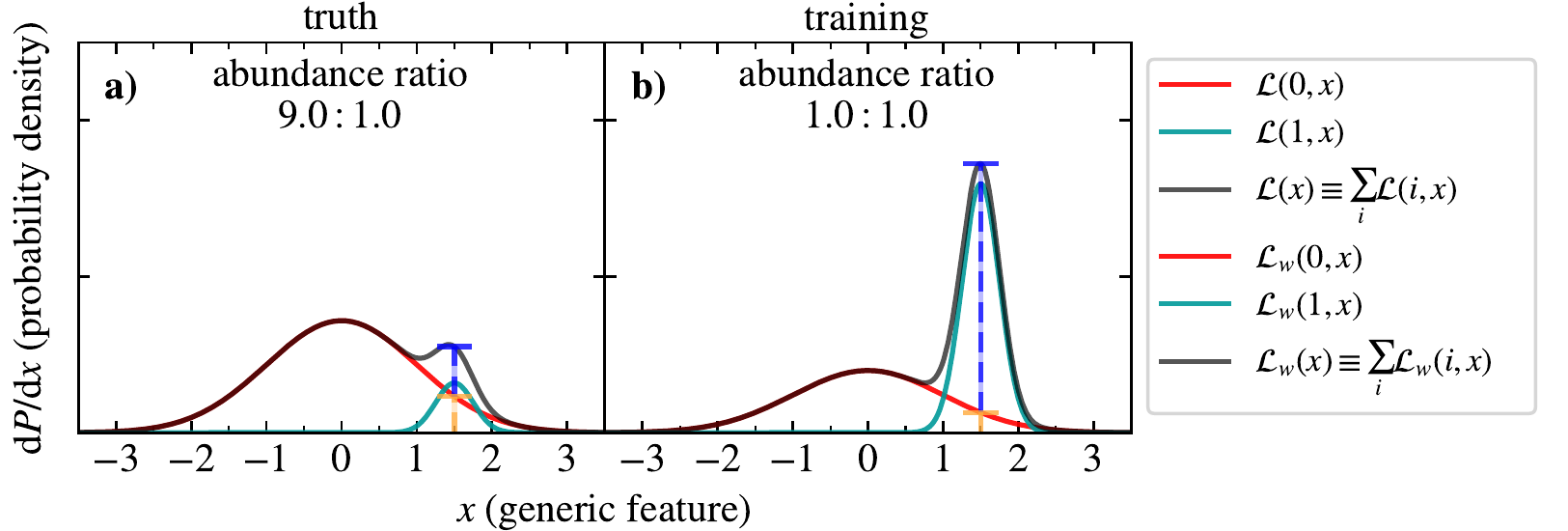}
	\end{center}
	
	\caption{Classification illustration using Gaussians}
	{The classification process for sources drawn from two Gaussians in illustrated form, including the way weighting the Gaussians to balance them shifts the probabilities (panels (\textbf{a}) and (\textbf{b}) for unweighted and weighted, respectively). 
	The orange and blue striped vertical lines illustrate how to calculate the probability that a source at $x=1.5$ comes from class 0 or 1.
	The probability that the source comes from class 0 is just the fraction of the total bar that is orange, and class 1 is the fraction that is blue.
	As a function of $x$ the probability that a source at $x$ comes from class $i$ is denoted by $P(i|x)$.
	Notice how the crossover point of the red and cyan lines, the decision boundary of a Bayes classifier, is shifted by weighting. 
	For this illustration, class 0 is from a Gaussian with mean 0 standard deviation 1, and contributes 90\% of the sources, while class 1 has mean 1.5 and standard deviation 0.25.
	}
	\label{fig:2gauss}
\end{figure*}

The fundamental mathematics behind classification are easiest to visualize using a 1-dimensional illustration of the how probabilities are calculated from data densities.
Figure~\ref{fig:2gauss} shows how the classification process works for data drawn from one of two overlapping Gaussian distributions.
In its panel~(\textbf{a}) the true densities of each type of object are plotted, labeled by $\mathcal{L}(i,x)$ (read as: ``the likelihood that a source will have feature $x$ and be from class $i$").
This represents what a training set with infinite data points would look like.
The figure's panel~(\textbf{b}) shows what weighting the data to balance the training set would do, with the subscript $w$ added to the likelihoods to emphasize this.

Note that a supervised classifier machine learning algorithm is taking the density of observed data, implicitly smoothing it, and using the smoothed data to calculate $P(i|x)$.
Even when only one label is assigned to each source instead of probabilities, that can be viewed as approximating the probability that the source belongs to the assigned class as 1, and the rest as 0.
Algebraically, if the sources for class $i$ have a density given by $\mathcal{L}(i,x)$ then the probabilities are given by
\begin{align}
	P(i|x) & = \frac{\mathcal{L}(i,x)}{\sum_j \mathcal{L}(j,x)}, \label{eqn:baseprob}
\end{align}
which is simply the number of sources class $i$ contributes at $x$ divided by the total number of sources there.

An example of $\mathcal{L}(i,x)$ can be found in color--magnitude diagrams (CMD) for galaxies \citep[for example:][]{Strateva:2001,Blanton:2003,Lake:2016}.
The CMDs usually contain two distinct groups of sources, one called the `red sequence' and the other the `blue cloud' (first observed as distinct in \citet{Strateva:2001}).
In this case, $\mathcal{L}(\text{red sequence},x)$ would be the CMD of just red sequence galaxies that have been spectroscopically identified before plotting them, and $x$ stands in for both the color and absolute magnitude of the galaxies.

$\mathcal{L}(i,x)$ can be split into its normalization $P(i)$ and its probability density function (PDF) $\mathcal{L}(x|i)$.
Doing so recasts Equation~\ref{eqn:baseprob} as
\begin{align}
	P(i|x) & = \frac{P(i)\, \mathcal{L}(x|i)}{\sum_j P(i)\, \mathcal{L}(x|j)}, \label{eqn:probBayes}
\end{align}
which gives $P(i|x)$ the same form as the Bayesian posterior probability, with $\mathcal{L}(x|i)$ the likelihood of the data given the model, and $P(i)$ is the prior probability of the model.
When there are $N$ data points drawn from $\mathcal{L}(i,x)$ without bias, then an unbiased maximum likelihood estimate of the prior probability is
\begin{align}
	P(i) & = \frac{n_i}{N}, \label{eqn:priorest}
\end{align}
where $n_i$ is the number of sources in class $i$ and $N$ is the number of sources in the data set.
Because machine learning algorithms are fit to real data sets of finite size, Equation~\ref{eqn:priorest} also serves as the effective prior that the fit model uses to compute probabilities.

In machine learning training begins by defining a `loss function' $L(\{\text{data}\}|\theta)$, also called the `badness of fit.' 
The training process consists of minimizing the loss over some set of model parameters, symbolized by $\theta$.
Two examples of loss functions commonly used in astronomy are $\chi^2$ and the negative log-likelihood.
A property that most loss functions share is that they take the form of a sum of losses for each individual data point
\begin{equation}
	L(\{\text{data}\}|\theta) = \sum_{n=1}^N L(i_n,x_n|\theta), \label{eqn:baseloss}
\end{equation}
where $L(i_n,x_n|\theta)$ is the loss one would use to fit the model to a data set with a single point.
Further specifying the form of $L(i_n,x_n|\theta)$ is unnecessary for the argument that follows.

The process of weighting is then changing the sum in Equation~\ref{eqn:baseloss} to a weighted sum
\begin{equation}
	L_w(\{\text{data}\}|\theta) = \sum_{n=1}^N w_n L(i_n,x_n|\theta). \label{eqn:weightloss}
\end{equation}
This is similar to what happens when Equation~\ref{eqn:baseloss} is rearranged to be a sum over F families with $N_f$ points each that are identical clones of one another
\begin{equation}
	L_C(\{\text{data}\}|\theta) = \sum_{f=1}^F N_f L(i_f, x_f|\theta). \label{eqn:cloneloss}
\end{equation}

It is worth making explicit that machine learning models are extremely flexible.
For example it has been proved mathematically that neural networks can approximate any function arbitrarily well, with enough neurons, as long as they satisfy two conditions: they use a non-linear activation function and have at least one hidden layer \citep{Hornik:1989}.
This flexibility means that all of the information in the model that makes it specific to a data set is encoded in the loss function used to train it.
In comparing Equations~\ref{eqn:weightloss} and \ref{eqn:cloneloss} it can be seen that the $w_n$ play exactly the same role as the $N_f$.
In other words, weighting the data has an effect that is indistinguishable, to the machine learning algorithm, from when the data contains real clones in it.

Changing the effective number of each class of data using weights replaces $n_i$ with a sum over the weights applied to examples from class $i$ in the effective prior, which yields a weighted prior
\begin{align}
	P_w(i) & = \frac{\sum_{n=1}^N w_n \delta_{i,j_n}}{\sum_{n=1}^N w_n} \nonumber \\
	& = \frac{\overline{w}_i P(i)}{\sum_i \overline{w}_i P(i)},\label{eqn:PriorWeightDef}
\end{align}
where $\overline{w}_i$ is the mean weight used for examples from class $i$, that is $\overline{w}_i \equiv \left.\sum_{n=1}^N w_n \delta_{i,j_n}\right/ n_i$, and $\delta_{i,j}$ is the Kronecker delta.
We note that another reason to use caution when weighting data is because the modified effective number of sources, $N_w = \sum_n w_n$, can taint many techniques for estimating the uncertainty of the model's predictions or parameters.
Equation~\ref{eqn:PriorWeightDef} means that the function the machine learning algorithm is trying to approximate is also modified to a weighted form
\begin{align}
	P_w(i|x) & = \frac{P_w(i)\, \mathcal{L}(x|i)}{\sum_j P_w(j)\, \mathcal{L}(x|j)}, \label{eqn:PostWeightDef}
\end{align}
which is biased away from $P(i|x)$.
The impact of this bias depends on how much the PDFs $\mathcal{L}(x|j)$ overlap with each other: the greater the overlap the larger the impact.

The bias in Equation~\ref{eqn:PostWeightDef} comes about from changing the normalization of each of the densities.
It can be seen by inspection that any technique that fixes the weighting induced bias can also work on biases among classes that are not functions of the relevant features of the data $x$. 
A good example of such a bias is when surveys that do not have the same sky coverage are combined, similar to bias that the $V_e/V_a$ estimator of \citet{Avni:1980}, that is used in the process of measuring luminosity functions, addresses.

It is, technically, possible to devise loss functions that do not produce biased results when weighting data.
For example, if the terms in the loss function are given by $[P_{\text{model}}(i_m|x_m)-P_{\text{true}}(i_m|x_m)]^2$, then weighting will have the effect of improving the accuracy of the model probabilities near the up-weighted points, as desired. 
Notice that the counter-example requires that some way to calculate the true probability is already in hand, making the choice to use supervised learning on data questionable.
Regardless, any machine learning algorithm that both responds correctly to real clones in the data and will converge to the true $P(i|x)$ given infinite data (and, perhaps, infinite parameters in the model) should produce results that are biased when the loss function is weighted.

\subsection{Predicting Performance Metrics to Assess the Accuracy of Model Probabilities}\label{sec:theory:overshoot}
An important reason to base source classification on measurable probabilities is that it allows a users to predict the performance metrics that \textit{any} classification scheme will have on real data.
A mismatch between predicted metrics and observed ones on labeled data is a strong indication that the model probabilities are inaccurate.
We define the observed minus predicted model metric to be the \textit{overshoot} of the metric.

The basis of the method is to take the model's probabilities literally.
To simplify the derivation, assume that all sources have distinct features labeled $x_n$.
Each source can then be taken as an independent draw from its own categorical distribution with unknown results, so their expected outcomes and variances will just sum.
Suppose that a classifier assigns the label $j_n$ to each of the $N$ sources, then the expected number of true positives $\text{eTP}$, false positives $\text{eFP}$, and false negatives $\text{eFN}$ for class $i$ are given by
\begin{align}
	\mathrm{eTP}_i & = \sum_{n=1}^N \delta_{j_n,i} P(i = j_n|x_n),\nonumber \\
	\mathrm{eFP}_i & = \sum_{n=1}^N \delta_{j_n,i} P(i \neq j_n|x_n),\text{ and} \nonumber \\
	\mathrm{eFN}_i & = \sum_{n=1}^N (1-\delta_{j_n,i}) P(i = j_n|x_n), \label{eqn:expstats}
\end{align}
respectively.
The corresponding variances to Equations~\ref{eqn:expstats}, $\mathrm{vTP}_i$, $\mathrm{vFP}_i$, and $\mathrm{vFN}_i$, are given by replacing the probabilities in the sums with $P\times(1-P)$.
Note that the number of positives for class $i$, $\mathrm{NP}_i = \mathrm{eTP}_i+ \mathrm{eFP}_i$, is not random for our present purposes because it is the number of sources that the classifier labeled $i$.

Equations~\ref{eqn:expstats} can be used to construct asymptotically unbiased estimators for expected values of the most commonly used metrics in machine learning: completeness (also known as ``sensitivity" and ``recall"), reliability (also known as ``precision"), and $F_1\text{-score}$ as
\begin{align}
	C_{\mathrm{mdl}}(i) & = \frac{\mathrm{eTP}_i}{\mathrm{eTP}_i + \mathrm{eFN}_i}, \nonumber \\
	R_{\mathrm{mdl}}(i) & = \frac{\mathrm{eTP}_i}{\mathrm{NP}_i}, \text{ and} \nonumber \\
	F_{1,\mathrm{mdl}}(i) & = \frac{2\mathrm{eTP}_i}{N_i + \mathrm{eTP}_i + \mathrm{eFN}_i}, \label{eqn:CRF1mdl}
\end{align}
respectively. 
Their respective variances can be estimated using propagation of errors; they are:
\begin{align}
	\operatorname{Var}(C_{\mathrm{mdl}}(i)) & = \frac{\mathrm{vTP}_i\mathrm{eFN}_i^2 + \mathrm{vFN}_i\mathrm{eTP}_i^2}{\left(\mathrm{eTP}_i + \mathrm{eFP}_i\right)^4}, \nonumber \\
	\operatorname{Var}(R_{\mathrm{mdl}}(i)) & = \frac{\mathrm{vTP}_i}{N_i^2}, \text{ and} \nonumber \\
	\operatorname{Var}(F_{1,\mathrm{mdl}}(i)) & = 4\frac{\mathrm{vTP}_i\left(\mathrm{NP}_i + \mathrm{eFN}_i\right)^2 + \mathrm{vFN}_i \mathrm{eTP}_i^2}{\left(N_i + \mathrm{eTP}_i + \mathrm{eFN}_i\right)^4}. \label{eqn:CRF1mdlvar}
\end{align}

When the true class labels are known, the predicted metric $C_{\mathrm{mdl}}$, $R_{\mathrm{mdl}}$, and $F_{1,\mathrm{mdl}}$ can be compared with the observed ones $C_{\mathrm{obs}}$, $R_{\mathrm{obs}}$, and $F_{1,\mathrm{obs}}$.
Subtracting the predicted metric from the observed one produces the overshoot in the metric, and the variances from Equations~\ref{eqn:CRF1mdlvar} make it possible to assess whether the overshoot is statistically significant.
Since all of this assumes that the probabilities used were accurate, significant overshoot is strong evidence that the probabilities used to make the predictions are not accurate.

\subsection{Measuring and Replacing Priors: the Deweighting Procedure}\label{sec:theory:reweight}
The steps to replace a prior are: divide by the old prior, multiply by the new, then normalize, thus
\begin{align}
	P(i|x) & = \frac{\frac{P(i)}{P_w(i)} P_w(i|x)}{\sum_j \frac{P(j)}{P_w(j)} P_w(j|x)}.\label{eqn:deweight}
\end{align}
Since, for a weighted data set, $P_w(i)\propto w_i P(i)$, we refer to Equation~\ref{eqn:deweight} as ``deweighting" the predictions.
With real data and real models, Equation~\ref{eqn:deweight} will only work perfectly in the limit of infinite data and, potentially, infinite fit parameters. 
In more realistic situations, deweighting the output of real models is statistically valid for the data as a whole, and may over or under-correct for any given source.

Equation~\ref{eqn:deweight} is a way to remove a known bias when the true prior $P(i)$ is known. 
This can be useful when weighting otherwise unbiased training data.
A more common occurrence is for the training data, itself, to be of biased composition.
In that case the bias to the prior may not be known, exactly.
One way to detect the bias is to compare an estimate of the prior used in training the machine learning algorithm to an estimate of the prior from the broader data set the model is applied to.
Doing so strays a bit into a semi-supervised machine learning algorithm.

For example, denote the prior of a particular model calculated using Equation~\ref{eqn:PriorWeightDef} as $P_M(i)$ and the probabilities predicted by the model as $P_M(i|x)$.
If the model is then applied to a data set with $N$ total sources, then the true expected prior for the data set is given by
\begin{align}
	P(i) & = \int \mathcal{L}(i,x)\d x \nonumber\\
	& = \int P(i|x)\, \mathcal{L}(x)\d x \nonumber \\
	& \equiv \mathrm{E}_x \left(P(i|x)\right) \nonumber \\
	& \approx \frac{1}{N} \sum_{m=1}^N P_M(i | x_m), \label{eqn:PriorEst2}
\end{align}
where the last line of Equation~\ref{eqn:PriorEst2} is replacing the expected value with the sample average.
If $P_M(i)$ and the output of \ref{eqn:PriorEst2} differ to a statistically significant degree then that is good evidence that the prior is inconsistent with the data set the model is being applied to.
Note that this is only the case if the domain of the data features $x$ are a match. 
If the domain of application significantly differs from the domain used for training then the priors are not actually comparable.

\begin{figure*}[htb]
	\begin{center}
	\includegraphics[width=\textwidth]{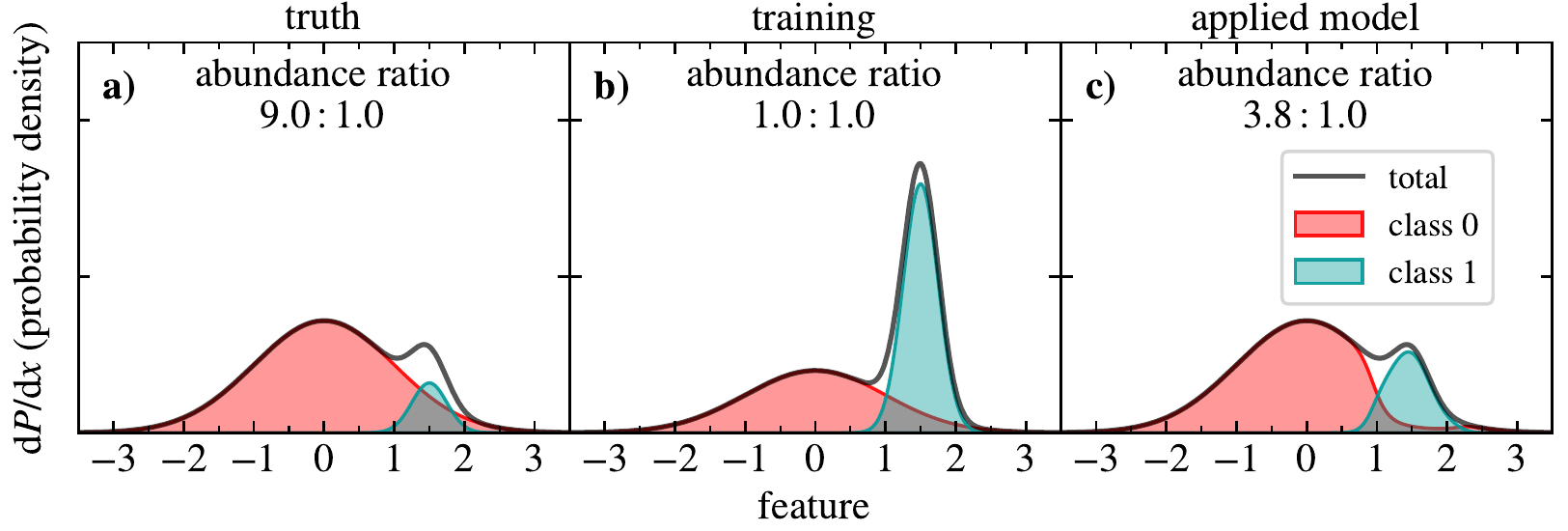}
	\end{center}
	
	\caption{Prediction Consistent Prior Illustration}
	{The classification process for sources drawn from two Gaussians in illustrated form, including the way weighting the Gaussians to balance them shifts the probabilities (panels (\textbf{a}) and (\textbf{b}) for unweighted and weighted, respectively). 
	Panels (\textbf{a}) and (\textbf{b}) are recreations of Figure~\ref{fig:2gauss}, and Panel~(\textbf{c}) shows the results of applying the weighted model to unweighted and unlabeled data. 
	For this illustration, class 0 is from a Gaussian with mean 0 standard deviation 1, and contributes 90\% of the sources, while class 1 has mean 1.5 and standard deviation 0.25.
	}
	\label{fig:priorcomp}
\end{figure*}

An illustration of the comparison described in the last paragraph can be found in Figure~\ref{fig:priorcomp}.
The abundance ratio for Panel~(\textbf{b}) is a proxy for the probabilities $P_M(i)$, and the ratio from Panel~(\textbf{c}) is a proxy for Equation~\ref{eqn:PriorEst2}. 
Their disagreement is a signal that the model is using the wrong prior. 

When there is clear evidence that the model's prior is a mismatch to the unlabeled data it is possible to find a prior that is consistent with the remaining aspects of the model, which we call a prediction consistent prior (PCP).
One way to do that is by combining Equations~\ref{eqn:deweight} and \ref{eqn:PriorEst2} to get
\begin{align}
	P(i) & = \frac{1}{N} \sum_{m=1}^N \frac{\frac{P(i)}{P_M(i)} P_M(i|x_m)}{\sum_j \frac{P(j)}{P_M(j)} P_M(j|x_m)}. \label{eqn:Pconsist}
\end{align}
Although it may be possible, in principle, to solve Equation~\ref{eqn:Pconsist} for $P(i)$ analytically, it is more practical to do so numerically using an algorithm like recursion or Newton's root finding algorithm.
The recursion method to solve Equation~\ref{eqn:Pconsist} starts with any reasonable estimate for $P(i)$, like $P_M(i)$ or all $P(i)$ equal.
The next step is to insert it into the right hand side (RHS) of Equation~\ref{eqn:Pconsist}.
Computing the sums on the RHS will give a new estimate of $P(i)$ that can then be inserted into the RHS of Equation~\ref{eqn:Pconsist} to get a new estimate.
This process can be repeated until some desired convergence level is reached.
This is a special case of the Expectation Maximization algorithm \citet{Dempster:1977}\footnote{For a more recent tutorial see \citet{Roche:2011}.}; one for which the models for the individual classes do not need updating, thus the name half-Expectation Maximization algorithm fits.

The recursion algorithm is guaranteed to converge, but that convergence can be slow \citep{Wu:1983}.
It is also not obvious that the solution it converges to is unique.
We prove that it is starting from the negative log-likelihood of an unlabeled data set
\begin{align}
	L (\{\text{data}\})&= -\sum_{m=1}^N \ln\left(\sum_i P(i) \mathcal{L}(x_m|i)\right) \nonumber\\
	&\hphantom{=}- \lambda\left(1-\sum_i P(i)\right), \label{eqn:unlblLoss}
\end{align}
with $\lambda$ a Lagrange multiplier that enforces normalization for the prior probabilities. 
Taking the derivatives of Equation~\ref{eqn:unlblLoss} with respect to $P(i)$ and $\lambda$ and setting them equal to zero yields $\lambda=N$ and equations equivalent to Equations~\ref{eqn:Pconsist}, as long as $P(i|x_m) = P_M(i) \mathcal{L}(x_m|i) / \mathcal{L}(x)$.
The matrix of second partials has elements
\begin{align}
	\frac{\partial^2 L}{\partial P(j) \partial P(k)} & = \sum_{m=1}^N \frac{\mathcal{L}(x_m|j) \mathcal{L}(x_m|k)}{\left[\sum_i P(i) \mathcal{L}(x_m|i)\right]^2}. \label{eqn:Hessian}
\end{align}
Because the likelihoods are all positive semi-definite, it is straightforward to show that the matrix is positive semi-definite in general, and positive definite when the set of predictions $\{P(i|x_m)\}$ spans the vector space indexed by $i$.
This means that the likelihood is a convex function over the parameters $P(i)$, which means its minimum is unique.
Because the region the $P(i)$ are also confined to is also convex, the maximum of the likelihood in that set will also be unique.
Along any direction in the prediction space not spanned by the predictions $L$ will be flat, permitting any solution along that direction.


The speed of convergence will depend on the size of the eigenvalues of the matrix of second partials, with both methods converging faster the farther from zero the eigenvalues are.
We also note that all $x$ dependent biases need to be the same for all PDFs for the prediction consistent prior to be fully effective.
$x$-dependent biases change the shape of the PDFs that the machine learning algorithm sees, changing $P_w(i|x)$ in ways that adjusting the global prior cannot compensate for.

\section{Data and Methods}\label{sec:data}
In this paper, we present four empirical demonstrations of the methods from Section~\ref{sec:theory}.
Section~\ref{sec:dat:sim} describes a tests on simple simulated data.
The reason for using a simulation is because it gives a case where the true probabilities $P(i|x)$ the model should be producing can be analytically calculated.
This allows for a particularly sensitive determination of any biases affecting the model.
The simulation needs to be simple because simplicity makes it easy to maximize the size of the regions where the classes overlap, increasing the impact of deweighting.

Section~\ref{sec:dat:mla} describes the details of the machine learning algorithm we used to produce classifiers in this work.
The algorithm used is a type of feed-forward neural network called a multi-layer perceptron. 
For each classification task we manually tuned the network's hyperparameters (specifically, the details of the data augmentation layer, the number of hidden layers, and the number of neurons per layer) in order to achieve good performance for each task.
The details of each tuning are given in their respective subsections.

Section~\ref{sec:dat:SGQ} describes a test on a similarly simple real data set with a moderately rare class: classifying sources as stars, galaxies, or AGN based on photometry.
AGN are actually a sub-type of galaxy; one where the matter falling into the super-massive black hole in the center is so hot and bright that it is comparable to the brightness of the rest of the galaxy.
This means that galaxies and AGN will have a lot of inherent similarity because the dividing line between them is not sharp, in reality.
This is the case because every known AGN is at the center of some host galaxy. 
Thus, when using the sort of moderate to low resolution photometry being used here, the light from every AGN will be contaminated by some amount of light from its host galaxy.
The exact level of contamination will depend sensitively on which wavelengths are being sampled, but that just affects at what total luminosities the light at that wavelength from the AGN and host are comparable. 
Thus, for any given wavelength, host SED, and AGN SED there is an accretion rate for the AGN where it will have the same observed flux at that wavelength as its host.
While the shape of the AGN's SED may also depend on accretion rate, it will, nevertheless, exist on a continuum that blends smoothly into the population of galaxies for which we cannot, yet, detect light from the accretion of the super-massive black hole at their center.

Like with the simulated data, this comparison is done in an artificially simplified space: $\mathrm{W1} - \mathrm{W2}$ versus $J-\mathrm{W1}$.
The $\mathrm{W1}$ filter measures the brightness of an object in light with a wavelength of approximately $3.4\micron$, $\mathrm{W2}$ detects light near $4.6\micron$, and $J$ near $1.2\micron$ \citep{Wright:2010,Skrutskie:2006}.
The motivation for choosing this color space was that \cite{Kovacs:2015} showed $J-\mathrm{W1}$ to be a good single color for separating stars from galaxies, and \cite{Stern:2012} showed $\mathrm{W1} - \mathrm{W2}$ to be a good single color for selecting AGN.
The rationale for intentionally limiting our model to this simple color space, artificially, is precisely because keeping distributions of different source types overlapping makes the effect that deweighting compensates for most apparent.
Some examples of features that could be added that would greatly increase the accuracy of a classification model include: the W1 magnitude (Figure 8, Panel a of \cite{Lake:2019} shows that at Galactic latitude $b > 30^\circ$ a source with $\mathrm{W1} < 18.2\Mag$ is more likely to be a star, and fainter sources are more likely extragalactic), pure morphological information (e.g. the $\chi^2$ information from 2MASS or AllWISE for how well the psf fit the source can indicate a source is extended, or contaminated, \citep{Cutri:2006,Cutri:2013}), parallaxes (by cross matching to, for example, \cite{Gaia:2018}), or just adding more colors.

Section~\ref{sec:dat:blazars} describes an expansion of the star/galaxy/AGN problem to include blazars, specifically flat spectrum radio quasars (FSRQs) and BL Lacertae objects (BL Lacs).
In the unified scheme described in \cite{Urry:1995}, Blazars are a special class of AGN where the magnetic fields coming from the super-massive black hole are forming a collimated jet of radiation that is pointed at us, making them unusually bright in radio and gamma radiation.
The base data set is the same as used in the star/galaxy/AGN problem, but the color space is different and two separate catalogs of blazars are used to identify which of the sources are blazars.
The color space used is: $\mathrm{W3}-\mathrm{W2}$, $\mathrm{W2}-\mathrm{W1}$, $r-\mathrm{W1}$, $r-i$, and $g-r$.
The new filters are sensitive to light at wavelengths at approximately: $12\micron$ for $\mathrm{W3}$, $460\nm$ for $g$, $610\nm$ for $r$, and $750\nm$ for $i$ \citep{Wright:2010,Doi:2010}.
The three \textit{WISE} filters are chosen to enable use of the space shown in the ``bubble plot" first shown in \citet{Wright:2010}.
The SDSS filters $r$ and $i$ were chosen for the $r-i$ versus $r-\mathrm{W1}$ color space \cite{Lake:2012} used to characterize the sources in the \textit{WISE}/DEIMOS spectroscopic survey, and $g$ was added for being comparably sensitive to $r$.
The reason for using a greatly expanded color space is because our inspection of many \textit{WISE}-SDSS color--color diagrams shows very little separation between the general AGN and blazar data, and it is hoped that accumulating multiple small-separation pieces of evidence can give the machine learning algorithm the maximum chance to be able to distinguish the blazars from the rest of the AGN.

Section~\ref{sec:dat:Clarke} describes a short test in constructing a PCP from the catalog of \citet{Clarke:2020}.

\subsection{Machine Learning Algorithm}\label{sec:dat:mla}
We implemented the machine learning algorithm, a multi-layer perceptron neural network, used to classify points in this data set in \code{Python} 3.6.9.
The packages used to implement and test it were: \code{PyTorch} 1.4.0 \citep{pytorch}, \code{numpy} 1.17.4, and \code{matplotlib} 3.2.1 \citep{matplotlib}.
The machine on which we ran the software was a typical laptop.\footnote{MacBook Pro (15-inch monitor, 2018, with a 2.2 GHz Intel Core i7 6 core processor, an Intel UHD Graphics 630 GPU, and a Radeon Pro 555X GPU) running maxOS Mojave (version 10.14.6) and Xcode 10.3.}

The machine learning algorithm used has a five step feed-forward structure: scale all features to be in the interval \mbox{[-1,1]}, augment features, run a multi-layer perceptron neural network, convert the output of the network from log-odds to probability, and classify based on probabilities. 
We have published the version of the script used to train this neural network on simulated data, alongside the script used to generate the simulated data, on Figshare under \href{https://doi.org/10.6084/m9.figshare.20237832}{DOI:10.6084/m9.figshare.20237832}.
We provide details of each step in the following paragraphs.

The feature scaling step consists of finding a linear transformation that maps the range of the training data to the interval \mbox{$[-1,1]$}.
This normalization scheme was chosen without evaluating alternatives (for example, mapping the standard deviation to 1 instead of the range) after testing had shown that normalizing provided a large increase in asymptotic accuracy for large training sets.
We also found that normalizing reduced the incidence of getting stuck in local minima.
It is considered common knowledge in the area of machine learning that normalizing the input to neural networks is critical to optimizing their performance; see, for example, \cite{Sola:1997}.

The augment features step consists of appending to each data point the terms of a polynomial out to second order. 
In other words, each vector of features now consists of lists with elements $x_0^{n_0} x_1^{n_1} x_2^{n_2} x_3^{n_3}$ .
In the interpolation convention for second order augmentation, each $n_k$ takes all values between $0$ and $2$ independently (excluding the constant one, where all $n_k$ are zero).
In the mathematician's convention, the $n_k$ must sum to either 1 or 2.
This was done because manual examination of the data spaces shows the data to be confined to compact regions in the feature spaces.
Thus the natural coordinates for describing relative densities among the data classes should be curved.
Using quadratic feature augmentation allows the decision boundaries of individual neurons to be conic sections (especially circles and ellipses).
Without this step, the neurons' decision boundaries would consist of flat hyper-planes.
Thus, the network should require fewer neurons and layers, at the cost of more parameters per neuron in the initial layer.
If the feature spaces used in this work had been higher dimensional then it may not have been a good tradeoff.

The neural network used in this experiment is a fully connected one, also called a ``multi-layer perceptron."
The activation function used for the neural network is the rectified linear unit (ReLU), which can be written in terms of the unit step function $\Theta(x)$ as $x\Theta(x)$.
This model was chosen because neural networks have an abundance of publicly available software tools to implement them \citep[e.g. pytorch and scikit-learn][]{pytorch,Pedregosa:2021} with good tutorials for how to perform logistic regression using them, and mathematical proof that they can approximate any function with arbitrary accuracy given enough training data and resources \citep{Hornik:1989}.

The output of the neural network was interpreted as the natural logarithm of the odds that the source is in the given class, versus any with numerically greater index, adjusted so that when all of the network's parameters are zero it will predict all classes to be equally probable. 
In formulae, if $y_i$ is the network's output with index $i$ then they are related to the probabilities by
\begin{align}
	y_i & = \ln \left(\frac{P(i | x)}{\sum_{j>i} P(j|x)}\right) + b_i,\nonumber \\
	P(i|x) & = \left\{\begin{array}{cc}
	 \frac{1}{1+e^{-(y_i-b_i)}} \prod_{j=0}^{i-1} \frac{1}{1+e^{y_j-b_j}} & i < N_i-1 \\
	 \prod_{j=0}^{i-1} \frac{1}{1+e^{y_j-b_j}} & i = N_i-1 \end{array} \right. , \text{ and} \label{eqn:logoddsprob}\\
	 b_i & = \ln(N_i - 1 - i), \nonumber 
\end{align}
where $N_i$ is the total number of classes and the index $i$ ranges from $0$ to $N_i-1$.

Using the convention in Equation~\ref{eqn:logoddsprob} for the network's output is a little more complicated to code than the standard element-wise logit of probabilities, but it has the advantage that none of the model parameters are rendered irrelevant.
Probabilities have a constraint that they must sum to 1.
The standard technique for calculating probabilities from the output layer of a neural network is for the output layer to have $N_i$ outputs and then evaluate the softmax function on them
\begin{align}
	P(i|x) &= \frac{e^{y_i} }{\sum_{j=0}^{N_i-1} e^{y_j}}. \label{eqn:softmax}
\end{align}
Evaluating the gradient $\frac{\partial P(i|x)}{\partial y_k}$, a necessary step in back-propagation \citep{leCun:1988}, shows that the constraint makes the gradient, as a matrix, singular.
This means that there is a direction in which the parameter update cannot perturb the network's parameters in the output layer.
This singularity will propagate backwards throughout the network.
We make no attempt to quantify how many parameters in a given network are rendered irrelevant by the use of softmax on the output, but it should be clear that it is more than zero.

The classification scheme used was a stochastic classifier: the source was assigned to a class using the probabilities the machine gave for it belonging to each class.
We did this after finding that the common bayes classifier (assign sources to whichever probability is highest for that source), was sacrificing rare source completeness and overall demographic accuracy in favor of reliability.

The software library used to implement the neural network and facilitate optimization is \code{PyTorch} version 1.4.0 \citep{pytorch}.
The loss function used to fit the neural network's parameters was the cross entropy.
The model parameter initialization was handled by \code{PyTorch}'s built in initializer with default settings.
The optimization step used \code{PyTorch}'s implementation of Adam, \code{torch.optim.adam}, described by \citet{adamoptimizer}, with the learning rate $\code{lr}=2\times 10^{-3}$ and \code{weight\_decay} parameter set to $10^{-4}$.
Setting the \code{weight\_decay} parameter gives the algorithm a bias towards the model parameters all being zero, reducing the variability of poorly constrained model parameters, and giving an additional reason for the $b_k$ to be non-zero in Equation~\ref{eqn:logoddsprob}.
Weight decay is a standard method of model regularization, that is, reducing the model variance in the face of insufficient data to determine all model parameters.
Mathematically, it is equivalent to adding a term proportional to the sum of the square of the model parameters to the loss, and the constant of proportionality is $\lambda / (2 N_p)$, where $\lambda$ is the weight decay and $N_p$ is the number of parameters.
This guarantees that there is always at least one optimum value for the parameters to converge to.
The machine learning algorithm was considered to have converged when the improvement in the loss produced by an Adam step was less than $10^{-8}$, or when it had taken 25,000 iterations.
The learning process was repeated five times, with only the lowest loss model kept, to reduce the chance of the model getting stuck in a local minimum.

\subsection{Simulations}\label{sec:dat:sim}
The simulations are meant to mimic a 4-dimensional color space where the data are drawn from three overlapping clusters, and each cluster is a distinct class.
Specifically, each data point has four Gaussian distributed random variables as features, only the first three of which are relevant.
The last dimension is irrelevant because the data are all distributed identically in that dimension. 
This was done to add some of the challenge machine learning algorithms have to overcome when analyzing real data.
The pseudo-random numbers for the simulations were generated using the \code{numpy} package version 1.17.4 \citep{numpy} called from \code{Python} 3.6.9.

The parameters of the Gaussian distributions, means ($\mu_i$) and standard deviations ($\sigma_i$) along dimension $i$, split by source class, are in Table~\ref{tbl:simpars}.
The fraction of points that each source contributed to each simulation is given in Table~\ref{tbl:simfracs}.
The fractions in Table~\ref{tbl:simfracs} were chosen to be similar to the real data set described in Section~\ref{sec:dat:SGQ}.
The column labeled ``fraction" is the nominal fraction contributed by sources of that class, and the remaining columns list the number of actual sources used in the simulations, as randomly generated parts of a non-random total.
Note that no development set was generated to assist in manually tuning hyperparameters because the simulations were fast, and so could be re-done after all of the hyperparameter tuning was done.

The `biased' simulation was one where the fraction of points simulated was equal for each class.
This provides a useful point of comparison with the weighted fit because, following \citet{Clarke:2020}, the weights used were the inverse of each class's fraction over the overall data,
\begin{equation}
	w_i = \frac{1}{P(i)}. \label{eqn:weightsused}
\end{equation}
Thus, the weighted fits are set to mimic a fit to data with a composition with equal contribution from each class.

\begin{figure*}[htb]
	\begin{center}
	\includegraphics[width=\textwidth]{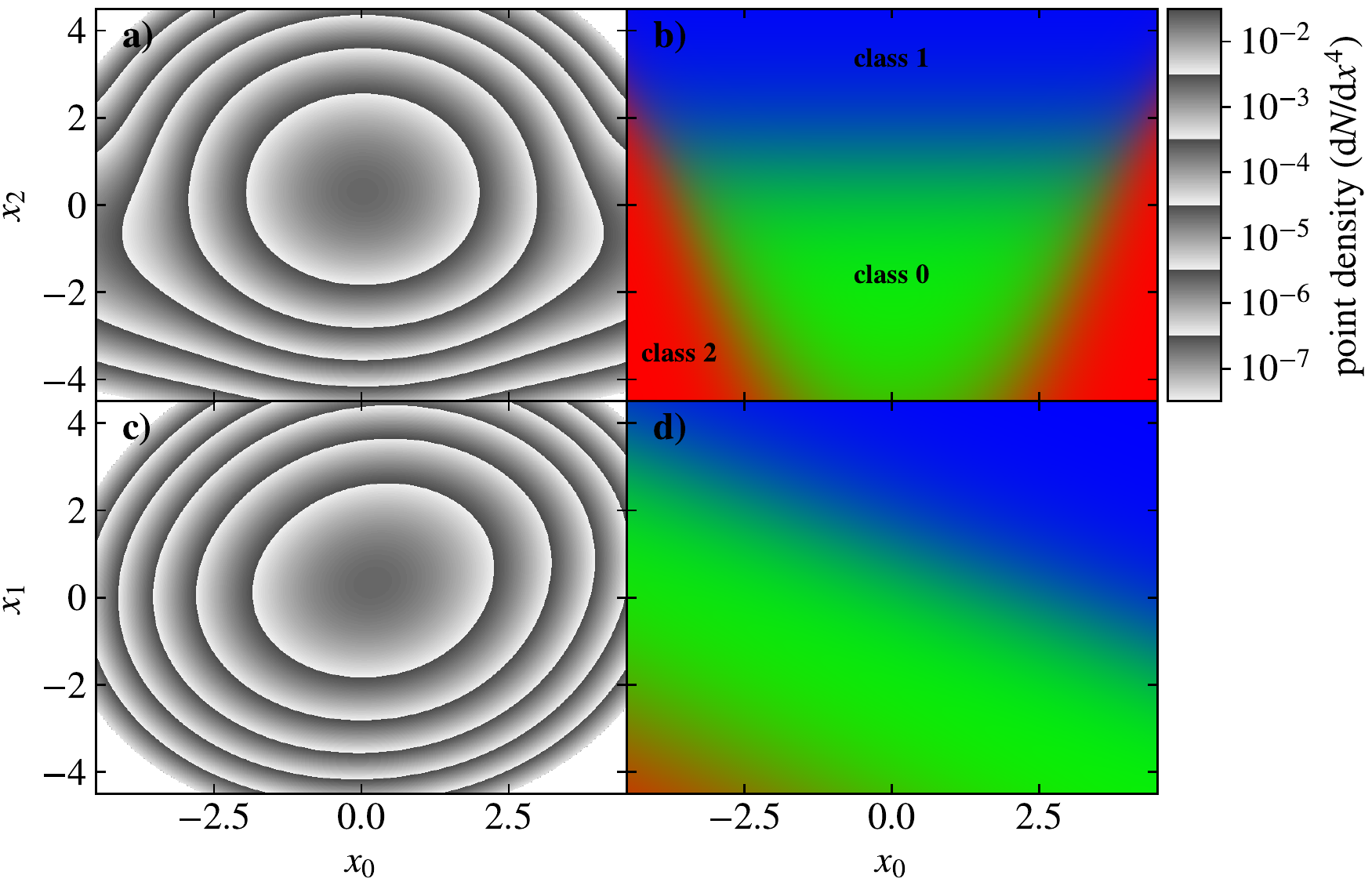}
	\end{center}
	
	\caption{Simulation Slices}{ Slices from the simulated data space.
	The lack of visible clustering is intentionally done to maximize the impact of weighting on the model the machine learning algorithm produces.
	The left column shows the net density of points (log scaled), and the right column shows the relative density of each class, encoded in the color with the classes 0, 1, and 2 set to green, blue, and red, respectively (green is near the origin, blue the top, and red the lower outsides).
	When inspecting the digital version of the figure, the red--green--blue (RGB) values of each point will sum to 255, and the fraction of that sum contributed by each basic color represents the probability of the corresponding class at that point.
	The top row is the $(x_0,x_2)$ space with $x_1 = 0$ and $x_3=0$.
	Similarly the bottom row is $(x_0,x_1)$ with $x_2=0$ and $x_3=0$. }
	\label{fig:simregions}
\end{figure*}

Along each axis there is a tremendous amount of overlap among the 1-dimensional Gaussians, and the boundaries between regions where each class dominates are complicated.
The complications come from giving class 2 a wider spread along dimension 2.
Slices of the space computed from its theoretical properties are shown in Figure~\ref{fig:simregions}.
The key takeaway from Figure~\ref{fig:simregions} is that the classes are very confused, that is, there is no clear clustering in the density plots (left column) and the boundaries between regions where a single class predominates are gradual (regions of single color in the right column). This is especially true for class 2, which only dominates out on the wings where the density of points is down from the peak by about 4 orders of magnitude.

\begin{table}[htb]
\begin{center}
\caption{Simulation Gaussian Parameters}\label{tbl:simpars}
\begin{tabular}{|c | m{8mm} m{8mm} m{4mm} m{4mm} | m{4mm} m{4mm} m{4mm} m{4mm} |}
\hline\hline
Class & $\mu_0\vphantom{\mu}^\mathrm{a}$ & $\mu_1\vphantom{\mu}^\mathrm{a}$ & $\mu_2\vphantom{\mu}^\mathrm{a}$ & $\mu_3\vphantom{\mu}^\mathrm{a}$ & $\sigma_0\vphantom{\sigma}^\mathrm{b}$ & $\sigma_1\vphantom{\sigma}^\mathrm{b}$ & $\sigma_2\vphantom{\sigma}^\mathrm{b}$ & $\sigma_3\vphantom{\sigma}^\mathrm{b}$ \\
\hline
0 & \hphantom{$-$}0 & \hphantom{$-$}0 & 0 & 0 & 1 & 1 & 1 & 1 \\
1 & \hphantom{$-$}1 & \hphantom{$-$}0.5 & 0 & 0 & 1 & 1 & 1 & 1 \\
2 & $-0.75$ & $-0.5$ & 0 & 0 & 1 & 1 & 2 & 1\\
\hline
\end{tabular} \\
\end{center}
$^\mathrm{a}$ Mean value of of a Gaussian distribution along the given dimension (subscript).\\
$^\mathrm{b}$ Standard deviation of a Gaussian distribution along the given dimension (subscript).

\end{table}

\begin{table}[htb]
\begin{center}
\caption{Simulation Class Characteristics}\label{tbl:simfracs}
\begin{tabular}{| c | lrrr |}
\hline\hline
Class & fraction & $N_{\mathrm{train}}\vphantom{N}^\mathrm{a}$ & $N_{\mathrm{biased}}\vphantom{N}^\mathrm{b}$ & $N_{\mathrm{test}}\vphantom{N}^\mathrm{c}$  \\
\hline
0 & $0.6$ & 599,622 & 332,724 & 11,962 \\
1 & $0.38$ & 380,367 & 333,046 & 7,657 \\
2 & $0.02$ & 20,011 & 334,230 & 381 \\
\hline
all & 1 & $10^6$ & $10^6$ & 20,000 \\
\hline
\end{tabular} \\
\end{center}
$^\mathrm{a}$ Number of sources in the training set.\\
$^\mathrm{b}$ Number of sources in the biased training set.\\
$^\mathrm{c}$ Number of sources in the testing set.

\end{table}

To classify the simulated data we used the interpolation convention second order data augmentation step.
This increases the number of features fed to the neural network to $3^4 - 1 = 80$ from 4.
The neural network had exactly one hidden layer of 32 neurons.
These choices produced a model with $[(3^4 - 1)\times 32 + 32] + [32 \times 2 + 2] = 2,658$ total free parameters.

\subsection{Star/Galaxy/AGN Classification}\label{sec:dat:SGQ}
\begin{figure}[htb]
	\begin{center}
	\includegraphics[width=0.48\textwidth]{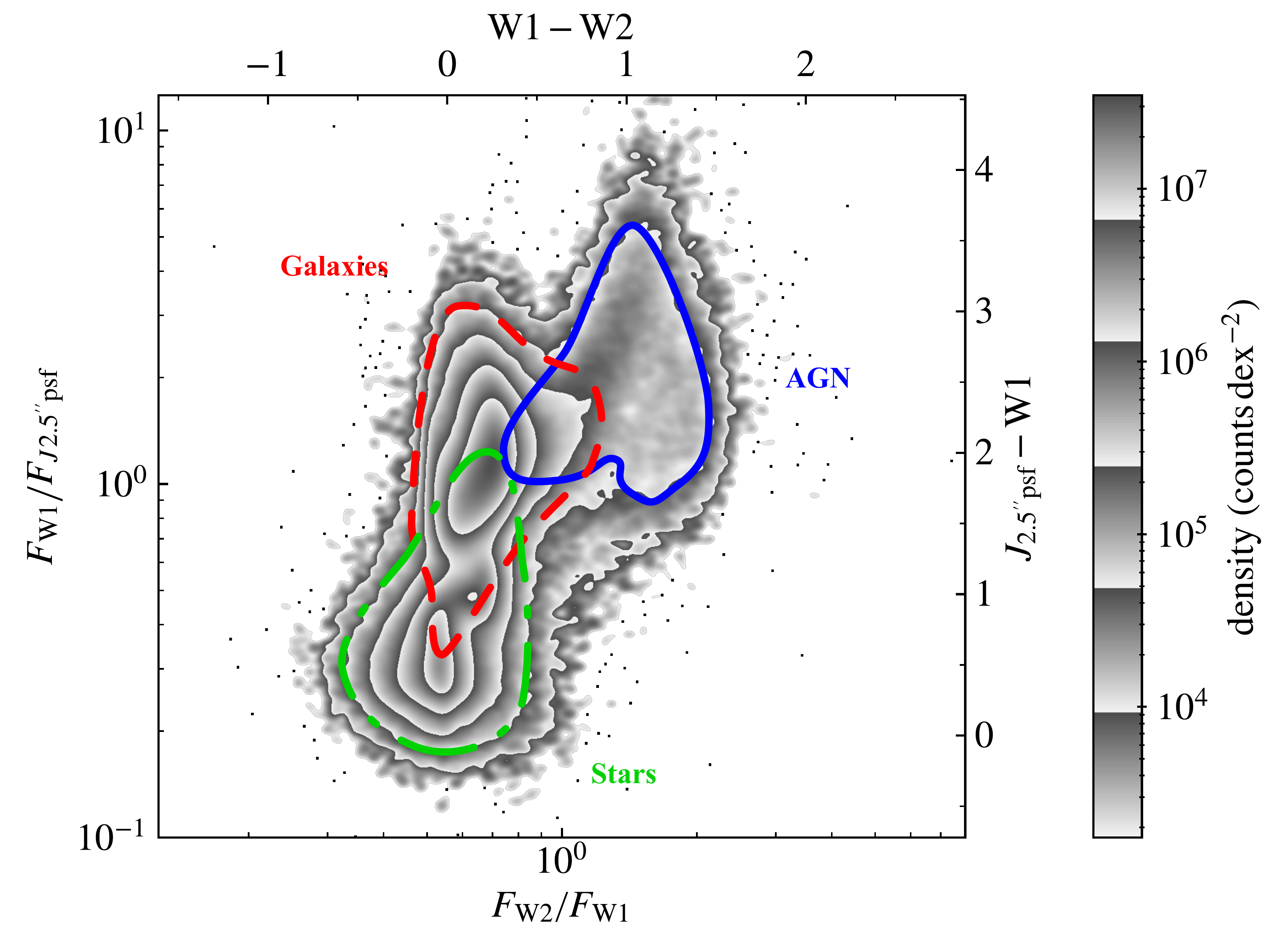}
	\end{center}
	
	\caption{\textit{WISE}-2MASS color--color diagram}
	{The distribution in $J-\mathrm{W1}$ versus $\mathrm{W1}-\mathrm{W2}$ space of all sources in the SDSS data release 16 \code{SpecObj} table view that have clean photometry in AllWISE and a minimum signal-to-noise ratio ($\mathrm{SNR}>5$) in $J$, W1, and W2, smoothed with a $0.01\operatorname{dex}$ Gaussian kernel in each direction where individual points are not plotted.
	 The sources divide into three major groups, with most stars near the stellar locus at the bottom of the plot, most galaxies directly above them, and most AGN in the lobe that projects above and to the right of the galaxies.
	 The colored contours follow, approximately, the isodensity contours at approximately $5\times 10^4\operatorname{counts}\operatorname{dex}^{-2}$ for the classes labeled (the red dashed line is for galaxies, the green dash-dotted line is for stars, and the blue solid line is for AGN). }
	\label{fig:W2Mcolors}
\end{figure}

The first real data case presented here uses the data shown in Figure~\ref{fig:W2Mcolors} for the star/galaxy/AGN classification problem.
The sources used were the entirety of the SDSS data release 16 \citep[SDSSdr16,][]{SDSSdr16} table named ``\code{SpecObj}" cross-matched to data from \textit{WISE}'s AllWISE data release \citep{Wright:2010,Cutri:2013} and the 2MASS \citep{Skrutskie:2006}.
The sources were cross-matched to the AllWISE database on the Infrared Science Archive (IRSA) using a radius of $3.0''$. 
When the SDSS table named ``TwoMass" had a matching entry for a source it was used for 2MASS photometry, and the AllWISE database provided the photometry otherwise.

For the minority of sources with bad W1 flux uncertainties\footnote{\url{http://wise2.ipac.caltech.edu/docs/release/allwise/expsup/sec2_2.html\#w1sat}} they were replaced using the same procedure as used in \citet{Lake:2018}.
In brief, if we let $F_{\mathrm{W1}}$ be the value from the column \code{w1flux}, $\sigma_{\mathrm{W1}}$ be the value from the column \code{w1sigflux}, $F_{0\ \mathrm{W1}} = 10^{0.4\times20.5}\operatorname{dn}$\footnote{\url{https://wise2.ipac.caltech.edu/docs/release/allsky/expsup/sec2\_3f.html}} is the AllWISE Atlas magnitude zero flux, and $\sigma_{\mathrm{W1p2}} \equiv 0.4 \ln(10) [\code{w1sigp2}] F_{0\ \mathrm{W1}} 10^{-0.4[\code{w1magp}]}$, a flux uncertainty estimated from the columns \code{w1magp} and \code{w1sigp2}, then we replace $\sigma_{\mathrm{W1}}$ with $\sigma_{\mathrm{W1p2}}$ when the following inequality is satisfied:
\begin{align}
	\sigma_{\mathrm{W1}} &> 2\sqrt{(0.02 F_{\mathrm{W1}})^2 + \sigma_{\mathrm{W1p2}}^2}. \label{eqn:w1uncrep}
\end{align}
The values $2$ and $0.02$ in Equation~\ref{eqn:w1uncrep} were empirically determined based on when one uncertainty estimate would typically outperform the other, with a bias toward using the standard uncertainty when they are close.

In addition to having signal-to-noise ratio (SNR) greater than 5 in W1, W2, and J, the sources were also required to satisfy: $\code{w1cc\_map}=0$ (the contamination and confusion bit map containing all quality and artifact flags checked for in the W1 images during data processing, a value of $0$ means no issues were found for the source, for details see the \textit{WISE} All-Sky column description\footnote{\url{https://wise2.ipac.caltech.edu/docs/release/allsky/expsup/sec2\_4ci.html}}), $\code{w2cc\_map}=0$ (the same as \code{w1cc\_map}, but for W2), and the SDSS \code{zwarning} flags must be either 0 or 16. 
\code{zwarning} is the bit map where quality flags from SDSS's processing of object spectra are stored, see the SDSS dr16 Schema browser for details on their meaning.\footnote{\url{https://skyserver.sdss.org/dr16/en/help/browser/browser.aspx}}
The \code{zwarning} flags were chosen following the selection criteria in \citet{Clarke:2020}.
When \code{zwarning} flag at bit number 5 (bit mask 16) is set when the spectrum has: ``fraction of points more than 5 sigma away from best model is too large $(> 0.05)$."
\citet{Clarke:2020} justify not rejecting sources with this flag, stating: ``If the flag is 16, this indicates the ``MANY\_OUTLIERS" warning which is only present for data taken with the SDSS spectrograph and not with the BOSS spectrograph, and usually indicates a high signal-to-noise spectrum or broad emission lines in a galaxy. Consequently, it rarely signifies a true error."

Note that the colors in Figure~\ref{fig:W2Mcolors} are not physical because no attempt is made to do resolved source or matched aperture photometry; the fluxes used were the point spread function (psf) photometry from both catalogs.
Thus, the $J-\mathrm{W1}$ axis is a combination of both color and morphological information.
The key feature to notice in Figure~\ref{fig:W2Mcolors} is that it separates out stars (bottom, dash-dotted green contour) from galaxies (above stars, dashed red contour) and AGN (projecting up and to the right of the galaxies, solid blue contour), but not completely; there is still overlap among the three major source classes.

The characteristics of the resulting 1,167,803 sources can be found in Table~\ref{tbl:SDSSdat}. 
The sources were randomly divided into training (train), development (dev), and test sets in an 80\%, 10\%, 10\% split, respectively.
During development and tweaking of the machine learning algorithm, the test set was left untouched and unexamined, with dev serving as a stand-in.
The purpose of using the dev set in place of the test set during algorithm development is to keep the assessments based on the actual test set unbiased by what would amount to a fitting process of the model's hyper-parameters to the test set.
The performance on the test set was only evaluated just before publication, at which time the dev set had completed its job as a proxy test set and was merged in with the train one in order to maximize the amount of training data used to produce the final model.

We chose to use data from the SDSS for demonstration because it provides large database with spectroscopically classified sources  to use as the true target values for the machine learning algorithm.
Spectroscopic identification is particularly important because it is considered a high fidelity classification technique.
The selection of objects in the \code{SpecObj} table view, which contains all of SDSS's spectroscopically characterized sources, is heterogeneous among multiple surveys (sdss/legacy, boss, eboss, and segue) with several programs, each.
Each program had its own targeting criteria based on factors including: color, magnitude, morphology, and sky coverage.\footnote{For example, compare the ``DR16 Coverage Figures" at \url{https://www.sdss.org/dr16/} or \citet{SDSSdr16}}
All of these complications mean that this data cannot be considered to be from a single parent sample.
The model we generate is, therefore, not an accurate picture of reality.
Hence, the reason why we are publishing neither the trained model nor a catalog from applying it to the AllWISE database.
It is, nevertheless, a sufficiently realistic scenario with which to test the performance of the techniques in Section~\ref{sec:theory}.

\begin{table}[htb]
\begin{center}
\caption{SDSS Data Characteristics}\label{tbl:SDSSdat}
\begin{tabular}{| l | lrrr |}
\hline\hline
Class & fraction & $N_{\mathrm{train}}\vphantom{N}^\mathrm{a}$ & $N_{\mathrm{dev}}\vphantom{N}^\mathrm{b}$ & $N_{\mathrm{test}}\vphantom{N}^\mathrm{c}$  \\
\hline
Galaxy & $0.688$ & 642,858 & 80,348 & 80,188 \\
Star & $0.291$ & 272,246 & 34,055 & 34,089 \\
AGN & $0.021$ & 19,139 & 2,377 & 2,503 \\
\hline
total & 1 & 934,243 & 116,780 & 116,780 \\
\hline
\end{tabular} \\
\end{center}
$^\mathrm{a}$ Number of sources in the training set.\\
$^\mathrm{b}$ Number of sources in the development set.\\
$^\mathrm{c}$ Number of sources in the testing set.

\end{table}

This data was augmented to second order using interpolation convention counting.
Because the shape of the clusters of points in this real data is far more complicated than the simulated data, the neural network had two hidden layers of 128 and 32 neurons.
This number of neurons was chosen by adding a second layer to cope with the additional complexity in the data shape, and then doubling the number of neurons in each layer until the regions assigned to each class were subjectively accurate.
The number of parameters in the model is: $[(3^2 - 1)\times 128 + 128] + [128 \times 32 + 32] + [32 \times 2 + 2] = 5,346$.

\subsection{Blazar Classification---Test Case for Extremely Rare Populations}\label{sec:dat:blazars}
We also test our method with a little bit more complicated classification compositions in the data. 
Here we introduce two more types of rare objects: at spectrum radio quasars (FSRQ) and BL Lacertae objects (BL Lac).
The data set for testing star/galaxy/AGN/FSRQ/BL Lac classification problem starts with the same SDSSdr16 \code{SpecObj} table cross-matched to AllWISE with a $3\arcsec$ maximum radius and the SDSSdr16 \code{PhotoObj} table view by object ID, but not 2MASS. 
It, therefore, has a similar biased starting point, but the altered additional selection criteria imposed here produces a different balance of sources.

To identify blazars we cross-matched this table to the combination of two separate blazar catalogs: the fifth edition of Roma-BZCAT \citep[BZCAT][]{Massaro:2015} and the fourth version of the Fermi AGN catalog \citep[4LAC,][]{Fermi:2019_4lac}.
To combine 4LAC and BZCAT, we cross-matched the BZCAT coordinates to the coordinates of the radio companion in the 4LAC catalog with a $10\arcsec$ distance to eliminate duplicates (also dropping the 4LAC source 4FGL J1956.1+0234, because it was a worse match to its companion than the alternative).
Whenever the source classification disagreed between the catalogs, we used the one from 4LAC, because it is newer, unless the 4LAC class was \code{bcu} (blazar candidate unknown type), then the BZCAT classification was used.
The only source identifications used were BL Lacs or FSRQs.
We assigned BZCAT sources to classes based on the beginning of its \code{Source classification} string of characters.
We assigned the source to the BL Lac class if its string starts with ``BL Lac" and FSRQ when the string starts with ``QSO RLoud". 
For 4LAC the class strings were: ``bll" and ``fsrq", respectively, regardless of letter case.

Any \code{SpecObj} source that is within $5\arcsec$ of a Rmoa-BZCAT source's coordinates or a 4LAC source's companion gets labeled as the merged catalog's blazar class, regardless of the SDSS class given.

Similar to Section~\ref{sec:dat:SGQ}, we imposed the following constraints: $\code{zwarning}\in\{0,16\}$, $\code{w?cc\_map}=0$, and $\code{w?flux} > 5\times \code{w?sigflux}$ (after replacing bad \code{w1sigflux} values, as in Section~\ref{sec:dat:SGQ}), where the ``\code{?}" symbol takes each of \code{1}, \code{2}, and \code{3} to correspond with \textit{WISE}'s three shortest wavelength channels.

To constrain the SDSS photometry, first we selected fluxes appropriate to the morphological classification in the \code{PhotoObj} table.
For sources that have $\code{type}=6$ (consistent with the point spread function [psf]) in the \code{PhotoObj} table we used the values in the \code{psfFlux\_?} and \code{psfFluxIvar\_?} columns to construct the source's flux and uncertainty in the $g$, $r$, and $i$ filters (``\code{?}" = \code{g}, \code{r}, and \code{i}, respectively); for all other sources we used the \code{modelFlux\_?} and \code{modelFluxIvar\_?} columns.
Next, we used most of the recommended cuts in the documentation describing how the \code{PhotoObj} table's \code{clean} column was populated.\footnote{\url{https://www.sdss.org/dr16/algorithms/photo\_flags\_recommend/}}
Where we differ from the recommended usage is: with two exceptions, we required $g$, $r$, and $i$ to pass all cuts individually; we ignored \code{DEBLEND\_NOPEAK} because our SNR cuts made it irrelevant; and the \code{BINNED1} flag only had to be set in one of the three filters. 
Like the \textit{WISE} filters, the SNR in $g$, $r$, and $i$ all had to exceed 5.
After quality cuts, all SDSS fluxes were dereddened using the appropriate value from the \code{extinction\_?} column from \code{PhotoObj}.

Finally, \textit{WISE} has a lower resolution than SDSS (about $6\arcsec$ versus $1.3\arcsec$), which can make flux contamination from coincident sources a problem. 
Because the goal of this work is to test the modeling process, and not publish the model or a catalog from applying the model, we made no effort to reduce the impact of contamination (especially since it makes the factors tested more relevant).
The basic statistics of the selected data can be found in Table~\ref{tbl:blzdat}.
The reason that stars and AGN switch rarity rank in this data set is because the W3 SNR cut is heavily biased against most stars.

\begin{table}[htb]
\begin{center}
\caption{Blazar Data Characteristics}\label{tbl:blzdat}
\begin{tabular}{| c | lrrr |}
\hline\hline
Class & fraction & $N_{\mathrm{train}}\vphantom{N}^\mathrm{a}$ & $N_{\mathrm{dev}}\vphantom{N}^\mathrm{b}$ & $N_{\mathrm{test}}\vphantom{N}^\mathrm{c}$  \\
\hline
Galaxy & $0.6991$ & 270,988 & 33,874 & 33,873 \\
Star & $0.0081$ & 3,125 & 391 & 390 \\
AGN & $0.2913$ & 112,904 & 14,113 & 14,113 \\
BL Lac & $0.0005$ & 199 & 25 & 24 \\
FSRQ & $0.0010$ & 372 & 46 & 46 \\
\hline
total & 1 & 934,243 & 116,780 & 116,780 \\
\hline
\end{tabular} \\
\end{center}
$^\mathrm{a}$ Number of sources in the training set.\\
$^\mathrm{b}$ Number of sources in the development set.\\
$^\mathrm{c}$ Number of sources in the testing set.

\end{table}

\begin{figure*}[p]
	\begin{center}
	\includegraphics[width=\textwidth]{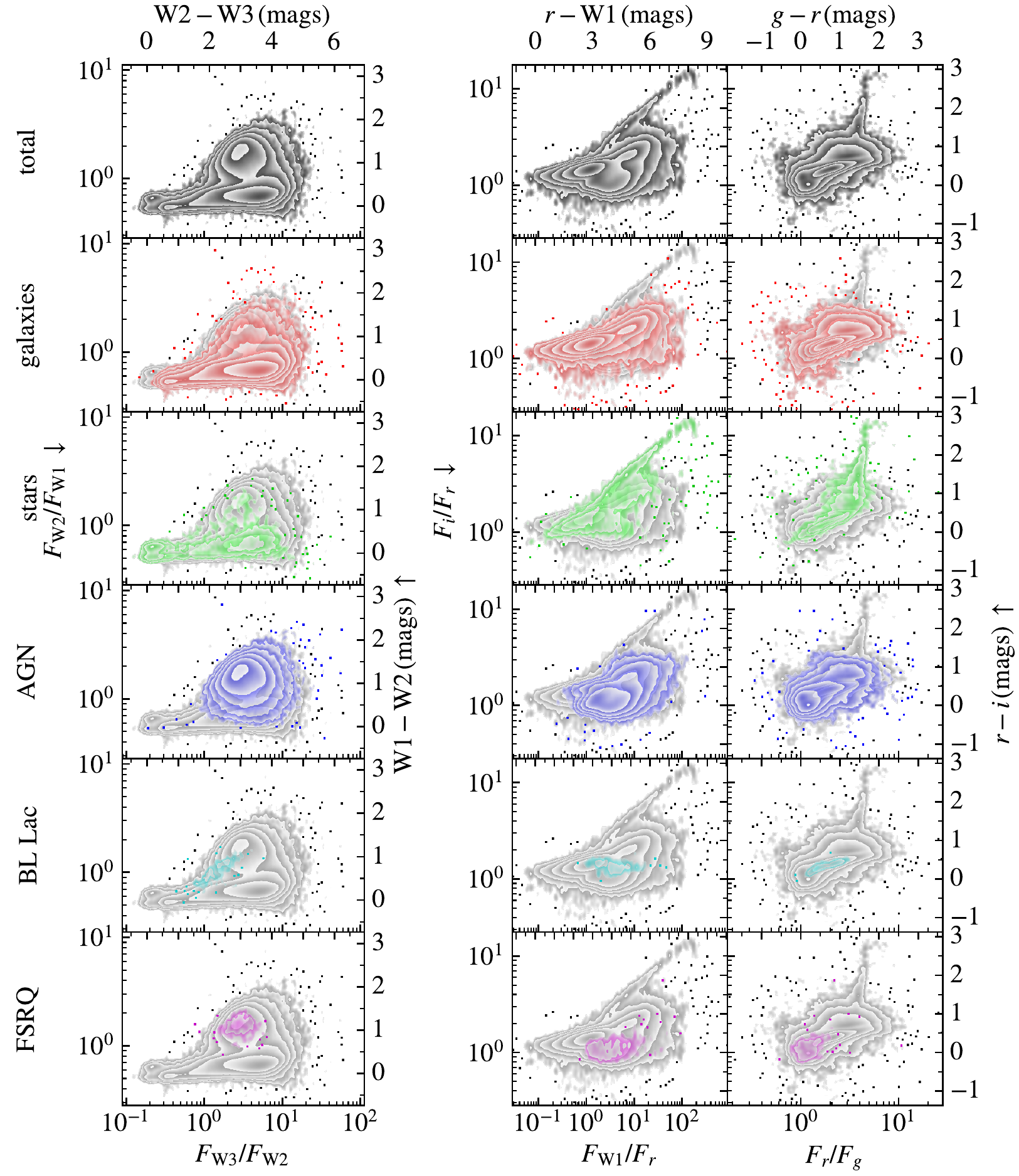}
	\end{center}
	
	\caption{\textit{WISE}-SDSS color--color diagrams}
	{The distribution in several color sub-spaces of all sources in the SDSS data release 16 \code{SpecObj} table view matched to AllWISE (details in the text), with each column smoothed using a 15, 25, and $15\operatorname{milli-dex}$ Gaussian kernel in each direction where individual points are not plotted.
	While just about any column provides good separation among stars, galaxies, and AGN, no single column does well on separating the FSRQ from AGN.
	\textit{WISE} filters are in the Vega magnitude system, and the SDSS ones are in the AB system. }
	\label{fig:WScolors}
\end{figure*}

Using the \code{w?flux} and \code{modelFlux\_?} or \code{psfFlux\_?} columns to construct magnitudes, we used the following colors as the base features for the neural network: $\mathrm{W}1 - \mathrm{W}2$, $\mathrm{W}2 - \mathrm{W}3$, $r-\mathrm{W}1$, $r-i$, and $g-r$.
Figure~\ref{fig:WScolors} contains a selection of density plots of the resulting data.
The left column plots $\mathrm{W}1 - \mathrm{W}2$ versus $\mathrm{W}2 - \mathrm{W}3$ and follows the bubble plot from \citet{Wright:2010}: main sequence stars are at the origin, most galaxies stretch in a horizontal line right of the origin, and the AGN lift above the galaxies in a bubble.
Note that this color space is the one that shows the largest difference between the blazar data (bottom two rows) and the other AGN (the row above the blazar ones).
The middle column plots $r-i$ versus $r - \mathrm{W}1$, following a similar pattern to color--color plot from \citet{Lake:2012}; most stars are on the stellar locus along the upper left edge of the data, and galaxies and AGN are in the less linear clusters of data below it with different centers.
The right column plots $r-i$ versus $g - r$ and shows a nearly horizontal linear trough in the data around $r-i\approx0.7$, which is close to some of the selection criteria for the SDSS Luminous Red Galaxy sample \citep[c.f. Figure~4 of][]{Eisenstein:2001}.

Figure~\ref{fig:WScolors} presents the clearest example of the impact of source contamination in the first column of the stars row.
First, the W3 SNR and quality cuts eliminated the vast majority of sources SDSS spectroscopically labeled as stars, leaving 1.16\% of stars that passed all other cuts for a total of 3,926 of them.
Of those that remain, 2,547 (65\%) are unusually red in $\mathrm{W}2-\mathrm{W}3$ ($\mathrm{W}2 - \mathrm{W}3 > 1\operatorname{mag}$).
The brightest neighbor in $r$ within $15\arcsec$ of the red stars is labeled as a galaxy and closer than $6\arcsec$ in SDSS's \code{PhotoPrimary} table for 831 (21\%) of them (1,215, $31\%$, for those when the brightest neighbor is closer than $10\arcsec$).
823 (21\%) have no SDSS neighbors within $15\arcsec$.
We have not explored deeply enough to make definitive statements, but it appears that anywhere from 1/2 to 2/3 are contamination from galaxies, AGN, or even some Galactic dust clouds, and the remainder are real infrared excess stars.

As the plots in Figure~\ref{fig:WScolors} show, in any given color--color plot the AGN and blazar data are nearly identical.
The primary difference seems to be the orientation of the tail of BL Lac sources (about one third of the blazars) that projects out down and to the left from the main blazar blob in the left column of plots in Figure~\ref{fig:WScolors}.

The network architecture from Section~\ref{sec:dat:SGQ} appears to have had enough excess capacity to handle the two additional classes and three features.
Because the number of basic features of 5, we had to use mathematician's convention for augmenting the features to avoid having too many free parameters of little use.
The number of parameters in this model is: $\left[\frac{1}{2} 5(5+3)\times 128 + 128\right] + [ 128 \times 32 + 32] + [32 \times 3 + 3] = 6,915$.
This is a decrease in the number of parameters from the SDSS model, but the increase in relevant feature count improves the separation of the data, making classification easier.
Indeed, with this much data the star/galaxy/AGN question can be answered in most cases, leaving most of the uncertainty for the AGN/blazar distinction.

\subsection{Literature Classification Catalog}\label{sec:dat:Clarke}
A catalog of classification probabilities for sources in the cross matched AllWISE and SDSS catalogs was produced in \citet{Clarke:2020}.
The machine learning algorithms used were based on the SDSS spectroscopic catalog, weighted to balance the classes. 
The catalog contains 111,293,033 sources, so it makes a good stress test of the half-EM algorithm, even if guaranteeing the accuracy of the new PCP would require a deeper analysis of the data that is beyond the scope of this work. 
We did not generate classifications for this data set, so no neural network was used.
We only use this data set to test the performance of the PCP algorithm derived from Equation~\ref{eqn:Pconsist}.

\section{Results}\label{sec:res}
For each of the data sets described in Sections~\ref{sec:dat:sim}--\ref{sec:dat:blazars} we have computed a number of performance metrics as a function of training set size.
For the simulated data, we graph five different cases from three models fit to data and corrections applied to the models that we are showing to be biased.
For the real data we only graph three cases.
The reason for the deeper exploration of the simulated data is because we could simulate data sets with balanced composition without discarding data, which would sacrifice most of the examples from the training set, making the training sets non-comparable by size.

Subsection~\ref{sec:res:sim} contains the results from the simulated data sets, showing the performance overshoot as a function of training set size, mean Kullback--Leibler divergence between the true probabilities and model probabilities by class, and a demonstration of the accuracy of the PCP computed using the half-EM algorithm.
Subsection~\ref{sec:res:SGQ} contains the results from the star-galaxy-AGN task when using \textit{WISE} and 2MASS data.
The only comparison here is the performance overshoot as a function of data set size, because we lacked reliable true values that each probability should take. 
Subsection~\ref{sec:res:blazars} contains an examination of the same performance overshoot as a function of data set size for the star-galaxy-AGN-FSRQ-BL Lac classification task using \textit{WISE} and SDSS data.
Finally, Subsection~\ref{sec:res:pcp} contains an examination of the performance and accuracy of the PCP produced by applying the half-EM algorithm to the data from \citet{Clarke:2020}.

\subsection{Simulation Results}\label{sec:res:sim}
\begin{figure*}[p]
	\begin{center}
	\includegraphics[width=\textwidth]{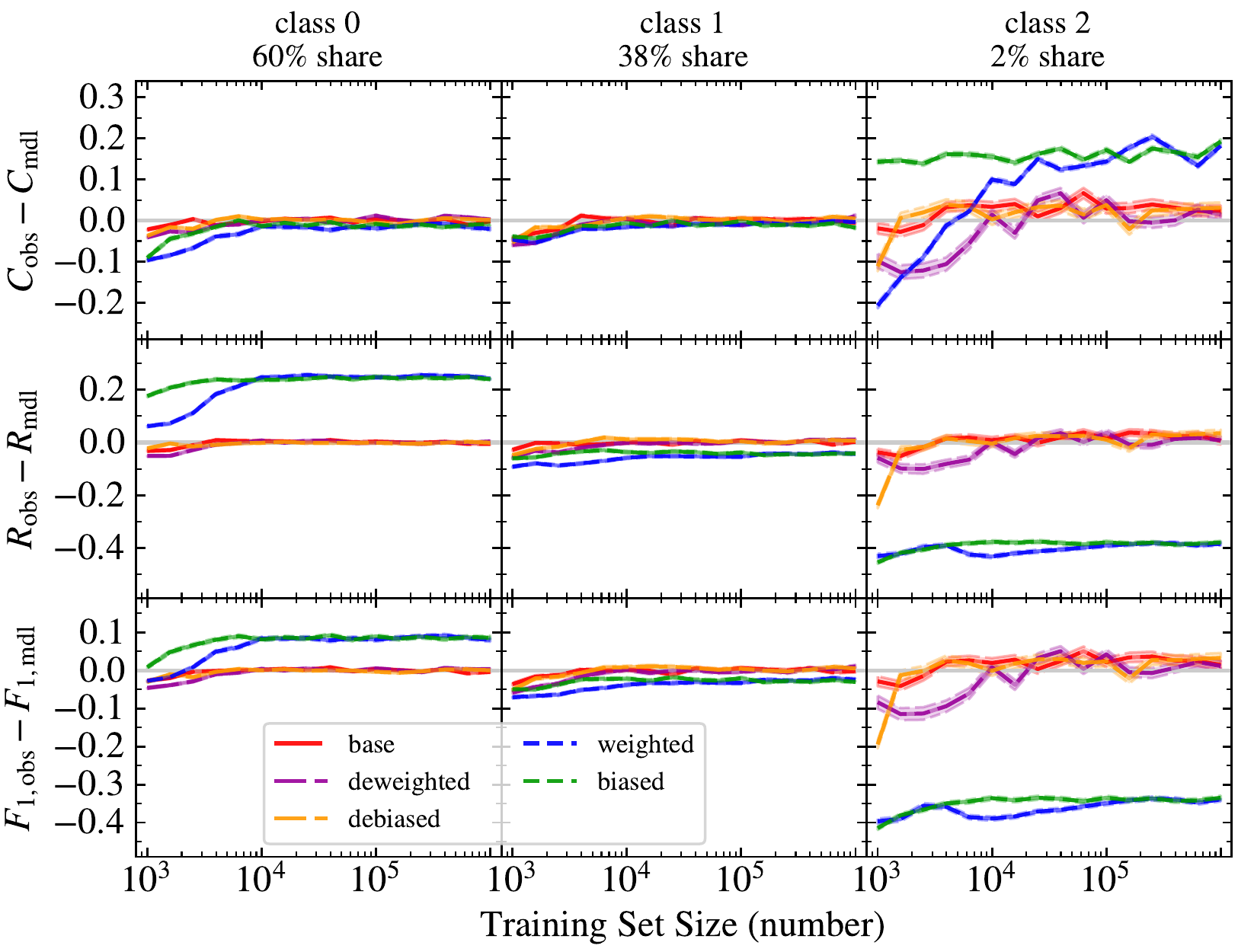}
	\end{center}
	
	\caption{Simulated Data Overshoot of Standard Performance Metrics}{ 
		Completeness, reliability, and $\text{F}_1$-score overshoot trends as a function of training set size, with the each column containing data for classes 0 through 2, respectively.
		The right hand column is class 2, and it is the rare class most strongly affected by weighting.
		All lines have a 1-$\sigma$ variance band around them, even if it is not visible.
		Notice how in all cases the deweighted and debiased lines follow the base lines, hitting their target values, and the weighted and biased ones do not.}
	\label{fig:simCRF1ovsh}
\end{figure*}

\begin{figure*}[htb]
	\begin{center}
	\includegraphics[width=\textwidth]{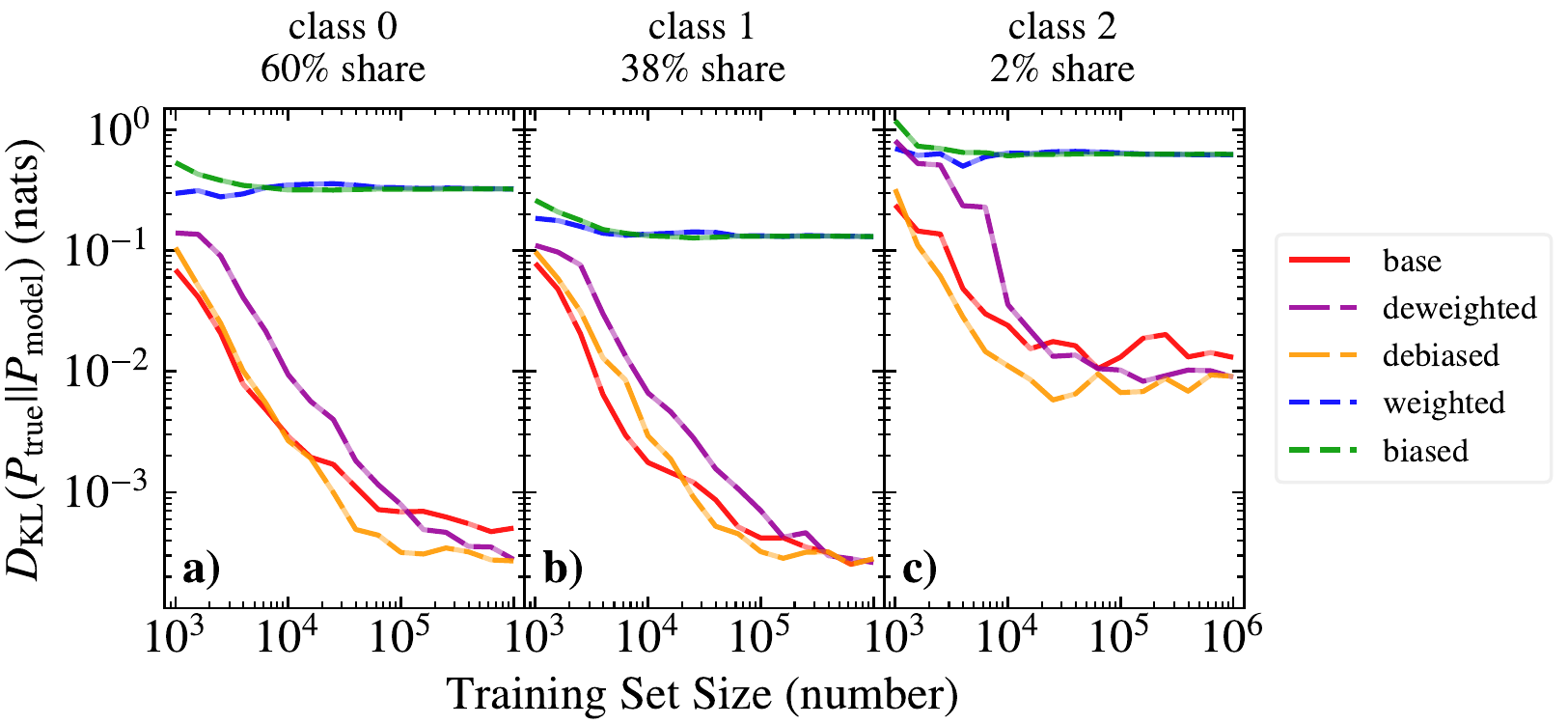}
	\end{center}
	
	\caption{Kullback--Leibler Divergence Trends}{ 
		Mean Kullback--Leibler divergence as a function of training set size, a measure of the amount of information lost when $P_{\mathrm{model}}$ is used to approximate the true probabilities $P_{\mathrm{true}}$.
		Note how the fits to weighted data sets and biased composition data sets asymptote to the same level for large sets, and similarly for the other three.
		Note also how the base fit outperforms all others for the common classes, and the debiased fit outperforms all others for the rare class.
		Panel (\textbf{a}) is class 0, (\textbf{b}) is class 1, and (\textbf{c}) is class 2. }
	\label{fig:simDKL}
\end{figure*}

We demonstrate training set balance bias and whether deweighting is an effective method to compensate for it using two techniques.
First, Figure~\ref{fig:simCRF1ovsh} shows each model's performance overshoot in the standard machine learning algorithm performance metrics as a function of training set size.
Second, Figure~\ref{fig:simDKL} shows a measure of distance between the various machine learning algorithm model probabilities and the analytic probability as a function of training set size.
Finally, we also did a short assessment of the half-EM method for constructing a PCP.

Figure~\ref{fig:simCRF1ovsh} contains the overshoot in the performance metrics on a test data set for: when the machine learning algorithm's model is applied to a representative data set without weighting (``base"), when the data was weighted to simulate each class being equally likely (``weighted"), when the probabilities from the weighted fit model were deweighted according to Equation~\ref{eqn:deweight} (``deweighted"), when the training data were of biased equal composition (``biased"), and when the probabilities from the biased fit model were debiased using the same process as the deweighted ones (``debiased").
Each column in Figure~\ref{fig:simCRF1ovsh} is dedicated to one source class (0, 1, and 2, respectively).
Each line has a shaded 1-$\sigma$ band around it, but most are not visible.

Figure~\ref{fig:simCRF1ovsh} shows that the deweighted and debiased models mostly follow the behavior of the base model, and the model fit to the biased training set behaves similarly to the weighted fit one.
It also shows that the weighted and biased fits' performance metrics indicate inaccurate probabilities in all metrics. 
They also show, specifically, that the weighted and biased models applied to the rare class (2, right hand column) are over-complete and under-reliable, as predicted.

%

%

The comparisons in Figure~\ref{fig:simCRF1ovsh} are not particularly sensitive measures of how accurately the models are computing the probabilities at each data point: a lot of information is discarded when classes are assigned. 
Having the probabilities that the models should produce in hand permits us to compute the distance between the model probabilities and true ones.
For the distance metric we selected the Kullback--Leibler divergence
\begin{align}
	D_{\mathrm{KL}}(P_{\mathrm{true}}||P_{\mathrm{model}}) & = \sum_{i=0}^2 P_{\mathrm{true}}(i|x_i) \ln\left(\frac{P_{\mathrm{true}}(i|x_i)}{P_{\mathrm{model}}(i|x_i)}\right),
\end{align}
averaged over all of the examples of each class, in nats ($\ln 2\operatorname{nats} = 1\operatorname{bit}$).
The two main reasons for choosing $D_{\mathrm{KL}}$ are that it has an interpretation as the ``information lost when $P_{\mathrm{model}}$ is used to approximate $P_{\mathrm{true}}$" \citep[p. 51]{Burnham:2002} and, under the name ``information gain," it is the most commonly used quantity when constructing decision trees \cite[Section 3.6.2.1]{Louppe:2014}.

The trends for the sample mean $D_{\mathrm{KL}}$ evaluated on the test set, split by class 0--2 in panels (\textbf{a})--(\textbf{c}), respectively, can be found in Figure~\ref{fig:simDKL}. 
For small training set sizes ($N < 10^4$) the biased and debiased lines mirror each-other, and similarly for the weighted and deweighted lines.
Between $10^4$ and $10^5$ the deweighted and debiased lines split to join the base (unbiased/unweighted) line. 
For sets larger than about $10^5$ the lines begin to approach horizontal asymptotes, representing the same residual bias in the models evident in Figure~\ref{fig:simCRF1ovsh}. 
Crucially, the biased and weighted models asymptote to the same value, and the debiased and deweighted models asymptote to the same lower value as the base model.

Also, note how the debiased lines outperform or match the base lines for most training set sizes, indeed for all training set sizes for class 2.

Finally, we can assess the performance of the half-EM algorithm on the simulated data. 
From Table~\ref{tbl:simfracs} we can see that the true prior used to simulate the data is 60, 38, and 2\%.
Applying the half-EM algorithm to the probabilities computed on the testing data yielded:
\begin{itemize}
  \item 60.6, 37.6, and 1.8\% for the weighted model,
  \item 60.9, 37.2, and 1.9\% for the biased set model, and
  \item 60.4, 37.5, and 2.1\% for the base model.
\end{itemize}
Given that the testing set size was 20,000 points, these results within the expected statistical variance $\sim1\%$.

\subsection{Star/Galaxy/AGN Results}\label{sec:res:SGQ}
\begin{figure*}[p]
	\begin{center}
	\includegraphics[width=\textwidth]{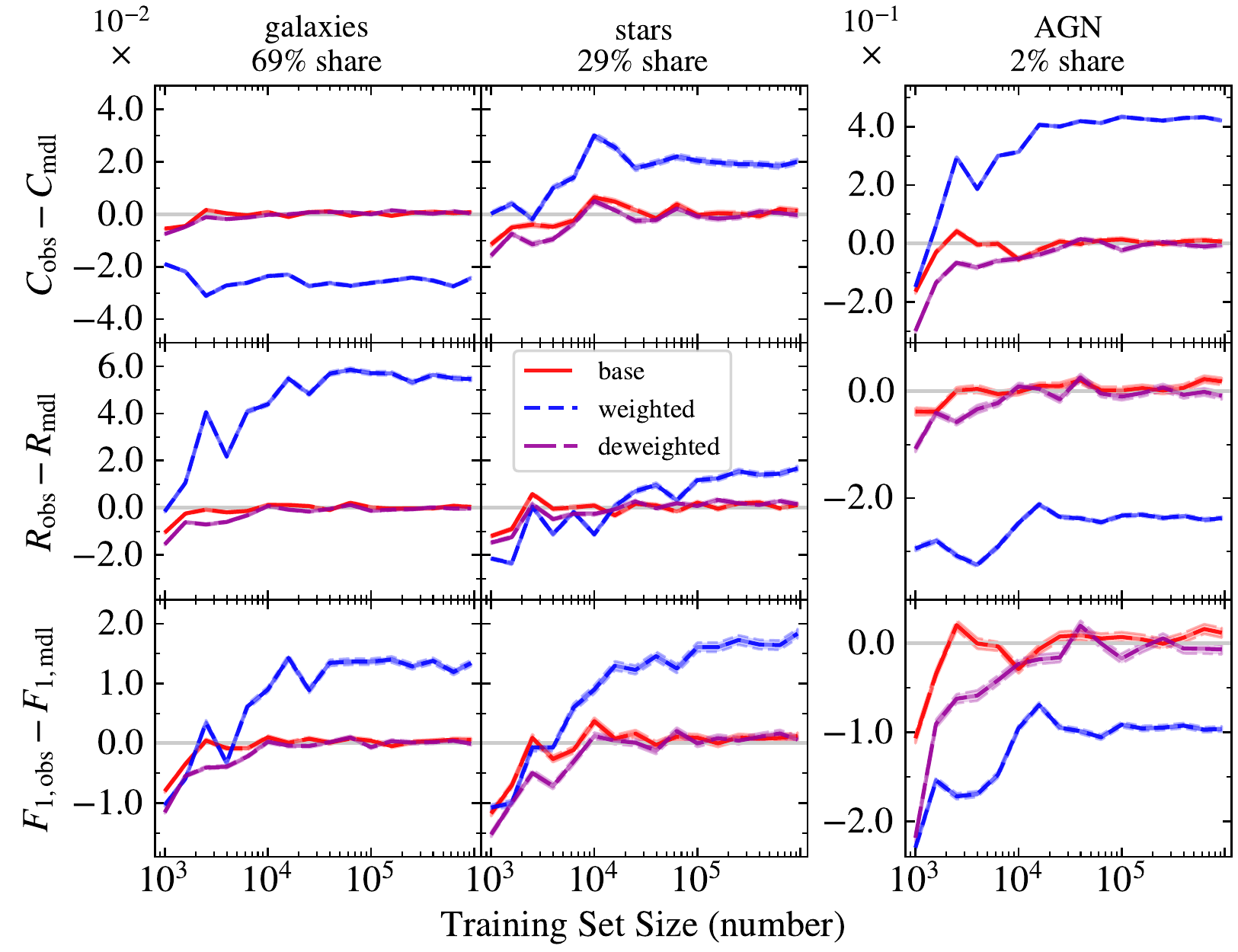}
	\end{center}
	
	\caption{Overshoot of Standard Performance Metrics}{
		Completeness, reliability, and $\text{F}_1$-score overshoot trends as a function of training set size to SDSS-selected sources. 
		The right hand column is AGN, and it is the rare class most strongly affected by weighting.
		All lines have a 1-$\sigma$ variance band around them, even if it is not visible.
		Notice how the deweighted line continues to follow the base one and the ideal one for real data, similar to simulated data. }
	\label{fig:sgqCRF1ovsh}
\end{figure*}

Lacking the ability to calculate probabilities analytically, the performance metrics applied to real data are less detailed than the ones available for simulated data.
Figure~\ref{fig:sgqCRF1ovsh} contains the same sort of comparisons between weighted and unweighted fits as Figure~\ref{fig:simCRF1ovsh}, with galaxies in the left column, stars in the middle, and AGN in the right.

The main difference between Figures~\ref{fig:simCRF1ovsh} and \ref{fig:sgqCRF1ovsh} is that the models the machine learning algorithm produced do such a good job on the star/galaxy split that the $y$-axes has to be zoomed in to make the details of the lines visible. 
In this case the base and deweighted models manage to outperform the weighted ones almost everywhere, with the rare weighted class being over-complete and under-reliable.
The lines also show evidence of residual bias, since they do not asymptote to zero.



%

\subsection{Blazars}\label{sec:res:blazars}
\begin{figure*}[p]
	\begin{center}
	\includegraphics[width=\textwidth]{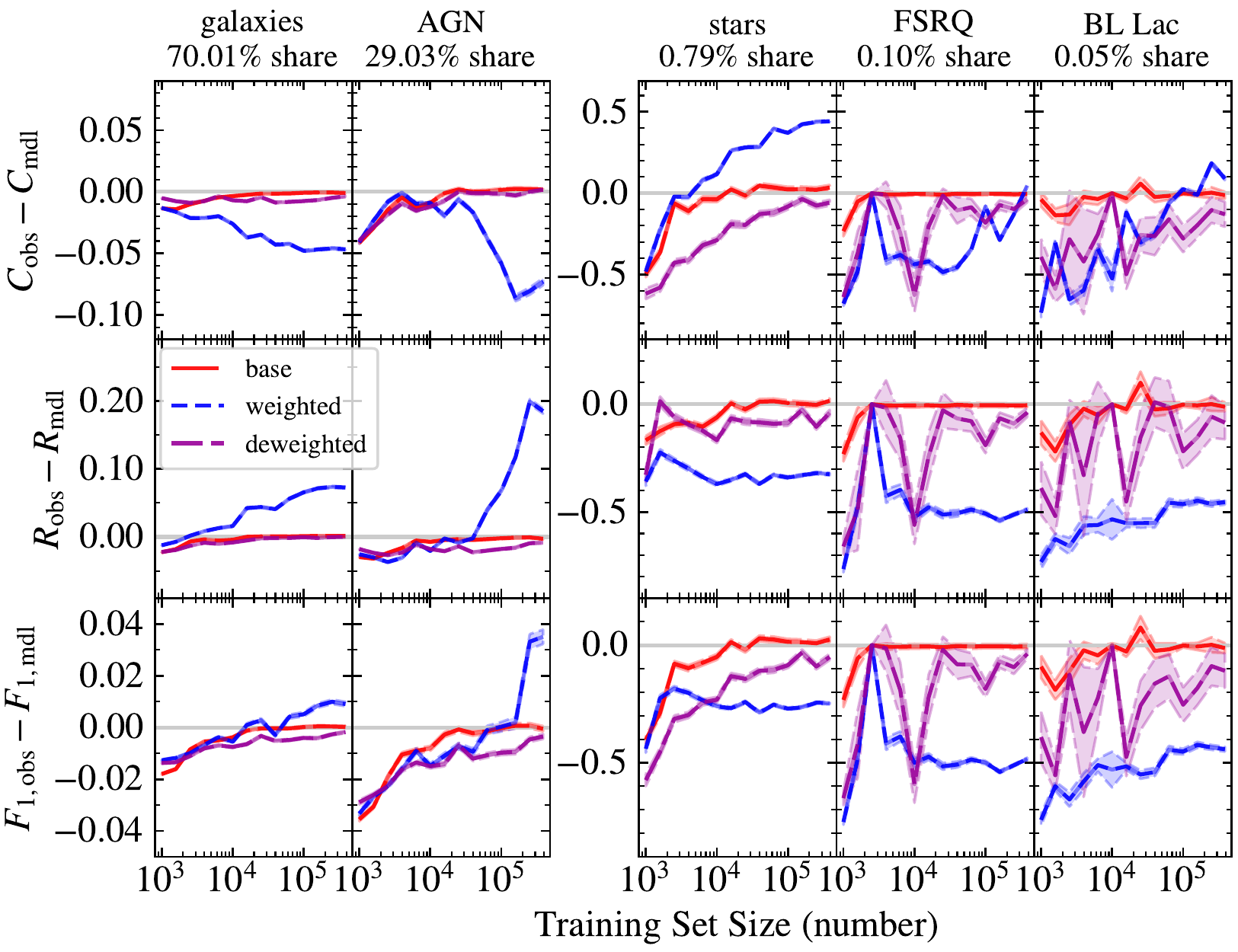}
	\end{center}
	
	\caption{Blazar Overshoot of Standard Performance Metrics}{
		Completeness, reliability, and $\text{F}_1$-score overshoot trends as a function of training set size to SDSS-selected sources with FSRQ and BL Lacs identified. 
		All lines have a 1-$\sigma$ variance band around them, even if it is not visible.
		The abundance in the training data of each column is listed at the top.
		Notice how the Corrected line continues to follow the Unweighted one for real data, similar to simulated data, but does not follow as closely, probably due to overfitting. }
	\label{fig:blazCRF1ovsh}
\end{figure*}

Figure~\ref{fig:blazCRF1ovsh} is similar to Figure~\ref{fig:sgqCRF1ovsh}, with columns reordered to reflect the altered abundance of source types in this differently selected data set. 
The dominant trends remain the same: the rare classes are over-complete and under-reliable in the weighted models, at the expense of the common classes which become under-complete and over-reliable as sources are moved from the latter to the former by the weighting.

The BL Lac column in Figure~\ref{fig:blazCRF1ovsh} may also show evidence of higher order effects of overweighting. 
Their extreme rarity means they get a weight that is extremely high, and their location in an otherwise under-dense region of the color-space means that the model may be treating the BL Lac data as a set of extremely sharp isolated peaks, instead of a unified cloud.
Up-weighting such a small minority of the data appears to have induced the machine learning algorithm to over-fit the data.
This seems to be the most likely explanation for why the deweighted BL Lacs are under-complete and over-reliable. 
The FSRQ, on the other hand, sit right over the highest density region of the AGN data, making it more difficult for isolated data peaks to dominate the fit locally, in spite of their similar rarity.

\subsection{Producing a Prediction Consistent Prior}\label{sec:res:pcp}
The catalog published in \citet{Clarke:2020} contains 111,293,033 sources, making it a good stress test for the half-EM algorithm.
The catalog was generated by applying a model to deeper data than the training set, so the results of this exercise are indicative, not definitive at the level of describing reality.
Averaging the probabilities published in the catalog produces the prior probabilities 41.8, 53.5, and 4.6\% for galaxies, stars, and quasars (QSOs), respectively, in the cross-matched SDSS-AllWISE photometric catalog.
This catalog is based on a model that weighted the training set to balance the classes, suggesting that the QSO count is likely over-estimated.
Assuming the model uses a flat prior produces a prediction consistent prior of 44.6, 55.0, and 0.4\% galaxies, stars, and QSOs.
This quasar fraction is a good deal lower than the 1--4\% broad-line AGN found in the unbiased \textit{WISE}/DEIMOS survey described in \citet{Lake:2012} and \citet{Lake:2018}, and is, therefore, unlikely to be accurate.
Using a prior based on the training set source balance, 72.9, 15.0, and 12.1\%, produces a prediction consistent prior of 37.6, 61.0, and 1.3\%, which is closer to the spectroscopic survey results.

Converging the above priors to an accuracy of $10^{-8}$, as measured by the mean absolute deviation, required 15 to 28 iterations at 6.3 to 8.5 seconds per iteration on the same laptop all other code in this paper was run on (reading in the data set from disk took longer than iterating).
This was done using the slower recursive algorithm.

In summary, the stark difference between either of the possible model priors and the prior based on the prediction is a strong indication that the prior used is inaccurate.
The question of which PCP is accurate is likely irrelevant in this case, because the selection function used to target the spectra almost certainly gave the PDFs of the training set data different shapes from the PDFs of the photometric data, requiring a more extensive semi-supervised learning algorithm to compensate.

\section{Discussion}\label{sec:disc}
Implicit in the arguments made in this paper is the view that classification decisions should be driven by accurate empirically observable probabilities.
One reason why is because it gives the end user the ability to accurately make his or her own decisions about where the dividing line should be based on her or his own completeness and reliability needs.
Another is that it permits the end user to make classification decisions based on well established statistical techniques, like expected value, that have inputs that vary from person to person.

We would go further, and claim that all classification schemes can be tied back to probability in some way, even if its only considering the classifications, themselves, as approximating the probability vectors a 1 in the assigned class and 0 in the others.
This intimate link with what are called ``posterior probabilities" implies that it is impossible to avoid setting a prior probability that the evidence can modify to produce the final prediction probabilities.
In machine learning, the question of how to set the prior is usually left to be implicitly set by the data set fed to the algorithm.

While the effect of weighting on the decision boundary positions is not explicitly mentioned in many sources, it can also be the purpose of applying up-weighting.\footnote{This is done, for example, in the ``Weighting of Examples for Classification" section of the documentation for the Wolfram Language, \url{https://reference.wolfram.com/language/tutorial/NeuralNetworksExampleWeighting.html}.}
In our experiments, weighting the data in this fashion succeeded in focusing the machine learning algorithm on the rare class, but this focus also reduced the overall performance of the classifier.
It is as though up-weighting parts of the data reduces the effective size of the data set.

It is in this context that the practice of weighting training data needs to be considered. 
Weighting the data modifies this implicitly set prior by mimicking a data set with a different composition.
Modifying the prior can be a valid technique.
For example, if the available training data is non-representative of the real data then modifying the prior can increase the accuracy of the model.
The question, then, is whether or not weighting the training data is the best way to modify the prior? 
The evidence and arguments presented here support the idea that the predictions are likely to be more accurate if the data is fit without weighting, and any adjustments to the prior are made after the model computes its probabilities.
Given that, it seems the way to get the most out of a machine learning algorithm classifier is to:
\begin{enumerate}
  \item be greedy about including data (especially from the rare classes), as long as any biases induced are known,
  \item avoid weighting data, if possible, and
  \item compensate for biases in the predictions by deweighting using Equation~\ref{eqn:deweight}.
\end{enumerate}
One example of a case where weighting cannot be avoided would be if the completeness (selection probability) cannot be computed based on output class and the data features.
In that case, it is not possible to deweight the machine learning algorithm model's prediction, so weighting the training data while fitting the machine learning algorithm is the only option for producing an accurate prior.

This advice cannot be regarded as definitive, though, because in modeling we did for this paper we did not actually observe one of the problems that weighting is supposed to solve: the machine learning algorithm ignoring the rare class outright.
We came close in preliminary work with the BL Lac and FSRQ classes when using the Bayes classifier (assign each object to whichever class has the highest probability).
In that case the problem likely originated from the fact that they were correctly being assigned low probabilities because there were no points in the feature space where the blazars outnumbered other classes, meaning no sources were assigned to those rare classes.
The problem was avoided here by using a stochastic classifier instead of the Bayes one. 
When the catalog of classifications is to be published, though, the higher expected error rate from using a non-Bayes classifier may be unacceptably high because the Bayes classifier is proven to be optimal in that metric \citep{Domingos:1997}.
If the observation generalizes that the point of weighting is to ensure sources are assigned to the rare class(es) in spite of their correctly calculated low probabilities, then the same effect can be achieved by weighting the machine learning algorithm's output probabilities, instead of the training data, without the apparent over-fitting discussed below.
This could be described as, effectively, using a different Bayesian prior for the probabilities before making classifications.
If the person generating the catalog is not the final user, though, then a better approach is to make soft classifications (i.e. publish the probabilities) so that the various end users can make their own classification decisions with their individual requirements for expected completeness and reliability class-by-class.

The other phenomenon we did not expect was the high variability of the deweighted overshoot curves for BL Lacs (right column of Figure~\ref{fig:blazCRF1ovsh}). 
We do not have a definitive explanation, for this.
One possible explanation is that the extreme up-weighting assigned to the BL Lacs combined with the location of the data in the W1--W2--W3 color space to induce over-fitting of this class. 
Inspection of Figure~\ref{fig:WScolors} reveals that FSRQ sit at same place as the peak of the AGN data. 
We believe that this is also nearly true for BL Lacs except in the W1--W2--W3 space, where the left hand tail of the BL-Lac data extends into a region where all other forms a data have low density.
This, with the extreme up-weighting applied to BL Lac data, meant that enough of the neural network's resources were dedicated to describing the BL Lac versus other class question that it resolved the individual training examples as delta functions, instead of smoothing them (i.e. the blazar data was over-fit). 
It is tempting to invoke ``adversarial examples" \citep{Wiyatno:2019}, but in a space this simple where there is no mismatch between training and testing data sets (i.e. the test set falls within the footprint of the training set), and the number of training examples (199 BL Lacs) greatly exceeds the dimensionality of the data (5) (i.e. they should fill the space) this seems less likely. 
Thus we feel it necessary to add the caveat that deweighting may not work when the weights used are too extreme.

The applications of the techniques in this paper are manifold, because finding rare but important occurrences in large samples is one of the primary use-cases for machine learning algorithms.
What is more, because these techniques address the accuracy of the prior probability, they can address many of the biases caused by letting imbalanced data set a biased prior, implicitly.

\section{Conclusion}\label{sec:conc}
We have shown that weighting the data to tune the performance of a classification machine learning algorithm should be done with care because it can modify the probabilities the model assigns in a way that mimics fitting the model on biased data.
The machine learning algorithm we used to show this was a small neural network implemented in Python.
We showed this by applying the neural network to: simulated data, source classification in a color--color plot, and source classification in a more extended color space.
We further observed that the bias from weighting or biased data sets can be removed by deweighting the probabilities the model assigns, and that the machine learning algorithm achieved its best performance when deweighting a fit to biased data, followed by unweighted data.
Finally, as long as the implicit prior probability is known, and the models are not otherwise biased, it is possible to construct an accurate prior probability from an unbiased and unlabeled data set using half of the EM algorithm.

In sum:
\begin{itemize}
	\item weighting data during training should be avoided because it biases results,
	\item if weighting or discarding data for balance cannot be avoided, then the bias induced can be corrected by deweighting the classifier's probabilities, and
	\item soft classification (publishing accurate probabilities) allows for more accurate demographic measurements than classifications, enabling both performance measurements and a determination of a prediction consistent prior on unlabeled data.
\end{itemize}

\section{Acknowledgements}
Funded by Chinese Academy of Sciences President’s International Fellowship Initiative. Grant No. 2019PM0017.

This project is partially supported by the CAS International Partnership Program No.114A11KYSB20160008 and CAS Interdisciplinary Innovation Team.

C.-W. Tsai was supported by a grant from the NSFC (No. 11973051).

This work is also supported by NSFC grant No. U1931110.

Funding for the Sloan Digital Sky Survey IV has been provided by the Alfred P. Sloan Foundation, the U.S. Department of Energy Office of Science, and the Participating Institutions. SDSS-IV acknowledges
support and resources from the Center for High-Performance Computing at
the University of Utah. The SDSS web site is www.sdss.org.

SDSS-IV is managed by the Astrophysical Research Consortium for the 
Participating Institutions of the SDSS Collaboration including the 
Brazilian Participation Group, the Carnegie Institution for Science, 
Carnegie Mellon University, the Chilean Participation Group, the French Participation Group, Harvard-Smithsonian Center for Astrophysics, 
Instituto de Astrof\'isica de Canarias, The Johns Hopkins University, Kavli Institute for the Physics and Mathematics of the Universe (IPMU) / 
University of Tokyo, the Korean Participation Group, Lawrence Berkeley National Laboratory, 
Leibniz Institut f\"ur Astrophysik Potsdam (AIP),  
Max-Planck-Institut f\"ur Astronomie (MPIA Heidelberg), 
Max-Planck-Institut f\"ur Astrophysik (MPA Garching), 
Max-Planck-Institut f\"ur Extraterrestrische Physik (MPE), 
National Astronomical Observatories of China, New Mexico State University, 
New York University, University of Notre Dame, 
Observat\'ario Nacional / MCTI, The Ohio State University, 
Pennsylvania State University, Shanghai Astronomical Observatory, 
United Kingdom Participation Group,
Universidad Nacional Aut\'onoma de M\'exico, University of Arizona, 
University of Colorado Boulder, University of Oxford, University of Portsmouth, 
University of Utah, University of Virginia, University of Washington, University of Wisconsin, 
Vanderbilt University, and Yale University.
This research has made use of the NASA/IPAC Infrared Science Archive, which is funded by the National Aeronautics and Space Administration and operated by the California Institute of Technology.
This publication makes use of data products from the Wide-field Infrared Survey Explorer, which is a joint project of the University of California, Los Angeles, and the Jet Propulsion Laboratory/California Institute of Technology, funded by the National Aeronautics and Space Administration.

This publication makes use of data products from the Wide-field Infrared Survey Explorer, which is a joint project of the University of California, Los Angeles, and the Jet Propulsion Laboratory/California Institute of Technology, and NEOWISE, which is a project of the Jet Propulsion Laboratory/California Institute of Technology. WISE and NEOWISE are funded by the National Aeronautics and Space Administration.

This publication makes use of data products from the Two Micron All Sky Survey, which is a joint project of the University of Massachusetts and the Infrared Processing and Analysis Center/California Institute of Technology, funded by the National Aeronautics and Space Administration and the National Science Foundation.


\bibliography{MLbiasBib}

\begin{thebibliography}{56}
\expandafter\ifx\csname natexlab\endcsname\relax\def\natexlab#1{#1}\fi
\providecommand{\url}[1]{\texttt{#1}}
\providecommand{\href}[2]{#2}
\providecommand{\path}[1]{#1}
\providecommand{\DOIprefix}{doi:}
\providecommand{\ArXivprefix}{arXiv:}
\providecommand{\URLprefix}{URL: }
\providecommand{\Pubmedprefix}{pmid:}
\providecommand{\doi}[1]{\href{http://dx.doi.org/#1}{\path{#1}}}
\providecommand{\Pubmed}[1]{\href{pmid:#1}{\path{#1}}}
\providecommand{\bibinfo}[2]{#2}
\ifx\xfnm\relax \def\xfnm[#1]{\unskip,\space#1}\fi
\bibitem[{{Ahumada} et~al.(2020){Ahumada}, {Prieto}, {Almeida}, {Anders},
  {Anderson}, {Andrews}, {Anguiano}, {Arcodia}, {Armengaud}, {Aubert}, {Avila},
  {Avila-Reese}, {Badenes}, {Balland}, {Barger}, {Barrera-Ballesteros}, {Basu},
  {Bautista}, {Beaton}, {Beers}, {Benavides}, {Bender}, {Bernardi}, {Bershady},
  {Beutler}, {Bidin}, {Bird}, {Bizyaev}, {Blanc}, {Blanton}, {Boquien},
  {Borissova}, {Bovy}, {Brandt}, {Brinkmann}, {Brownstein}, {Bundy}, {Bureau},
  {Burgasser}, {Burtin}, {Cano-D{\'\i}az}, {Capasso}, {Cappellari}, {Carrera},
  {Chabanier}, {Chaplin}, {Chapman}, {Cherinka}, {Chiappini}, {Doohyun Choi},
  {Chojnowski}, {Chung}, {Clerc}, {Coffey}, {Comerford}, {Comparat}, {da
  Costa}, {Cousinou}, {Covey}, {Crane}, {Cunha}, {Ilha}, {Dai}, {Damsted},
  {Darling}, {Davidson}, {Davies}, {Dawson}, {De}, {de la Macorra}, {De Lee},
  {Queiroz}, {Deconto Machado}, {de la Torre}, {Dell'Agli}, {du Mas des
  Bourboux}, {Diamond-Stanic}, {Dillon}, {Donor}, {Drory}, {Duckworth},
  {Dwelly}, {Ebelke}, {Eftekharzadeh}, {Davis Eigenbrot}, {Elsworth},
  {Eracleous}, {Erfanianfar}, {Escoffier}, {Fan}, {Farr},
  {Fern{\'a}ndez-Trincado}, {Feuillet}, {Finoguenov}, {Fofie},
  {Fraser-McKelvie}, {Frinchaboy}, {Fromenteau}, {Fu}, {Galbany}, {Garcia},
  {Garc{\'\i}a-Hern{\'a}ndez}, {Oehmichen}, {Ge}, {Maia}, {Geisler}, {Gelfand},
  {Goddy}, {Gonzalez-Perez}, {Grabowski}, {Green}, {Grier}, {Guo}, {Guy},
  {Harding}, {Hasselquist}, {Hawken}, {Hayes}, {Hearty}, {Hekker}, {Hogg},
  {Holtzman}, {Horta}, {Hou}, {Hsieh}, {Huber}, {Hunt}, {Chitham}, {Imig},
  {Jaber}, {Angel}, {Johnson}, {Jones}, {J{\"o}nsson}, {Jullo}, {Kim},
  {Kinemuchi}, {Kirkpatrick}, {Kite}, {Klaene}, {Kneib}, {Kollmeier}, {Kong},
  {Kounkel}, {Krishnarao}, {Lacerna}, {Lan}, {Lane}, {Law}, {Le Goff}, {Leung},
  {Lewis}, {Li}, {Lian}, {Lin}, {Long}, {Longa-Pe{\~n}a}, {Lundgren}, {Lyke},
  {Ted Mackereth}, {MacLeod}, {Majewski}, {Manchado}, {Maraston}, {Martini},
  {Masseron}, {Masters}, {Mathur}, {McDermid}, {Merloni}, {Merrifield},
  {M{\'e}sz{\'a}ros}, {Miglio}, {Minniti}, {Minsley}, {Miyaji}, {Mohammad},
  {Mosser}, {Mueller}, {Muna}, {Mu{\~n}oz-Guti{\'e}rrez}, {Myers}, {Nadathur},
  {Nair}, {Nandra}, {do Nascimento}, {Nevin}, {Newman}, {Nidever}, {Nitschelm},
  {Noterdaeme}, {O'Connell}, {Olmstead}, {Oravetz}, {Oravetz}, {Osorio},
  {Pace}, {Padilla}, {Palanque-Delabrouille}, {Palicio}, {Pan}, {Pan},
  {Parker}, {Paviot}, {Peirani}, {Ram{\'r}ez}, {Penny}, {Percival},
  {Perez-Fournon}, {P{\'e}rez-R{\`a}fols}, {Petitjean}, {Pieri},
  {Pinsonneault}, {Poovelil}, {Povick}, {Prakash}, {Price-Whelan}, {Raddick},
  {Raichoor}, {Ray}, {Rembold}, {Rezaie}, {Riffel}, {Riffel}, {Rix}, {Robin},
  {Roman-Lopes}, {Rom{\'a}n-Z{\'u}{\~n}iga}, {Rose}, {Ross}, {Rossi},
  {Rowlands}, {Rubin}, {Salvato}, {S{\'a}nchez}, {S{\'a}nchez-Menguiano},
  {S{\'a}nchez-Gallego}, {Sayres}, {Schaefer}, {Schiavon}, {Schimoia},
  {Schlafly}, {Schlegel}, {Schneider}, {Schultheis}, {Schwope}, {Seo},
  {Serenelli}, {Shafieloo}, {Shamsi}, {Shao}, {Shen}, {Shetrone}, {Shirley},
  {Aguirre}, {Simon}, {Skrutskie}, {Slosar}, {Smethurst}, {Sobeck}, {Sodi},
  {Souto}, {Stark}, {Stassun}, {Steinmetz}, {Stello}, {Stermer},
  {Storchi-Bergmann}, {Streblyanska}, {Stringfellow}, {Stutz}, {Su{\'a}rez},
  {Sun}, {Taghizadeh-Popp}, {Talbot}, {Tayar}, {Thakar}, {Theriault}, {Thomas},
  {Thomas}, {Tinker}, {Tojeiro}, {Toledo}, {Tremonti}, {Troup}, {Tuttle},
  {Unda-Sanzana}, {Valentini}, {Vargas-Gonz{\'a}lez}, {Vargas-Maga{\~n}a},
  {V{\'a}zquez-Mata}, {Vivek}, {Wake}, {Wang}, {Weaver}, {Weijmans}, {Wild},
  {Wilson}, {Wilson}, {Wolthuis}, {Wood-Vasey}, {Yan}, {Yang}, {Y{\`e}che},
  {Zamora}, {Zarrouk}, {Zasowski}, {Zhang}, {Zhao}, {Zhao}, {Zheng}, {Zheng},
  {Zhu} and {Zou}}]{SDSSdr16}
\bibinfo{author}{{Ahumada}, R.}, \bibinfo{author}{{Prieto}, C.A.},
  \bibinfo{author}{{Almeida}, A.}, \bibinfo{author}{{Anders}, F.},
  \bibinfo{author}{{Anderson}, S.F.}, \bibinfo{author}{{Andrews}, B.H.},
  \bibinfo{author}{{Anguiano}, B.}, \bibinfo{author}{{Arcodia}, R.},
  \bibinfo{author}{{Armengaud}, E.}, \bibinfo{author}{{Aubert}, M.},
  \bibinfo{author}{{Avila}, S.}, \bibinfo{author}{{Avila-Reese}, V.},
  \bibinfo{author}{{Badenes}, C.}, \bibinfo{author}{{Balland}, C.},
  \bibinfo{author}{{Barger}, K.}, \bibinfo{author}{{Barrera-Ballesteros},
  J.K.}, \bibinfo{author}{{Basu}, S.}, \bibinfo{author}{{Bautista}, J.},
  \bibinfo{author}{{Beaton}, R.L.}, \bibinfo{author}{{Beers}, T.C.},
  \bibinfo{author}{{Benavides}, B.I.T.}, \bibinfo{author}{{Bender}, C.F.},
  \bibinfo{author}{{Bernardi}, M.}, \bibinfo{author}{{Bershady}, M.},
  \bibinfo{author}{{Beutler}, F.}, \bibinfo{author}{{Bidin}, C.M.},
  \bibinfo{author}{{Bird}, J.}, \bibinfo{author}{{Bizyaev}, D.},
  \bibinfo{author}{{Blanc}, G.A.}, \bibinfo{author}{{Blanton}, M.R.},
  \bibinfo{author}{{Boquien}, M.}, \bibinfo{author}{{Borissova}, J.},
  \bibinfo{author}{{Bovy}, J.}, \bibinfo{author}{{Brandt}, W.N.},
  \bibinfo{author}{{Brinkmann}, J.}, \bibinfo{author}{{Brownstein}, J.R.},
  \bibinfo{author}{{Bundy}, K.}, \bibinfo{author}{{Bureau}, M.},
  \bibinfo{author}{{Burgasser}, A.}, \bibinfo{author}{{Burtin}, E.},
  \bibinfo{author}{{Cano-D{\'\i}az}, M.}, \bibinfo{author}{{Capasso}, R.},
  \bibinfo{author}{{Cappellari}, M.}, \bibinfo{author}{{Carrera}, R.},
  \bibinfo{author}{{Chabanier}, S.}, \bibinfo{author}{{Chaplin}, W.},
  \bibinfo{author}{{Chapman}, M.}, \bibinfo{author}{{Cherinka}, B.},
  \bibinfo{author}{{Chiappini}, C.}, \bibinfo{author}{{Doohyun Choi}, P.},
  \bibinfo{author}{{Chojnowski}, S.D.}, \bibinfo{author}{{Chung}, H.},
  \bibinfo{author}{{Clerc}, N.}, \bibinfo{author}{{Coffey}, D.},
  \bibinfo{author}{{Comerford}, J.M.}, \bibinfo{author}{{Comparat}, J.},
  \bibinfo{author}{{da Costa}, L.}, \bibinfo{author}{{Cousinou}, M.C.},
  \bibinfo{author}{{Covey}, K.}, \bibinfo{author}{{Crane}, J.D.},
  \bibinfo{author}{{Cunha}, K.}, \bibinfo{author}{{Ilha}, G.d.S.},
  \bibinfo{author}{{Dai}, Y.S.}, \bibinfo{author}{{Damsted}, S.B.},
  \bibinfo{author}{{Darling}, J.}, \bibinfo{author}{{Davidson}, James~W., J.},
  \bibinfo{author}{{Davies}, R.}, \bibinfo{author}{{Dawson}, K.},
  \bibinfo{author}{{De}, N.}, \bibinfo{author}{{de la Macorra}, A.},
  \bibinfo{author}{{De Lee}, N.}, \bibinfo{author}{{Queiroz}, A.B.d.A.},
  \bibinfo{author}{{Deconto Machado}, A.}, \bibinfo{author}{{de la Torre}, S.},
  \bibinfo{author}{{Dell'Agli}, F.}, \bibinfo{author}{{du Mas des Bourboux},
  H.}, \bibinfo{author}{{Diamond-Stanic}, A.M.}, \bibinfo{author}{{Dillon},
  S.}, \bibinfo{author}{{Donor}, J.}, \bibinfo{author}{{Drory}, N.},
  \bibinfo{author}{{Duckworth}, C.}, \bibinfo{author}{{Dwelly}, T.},
  \bibinfo{author}{{Ebelke}, G.}, \bibinfo{author}{{Eftekharzadeh}, S.},
  \bibinfo{author}{{Davis Eigenbrot}, A.}, \bibinfo{author}{{Elsworth}, Y.P.},
  \bibinfo{author}{{Eracleous}, M.}, \bibinfo{author}{{Erfanianfar}, G.},
  \bibinfo{author}{{Escoffier}, S.}, \bibinfo{author}{{Fan}, X.},
  \bibinfo{author}{{Farr}, E.}, \bibinfo{author}{{Fern{\'a}ndez-Trincado},
  J.G.}, \bibinfo{author}{{Feuillet}, D.}, \bibinfo{author}{{Finoguenov}, A.},
  \bibinfo{author}{{Fofie}, P.}, \bibinfo{author}{{Fraser-McKelvie}, A.},
  \bibinfo{author}{{Frinchaboy}, P.M.}, \bibinfo{author}{{Fromenteau}, S.},
  \bibinfo{author}{{Fu}, H.}, \bibinfo{author}{{Galbany}, L.},
  \bibinfo{author}{{Garcia}, R.A.},
  \bibinfo{author}{{Garc{\'\i}a-Hern{\'a}ndez}, D.A.},
  \bibinfo{author}{{Oehmichen}, L.A.G.}, \bibinfo{author}{{Ge}, J.},
  \bibinfo{author}{{Maia}, M.A.G.}, \bibinfo{author}{{Geisler}, D.},
  \bibinfo{author}{{Gelfand}, J.}, \bibinfo{author}{{Goddy}, J.},
  \bibinfo{author}{{Gonzalez-Perez}, V.}, \bibinfo{author}{{Grabowski}, K.},
  \bibinfo{author}{{Green}, P.}, \bibinfo{author}{{Grier}, C.J.},
  \bibinfo{author}{{Guo}, H.}, \bibinfo{author}{{Guy}, J.},
  \bibinfo{author}{{Harding}, P.}, \bibinfo{author}{{Hasselquist}, S.},
  \bibinfo{author}{{Hawken}, A.J.}, \bibinfo{author}{{Hayes}, C.R.},
  \bibinfo{author}{{Hearty}, F.}, \bibinfo{author}{{Hekker}, S.},
  \bibinfo{author}{{Hogg}, D.W.}, \bibinfo{author}{{Holtzman}, J.A.},
  \bibinfo{author}{{Horta}, D.}, \bibinfo{author}{{Hou}, J.},
  \bibinfo{author}{{Hsieh}, B.C.}, \bibinfo{author}{{Huber}, D.},
  \bibinfo{author}{{Hunt}, J.A.S.}, \bibinfo{author}{{Chitham}, J.I.},
  \bibinfo{author}{{Imig}, J.}, \bibinfo{author}{{Jaber}, M.},
  \bibinfo{author}{{Angel}, C.E.J.}, \bibinfo{author}{{Johnson}, J.A.},
  \bibinfo{author}{{Jones}, A.M.}, \bibinfo{author}{{J{\"o}nsson}, H.},
  \bibinfo{author}{{Jullo}, E.}, \bibinfo{author}{{Kim}, Y.},
  \bibinfo{author}{{Kinemuchi}, K.}, \bibinfo{author}{{Kirkpatrick},
  Charles~C., I.}, \bibinfo{author}{{Kite}, G.W.}, \bibinfo{author}{{Klaene},
  M.}, \bibinfo{author}{{Kneib}, J.P.}, \bibinfo{author}{{Kollmeier}, J.A.},
  \bibinfo{author}{{Kong}, H.}, \bibinfo{author}{{Kounkel}, M.},
  \bibinfo{author}{{Krishnarao}, D.}, \bibinfo{author}{{Lacerna}, I.},
  \bibinfo{author}{{Lan}, T.W.}, \bibinfo{author}{{Lane}, R.R.},
  \bibinfo{author}{{Law}, D.R.}, \bibinfo{author}{{Le Goff}, J.M.},
  \bibinfo{author}{{Leung}, H.W.}, \bibinfo{author}{{Lewis}, H.},
  \bibinfo{author}{{Li}, C.}, \bibinfo{author}{{Lian}, J.},
  \bibinfo{author}{{Lin}, L.}, \bibinfo{author}{{Long}, D.},
  \bibinfo{author}{{Longa-Pe{\~n}a}, P.}, \bibinfo{author}{{Lundgren}, B.},
  \bibinfo{author}{{Lyke}, B.W.}, \bibinfo{author}{{Ted Mackereth}, J.},
  \bibinfo{author}{{MacLeod}, C.L.}, \bibinfo{author}{{Majewski}, S.R.},
  \bibinfo{author}{{Manchado}, A.}, \bibinfo{author}{{Maraston}, C.},
  \bibinfo{author}{{Martini}, P.}, \bibinfo{author}{{Masseron}, T.},
  \bibinfo{author}{{Masters}, K.L.}, \bibinfo{author}{{Mathur}, S.},
  \bibinfo{author}{{McDermid}, R.M.}, \bibinfo{author}{{Merloni}, A.},
  \bibinfo{author}{{Merrifield}, M.}, \bibinfo{author}{{M{\'e}sz{\'a}ros}, S.},
  \bibinfo{author}{{Miglio}, A.}, \bibinfo{author}{{Minniti}, D.},
  \bibinfo{author}{{Minsley}, R.}, \bibinfo{author}{{Miyaji}, T.},
  \bibinfo{author}{{Mohammad}, F.G.}, \bibinfo{author}{{Mosser}, B.},
  \bibinfo{author}{{Mueller}, E.M.}, \bibinfo{author}{{Muna}, D.},
  \bibinfo{author}{{Mu{\~n}oz-Guti{\'e}rrez}, A.}, \bibinfo{author}{{Myers},
  A.D.}, \bibinfo{author}{{Nadathur}, S.}, \bibinfo{author}{{Nair}, P.},
  \bibinfo{author}{{Nandra}, K.}, \bibinfo{author}{{do Nascimento}, J.C.},
  \bibinfo{author}{{Nevin}, R.J.}, \bibinfo{author}{{Newman}, J.A.},
  \bibinfo{author}{{Nidever}, D.L.}, \bibinfo{author}{{Nitschelm}, C.},
  \bibinfo{author}{{Noterdaeme}, P.}, \bibinfo{author}{{O'Connell}, J.E.},
  \bibinfo{author}{{Olmstead}, M.D.}, \bibinfo{author}{{Oravetz}, D.},
  \bibinfo{author}{{Oravetz}, A.}, \bibinfo{author}{{Osorio}, Y.},
  \bibinfo{author}{{Pace}, Z.J.}, \bibinfo{author}{{Padilla}, N.},
  \bibinfo{author}{{Palanque-Delabrouille}, N.}, \bibinfo{author}{{Palicio},
  P.A.}, \bibinfo{author}{{Pan}, H.A.}, \bibinfo{author}{{Pan}, K.},
  \bibinfo{author}{{Parker}, J.}, \bibinfo{author}{{Paviot}, R.},
  \bibinfo{author}{{Peirani}, S.}, \bibinfo{author}{{Ram{\'r}ez}, K.P.},
  \bibinfo{author}{{Penny}, S.}, \bibinfo{author}{{Percival}, W.J.},
  \bibinfo{author}{{Perez-Fournon}, I.},
  \bibinfo{author}{{P{\'e}rez-R{\`a}fols}, I.}, \bibinfo{author}{{Petitjean},
  P.}, \bibinfo{author}{{Pieri}, M.M.}, \bibinfo{author}{{Pinsonneault}, M.},
  \bibinfo{author}{{Poovelil}, V.J.}, \bibinfo{author}{{Povick}, J.T.},
  \bibinfo{author}{{Prakash}, A.}, \bibinfo{author}{{Price-Whelan}, A.M.},
  \bibinfo{author}{{Raddick}, M.J.}, \bibinfo{author}{{Raichoor}, A.},
  \bibinfo{author}{{Ray}, A.}, \bibinfo{author}{{Rembold}, S.B.},
  \bibinfo{author}{{Rezaie}, M.}, \bibinfo{author}{{Riffel}, R.A.},
  \bibinfo{author}{{Riffel}, R.}, \bibinfo{author}{{Rix}, H.W.},
  \bibinfo{author}{{Robin}, A.C.}, \bibinfo{author}{{Roman-Lopes}, A.},
  \bibinfo{author}{{Rom{\'a}n-Z{\'u}{\~n}iga}, C.}, \bibinfo{author}{{Rose},
  B.}, \bibinfo{author}{{Ross}, A.J.}, \bibinfo{author}{{Rossi}, G.},
  \bibinfo{author}{{Rowlands}, K.}, \bibinfo{author}{{Rubin}, K.H.R.},
  \bibinfo{author}{{Salvato}, M.}, \bibinfo{author}{{S{\'a}nchez}, A.G.},
  \bibinfo{author}{{S{\'a}nchez-Menguiano}, L.},
  \bibinfo{author}{{S{\'a}nchez-Gallego}, J.R.}, \bibinfo{author}{{Sayres},
  C.}, \bibinfo{author}{{Schaefer}, A.}, \bibinfo{author}{{Schiavon}, R.P.},
  \bibinfo{author}{{Schimoia}, J.S.}, \bibinfo{author}{{Schlafly}, E.},
  \bibinfo{author}{{Schlegel}, D.}, \bibinfo{author}{{Schneider}, D.P.},
  \bibinfo{author}{{Schultheis}, M.}, \bibinfo{author}{{Schwope}, A.},
  \bibinfo{author}{{Seo}, H.J.}, \bibinfo{author}{{Serenelli}, A.},
  \bibinfo{author}{{Shafieloo}, A.}, \bibinfo{author}{{Shamsi}, S.J.},
  \bibinfo{author}{{Shao}, Z.}, \bibinfo{author}{{Shen}, S.},
  \bibinfo{author}{{Shetrone}, M.}, \bibinfo{author}{{Shirley}, R.},
  \bibinfo{author}{{Aguirre}, V.S.}, \bibinfo{author}{{Simon}, J.D.},
  \bibinfo{author}{{Skrutskie}, M.F.}, \bibinfo{author}{{Slosar}, A.},
  \bibinfo{author}{{Smethurst}, R.}, \bibinfo{author}{{Sobeck}, J.},
  \bibinfo{author}{{Sodi}, B.C.}, \bibinfo{author}{{Souto}, D.},
  \bibinfo{author}{{Stark}, D.V.}, \bibinfo{author}{{Stassun}, K.G.},
  \bibinfo{author}{{Steinmetz}, M.}, \bibinfo{author}{{Stello}, D.},
  \bibinfo{author}{{Stermer}, J.}, \bibinfo{author}{{Storchi-Bergmann}, T.},
  \bibinfo{author}{{Streblyanska}, A.}, \bibinfo{author}{{Stringfellow}, G.S.},
  \bibinfo{author}{{Stutz}, A.}, \bibinfo{author}{{Su{\'a}rez}, G.},
  \bibinfo{author}{{Sun}, J.}, \bibinfo{author}{{Taghizadeh-Popp}, M.},
  \bibinfo{author}{{Talbot}, M.S.}, \bibinfo{author}{{Tayar}, J.},
  \bibinfo{author}{{Thakar}, A.R.}, \bibinfo{author}{{Theriault}, R.},
  \bibinfo{author}{{Thomas}, D.}, \bibinfo{author}{{Thomas}, Z.C.},
  \bibinfo{author}{{Tinker}, J.}, \bibinfo{author}{{Tojeiro}, R.},
  \bibinfo{author}{{Toledo}, H.H.}, \bibinfo{author}{{Tremonti}, C.A.},
  \bibinfo{author}{{Troup}, N.W.}, \bibinfo{author}{{Tuttle}, S.},
  \bibinfo{author}{{Unda-Sanzana}, E.}, \bibinfo{author}{{Valentini}, M.},
  \bibinfo{author}{{Vargas-Gonz{\'a}lez}, J.},
  \bibinfo{author}{{Vargas-Maga{\~n}a}, M.},
  \bibinfo{author}{{V{\'a}zquez-Mata}, J.A.}, \bibinfo{author}{{Vivek}, M.},
  \bibinfo{author}{{Wake}, D.}, \bibinfo{author}{{Wang}, Y.},
  \bibinfo{author}{{Weaver}, B.A.}, \bibinfo{author}{{Weijmans}, A.M.},
  \bibinfo{author}{{Wild}, V.}, \bibinfo{author}{{Wilson}, J.C.},
  \bibinfo{author}{{Wilson}, R.F.}, \bibinfo{author}{{Wolthuis}, N.},
  \bibinfo{author}{{Wood-Vasey}, W.M.}, \bibinfo{author}{{Yan}, R.},
  \bibinfo{author}{{Yang}, M.}, \bibinfo{author}{{Y{\`e}che}, C.},
  \bibinfo{author}{{Zamora}, O.}, \bibinfo{author}{{Zarrouk}, P.},
  \bibinfo{author}{{Zasowski}, G.}, \bibinfo{author}{{Zhang}, K.},
  \bibinfo{author}{{Zhao}, C.}, \bibinfo{author}{{Zhao}, G.},
  \bibinfo{author}{{Zheng}, Z.}, \bibinfo{author}{{Zheng}, Z.},
  \bibinfo{author}{{Zhu}, G.}, \bibinfo{author}{{Zou}, H.},
  \bibinfo{year}{2020}.
\newblock \bibinfo{title}{{The 16th Data Release of the Sloan Digital Sky
  Surveys: First Release from the APOGEE-2 Southern Survey and Full Release of
  eBOSS Spectra}}.
\newblock \bibinfo{journal}{\apjs} \bibinfo{volume}{249}, \bibinfo{pages}{3}.
\newblock \DOIprefix\doi{10.3847/1538-4365/ab929e},
  \href{http://arxiv.org/abs/1912.02905}{\tt arXiv:1912.02905}.
\bibitem[{{Avni} and {Bahcall}(1980)}]{Avni:1980}
\bibinfo{author}{{Avni}, Y.}, \bibinfo{author}{{Bahcall}, J.N.},
  \bibinfo{year}{1980}.
\newblock \bibinfo{title}{{On the simultaneous analysis of several complete
  samples - The V/Vmax and Ve/Va variables, with applications to quasars}}.
\newblock \bibinfo{journal}{\apj} \bibinfo{volume}{235},
  \bibinfo{pages}{694--716}.
\newblock \DOIprefix\doi{10.1086/157673}.
\bibitem[{{Bailer-Jones} et~al.(2019){Bailer-Jones}, {Fouesneau} and
  {Andrae}}]{Bailer-Jones:2019}
\bibinfo{author}{{Bailer-Jones}, C.A.L.}, \bibinfo{author}{{Fouesneau}, M.},
  \bibinfo{author}{{Andrae}, R.}, \bibinfo{year}{2019}.
\newblock \bibinfo{title}{{Quasar and galaxy classification in Gaia Data
  Release 2}}.
\newblock \bibinfo{journal}{\mnras} \bibinfo{volume}{490},
  \bibinfo{pages}{5615--5633}.
\newblock \DOIprefix\doi{10.1093/mnras/stz2947},
  \href{http://arxiv.org/abs/1910.05255}{\tt arXiv:1910.05255}.
\bibitem[{{Blanton} et~al.(2003){Blanton}, {Hogg}, {Bahcall}, {Baldry},
  {Brinkmann}, {Csabai}, {Eisenstein}, {Fukugita}, {Gunn}, {Ivezi{\'c}},
  {Lamb}, {Lupton}, {Loveday}, {Munn}, {Nichol}, {Okamura}, {Schlegel},
  {Shimasaku}, {Strauss}, {Vogeley} and {Weinberg}}]{Blanton:2003}
\bibinfo{author}{{Blanton}, M.R.}, \bibinfo{author}{{Hogg}, D.W.},
  \bibinfo{author}{{Bahcall}, N.A.}, \bibinfo{author}{{Baldry}, I.K.},
  \bibinfo{author}{{Brinkmann}, J.}, \bibinfo{author}{{Csabai}, I.},
  \bibinfo{author}{{Eisenstein}, D.}, \bibinfo{author}{{Fukugita}, M.},
  \bibinfo{author}{{Gunn}, J.E.}, \bibinfo{author}{{Ivezi{\'c}}, {\v Z}.},
  \bibinfo{author}{{Lamb}, D.Q.}, \bibinfo{author}{{Lupton}, R.H.},
  \bibinfo{author}{{Loveday}, J.}, \bibinfo{author}{{Munn}, J.A.},
  \bibinfo{author}{{Nichol}, R.C.}, \bibinfo{author}{{Okamura}, S.},
  \bibinfo{author}{{Schlegel}, D.J.}, \bibinfo{author}{{Shimasaku}, K.},
  \bibinfo{author}{{Strauss}, M.A.}, \bibinfo{author}{{Vogeley}, M.S.},
  \bibinfo{author}{{Weinberg}, D.H.}, \bibinfo{year}{2003}.
\newblock \bibinfo{title}{{The Broadband Optical Properties of Galaxies with
  Redshifts $0.02<z<0.22$}}.
\newblock \bibinfo{journal}{\apj} \bibinfo{volume}{594},
  \bibinfo{pages}{186--207}.
\newblock \DOIprefix\doi{10.1086/375528},
  \href{http://arxiv.org/abs/astro-ph/0209479}{\tt arXiv:astro-ph/0209479}.
\bibitem[{{Burnham} and {Anderson}(2002)}]{Burnham:2002}
\bibinfo{author}{{Burnham}, K.P.}, \bibinfo{author}{{Anderson}, D.R.},
  \bibinfo{year}{2002}.
\newblock \bibinfo{title}{{Model Selection and Multi-Model Inference}}.
\newblock \bibinfo{edition}{2} ed., \bibinfo{publisher}{Springer-Verlag New
  York, Inc.}, \bibinfo{address}{175 Fifth Avenue, New York, NY 10010}.
\bibitem[{{Chambers} et~al.(2016){Chambers}, {Magnier}, {Metcalfe},
  {Flewelling}, {Huber}, {Waters}, {Denneau}, {Draper}, {Farrow}, {Finkbeiner},
  {Holmberg}, {Koppenhoefer}, {Price}, {Rest}, {Saglia}, {Schlafly}, {Smartt},
  {Sweeney}, {Wainscoat}, {Burgett}, {Chastel}, {Grav}, {Heasley}, {Hodapp},
  {Jedicke}, {Kaiser}, {Kudritzki}, {Luppino}, {Lupton}, {Monet}, {Morgan},
  {Onaka}, {Shiao}, {Stubbs}, {Tonry}, {White}, {Ba{\~n}ados}, {Bell},
  {Bender}, {Bernard}, {Boegner}, {Boffi}, {Botticella}, {Calamida},
  {Casertano}, {Chen}, {Chen}, {Cole}, {Deacon}, {Frenk}, {Fitzsimmons},
  {Gezari}, {Gibbs}, {Goessl}, {Goggia}, {Gourgue}, {Goldman}, {Grant},
  {Grebel}, {Hambly}, {Hasinger}, {Heavens}, {Heckman}, {Henderson}, {Henning},
  {Holman}, {Hopp}, {Ip}, {Isani}, {Jackson}, {Keyes}, {Koekemoer}, {Kotak},
  {Le}, {Liska}, {Long}, {Lucey}, {Liu}, {Martin}, {Masci}, {McLean}, {Mindel},
  {Misra}, {Morganson}, {Murphy}, {Obaika}, {Narayan}, {Nieto-Santisteban},
  {Norberg}, {Peacock}, {Pier}, {Postman}, {Primak}, {Rae}, {Rai}, {Riess},
  {Riffeser}, {Rix}, {R{\"o}ser}, {Russel}, {Rutz}, {Schilbach}, {Schultz},
  {Scolnic}, {Strolger}, {Szalay}, {Seitz}, {Small}, {Smith}, {Soderblom},
  {Taylor}, {Thomson}, {Taylor}, {Thakar}, {Thiel}, {Thilker}, {Unger},
  {Urata}, {Valenti}, {Wagner}, {Walder}, {Walter}, {Watters}, {Werner},
  {Wood-Vasey} and {Wyse}}]{PanStarrsPS1}
\bibinfo{author}{{Chambers}, K.C.}, \bibinfo{author}{{Magnier}, E.A.},
  \bibinfo{author}{{Metcalfe}, N.}, \bibinfo{author}{{Flewelling}, H.A.},
  \bibinfo{author}{{Huber}, M.E.}, \bibinfo{author}{{Waters}, C.Z.},
  \bibinfo{author}{{Denneau}, L.}, \bibinfo{author}{{Draper}, P.W.},
  \bibinfo{author}{{Farrow}, D.}, \bibinfo{author}{{Finkbeiner}, D.P.},
  \bibinfo{author}{{Holmberg}, C.}, \bibinfo{author}{{Koppenhoefer}, J.},
  \bibinfo{author}{{Price}, P.A.}, \bibinfo{author}{{Rest}, A.},
  \bibinfo{author}{{Saglia}, R.P.}, \bibinfo{author}{{Schlafly}, E.F.},
  \bibinfo{author}{{Smartt}, S.J.}, \bibinfo{author}{{Sweeney}, W.},
  \bibinfo{author}{{Wainscoat}, R.J.}, \bibinfo{author}{{Burgett}, W.S.},
  \bibinfo{author}{{Chastel}, S.}, \bibinfo{author}{{Grav}, T.},
  \bibinfo{author}{{Heasley}, J.N.}, \bibinfo{author}{{Hodapp}, K.W.},
  \bibinfo{author}{{Jedicke}, R.}, \bibinfo{author}{{Kaiser}, N.},
  \bibinfo{author}{{Kudritzki}, R.P.}, \bibinfo{author}{{Luppino}, G.A.},
  \bibinfo{author}{{Lupton}, R.H.}, \bibinfo{author}{{Monet}, D.G.},
  \bibinfo{author}{{Morgan}, J.S.}, \bibinfo{author}{{Onaka}, P.M.},
  \bibinfo{author}{{Shiao}, B.}, \bibinfo{author}{{Stubbs}, C.W.},
  \bibinfo{author}{{Tonry}, J.L.}, \bibinfo{author}{{White}, R.},
  \bibinfo{author}{{Ba{\~n}ados}, E.}, \bibinfo{author}{{Bell}, E.F.},
  \bibinfo{author}{{Bender}, R.}, \bibinfo{author}{{Bernard}, E.J.},
  \bibinfo{author}{{Boegner}, M.}, \bibinfo{author}{{Boffi}, F.},
  \bibinfo{author}{{Botticella}, M.T.}, \bibinfo{author}{{Calamida}, A.},
  \bibinfo{author}{{Casertano}, S.}, \bibinfo{author}{{Chen}, W.P.},
  \bibinfo{author}{{Chen}, X.}, \bibinfo{author}{{Cole}, S.},
  \bibinfo{author}{{Deacon}, N.}, \bibinfo{author}{{Frenk}, C.},
  \bibinfo{author}{{Fitzsimmons}, A.}, \bibinfo{author}{{Gezari}, S.},
  \bibinfo{author}{{Gibbs}, V.}, \bibinfo{author}{{Goessl}, C.},
  \bibinfo{author}{{Goggia}, T.}, \bibinfo{author}{{Gourgue}, R.},
  \bibinfo{author}{{Goldman}, B.}, \bibinfo{author}{{Grant}, P.},
  \bibinfo{author}{{Grebel}, E.K.}, \bibinfo{author}{{Hambly}, N.C.},
  \bibinfo{author}{{Hasinger}, G.}, \bibinfo{author}{{Heavens}, A.F.},
  \bibinfo{author}{{Heckman}, T.M.}, \bibinfo{author}{{Henderson}, R.},
  \bibinfo{author}{{Henning}, T.}, \bibinfo{author}{{Holman}, M.},
  \bibinfo{author}{{Hopp}, U.}, \bibinfo{author}{{Ip}, W.H.},
  \bibinfo{author}{{Isani}, S.}, \bibinfo{author}{{Jackson}, M.},
  \bibinfo{author}{{Keyes}, C.D.}, \bibinfo{author}{{Koekemoer}, A.M.},
  \bibinfo{author}{{Kotak}, R.}, \bibinfo{author}{{Le}, D.},
  \bibinfo{author}{{Liska}, D.}, \bibinfo{author}{{Long}, K.S.},
  \bibinfo{author}{{Lucey}, J.R.}, \bibinfo{author}{{Liu}, M.},
  \bibinfo{author}{{Martin}, N.F.}, \bibinfo{author}{{Masci}, G.},
  \bibinfo{author}{{McLean}, B.}, \bibinfo{author}{{Mindel}, E.},
  \bibinfo{author}{{Misra}, P.}, \bibinfo{author}{{Morganson}, E.},
  \bibinfo{author}{{Murphy}, D.N.A.}, \bibinfo{author}{{Obaika}, A.},
  \bibinfo{author}{{Narayan}, G.}, \bibinfo{author}{{Nieto-Santisteban}, M.A.},
  \bibinfo{author}{{Norberg}, P.}, \bibinfo{author}{{Peacock}, J.A.},
  \bibinfo{author}{{Pier}, E.A.}, \bibinfo{author}{{Postman}, M.},
  \bibinfo{author}{{Primak}, N.}, \bibinfo{author}{{Rae}, C.},
  \bibinfo{author}{{Rai}, A.}, \bibinfo{author}{{Riess}, A.},
  \bibinfo{author}{{Riffeser}, A.}, \bibinfo{author}{{Rix}, H.W.},
  \bibinfo{author}{{R{\"o}ser}, S.}, \bibinfo{author}{{Russel}, R.},
  \bibinfo{author}{{Rutz}, L.}, \bibinfo{author}{{Schilbach}, E.},
  \bibinfo{author}{{Schultz}, A.S.B.}, \bibinfo{author}{{Scolnic}, D.},
  \bibinfo{author}{{Strolger}, L.}, \bibinfo{author}{{Szalay}, A.},
  \bibinfo{author}{{Seitz}, S.}, \bibinfo{author}{{Small}, E.},
  \bibinfo{author}{{Smith}, K.W.}, \bibinfo{author}{{Soderblom}, D.R.},
  \bibinfo{author}{{Taylor}, P.}, \bibinfo{author}{{Thomson}, R.},
  \bibinfo{author}{{Taylor}, A.N.}, \bibinfo{author}{{Thakar}, A.R.},
  \bibinfo{author}{{Thiel}, J.}, \bibinfo{author}{{Thilker}, D.},
  \bibinfo{author}{{Unger}, D.}, \bibinfo{author}{{Urata}, Y.},
  \bibinfo{author}{{Valenti}, J.}, \bibinfo{author}{{Wagner}, J.},
  \bibinfo{author}{{Walder}, T.}, \bibinfo{author}{{Walter}, F.},
  \bibinfo{author}{{Watters}, S.P.}, \bibinfo{author}{{Werner}, S.},
  \bibinfo{author}{{Wood-Vasey}, W.M.}, \bibinfo{author}{{Wyse}, R.},
  \bibinfo{year}{2016}.
\newblock \bibinfo{title}{{The Pan-STARRS1 Surveys}}.
\newblock \bibinfo{journal}{arXiv e-prints} ,
  \bibinfo{pages}{arXiv:1612.05560}\href{http://arxiv.org/abs/1612.05560}{\tt
  arXiv:1612.05560}.
\bibitem[{{Cheng} et~al.(2021){Cheng}, {Huertas-Company}, {Conselice},
  {Arag{\'o}n-Salamanca}, {Robertson} and {Ramachandra}}]{Cheng:2021a}
\bibinfo{author}{{Cheng}, T.Y.}, \bibinfo{author}{{Huertas-Company}, M.},
  \bibinfo{author}{{Conselice}, C.J.}, \bibinfo{author}{{Arag{\'o}n-Salamanca},
  A.}, \bibinfo{author}{{Robertson}, B.E.}, \bibinfo{author}{{Ramachandra},
  N.}, \bibinfo{year}{2021}.
\newblock \bibinfo{title}{{Beyond the hubble sequence - exploring galaxy
  morphology with unsupervised machine learning}}.
\newblock \bibinfo{journal}{\mnras} \bibinfo{volume}{503},
  \bibinfo{pages}{4446--4465}.
\newblock \DOIprefix\doi{10.1093/mnras/stab734},
  \href{http://arxiv.org/abs/2009.11932}{\tt arXiv:2009.11932}.
\bibitem[{{Clarke} et~al.(2020){Clarke}, {Scaife}, {Greenhalgh} and
  {Griguta}}]{Clarke:2020}
\bibinfo{author}{{Clarke}, A.O.}, \bibinfo{author}{{Scaife}, A.M.M.},
  \bibinfo{author}{{Greenhalgh}, R.}, \bibinfo{author}{{Griguta}, V.},
  \bibinfo{year}{2020}.
\newblock \bibinfo{title}{{Identifying galaxies, quasars, and stars with
  machine learning: A new catalogue of classifications for 111 million SDSS
  sources without spectra}}.
\newblock \bibinfo{journal}{\aap} \bibinfo{volume}{639}, \bibinfo{pages}{A84}.
\newblock \DOIprefix\doi{10.1051/0004-6361/201936770},
  \href{http://arxiv.org/abs/1909.10963}{\tt arXiv:1909.10963}.
\bibitem[{{Cutri} et~al.(2006){Cutri}, {Skrutskie}, {Van~Dyk}, {Beichman},
  {Carpenter}, {Chester}, {Cambresy}, {Evans}, {Fowler}, {Gizis}, {Howard},
  {Huchra}, {Jarrett}, {Kopan}, {Kirkpatrick}, {Light}, {Marsh}, {McCallon},
  {Schneider}, {Stiening}, {Sykes}, {Weinberg}, {Wheaton}, {Wheelock} and
  {Zacharias}}]{Cutri:2006}
\bibinfo{author}{{Cutri}, R.M.}, \bibinfo{author}{{Skrutskie}, M.F.},
  \bibinfo{author}{{Van~Dyk}, S.}, \bibinfo{author}{{Beichman}, C.A.},
  \bibinfo{author}{{Carpenter}, J.M.}, \bibinfo{author}{{Chester}, T.},
  \bibinfo{author}{{Cambresy}, L.}, \bibinfo{author}{{Evans}, T.},
  \bibinfo{author}{{Fowler}, J.}, \bibinfo{author}{{Gizis}, J.},
  \bibinfo{author}{{Howard}, E.}, \bibinfo{author}{{Huchra}, J.},
  \bibinfo{author}{{Jarrett}, T.}, \bibinfo{author}{{Kopan}, E.L.},
  \bibinfo{author}{{Kirkpatrick}, J.D.}, \bibinfo{author}{{Light}, R.M.},
  \bibinfo{author}{{Marsh}, K.A.}, \bibinfo{author}{{McCallon}, H.},
  \bibinfo{author}{{Schneider}, S.}, \bibinfo{author}{{Stiening}, R.},
  \bibinfo{author}{{Sykes}, M.}, \bibinfo{author}{{Weinberg}, M.},
  \bibinfo{author}{{Wheaton}, W.A.}, \bibinfo{author}{{Wheelock}, S.},
  \bibinfo{author}{{Zacharias}, N.}, \bibinfo{year}{2006}.
\newblock \bibinfo{title}{{Explanatory Supplement to the 2MASS All Sky Data
  Release and Extended Mission Products}}.
\newblock \bibinfo{type}{Technical Report}.
\newblock \URLprefix
  \url{https://irsa.ipac.caltech.edu/data/2MASS/docs/releases/allsky/doc/explsup.html}.
\bibitem[{{Cutri} et~al.(2013){Cutri}, {Wright}, {Conrow}, {Fowler},
  {Eisenhardt}, {Grillmair}, {Kirkpatrick}, {Masci}, {McCallon}, {Wheelock},
  {Fajardo-Acosta}, {Yan}, {Benford}, {Harbut}, {Jarrett}, {Lake}, {Leisawitz},
  {Ressler}, {Stanford}, {Tsai}, {Liu}, {Helou}, {Mainzer}, {Gettings},
  {Gonzalez}, {Hoffman}, {Marsh}, {Padgett}, {Skrutskie}, {Beck}, {Papin} and
  {Wittman}}]{Cutri:2013}
\bibinfo{author}{{Cutri}, R.M.}, \bibinfo{author}{{Wright}, E.L.},
  \bibinfo{author}{{Conrow}, T.}, \bibinfo{author}{{Fowler}, J.W.},
  \bibinfo{author}{{Eisenhardt}, P.R.M.}, \bibinfo{author}{{Grillmair}, C.},
  \bibinfo{author}{{Kirkpatrick}, J.D.}, \bibinfo{author}{{Masci}, F.},
  \bibinfo{author}{{McCallon}, H.L.}, \bibinfo{author}{{Wheelock}, S.L.},
  \bibinfo{author}{{Fajardo-Acosta}, S.}, \bibinfo{author}{{Yan}, L.},
  \bibinfo{author}{{Benford}, D.}, \bibinfo{author}{{Harbut}, M.},
  \bibinfo{author}{{Jarrett}, T.}, \bibinfo{author}{{Lake}, S.},
  \bibinfo{author}{{Leisawitz}, D.}, \bibinfo{author}{{Ressler}, M.E.},
  \bibinfo{author}{{Stanford}, S.A.}, \bibinfo{author}{{Tsai}, C.W.},
  \bibinfo{author}{{Liu}, F.}, \bibinfo{author}{{Helou}, G.},
  \bibinfo{author}{{Mainzer}, A.}, \bibinfo{author}{{Gettings}, D.},
  \bibinfo{author}{{Gonzalez}, A.}, \bibinfo{author}{{Hoffman}, D.},
  \bibinfo{author}{{Marsh}, K.A.}, \bibinfo{author}{{Padgett}, D.},
  \bibinfo{author}{{Skrutskie}, M.F.}, \bibinfo{author}{{Beck}, R.P.},
  \bibinfo{author}{{Papin}, M.}, \bibinfo{author}{{Wittman}, M.},
  \bibinfo{year}{2013}.
\newblock \bibinfo{title}{{Explanatory Supplement to the AllWISE Data Release
  Products}}.
\newblock \bibinfo{type}{Technical Report}.
\newblock \URLprefix
  \url{http://wise2.ipac.caltech.edu/docs/release/allwise/expsup/index.html}.
\bibitem[{{Delchambre} et~al.(2022){Delchambre}, {Bailer-Jones},
  {Bellas-Velidis}, {Drimmel}, {Garabato}, {Carballo}, {Hatzidimitriou},
  {Marshall}, {Andrae}, {Dafonte}, {Livanou}, {Fouesneau}, {Licata},
  {Lindstrom}, {Manteiga}, {Robin}, {Silvelo}, {Abreu Aramburu}, {Alvarez},
  {Bakker}, {Bijaoui}, {Brouillet}, {Brugaletta}, {Burlacu}, {Casamiquela},
  {Chaoul}, {Chiavassa}, {Contursi}, {Cooper}, {Creevey}, {Dapergolas}, {de
  Laverny}, {Demouchy}, {Dharmawardena}, {Edvardsson}, {Fremat},
  {Garcia-Lario}, {Garcia-Torres}, {Gavel}, {Gomez}, {Gonzalez-Santamaria},
  {Heiter}, {Jean-Antoine Piccolo}, {Kontizas}, {Kordopatis}, {Korn},
  {Lanzafame}, {Lebreton}, {Lobel}, {Lorca}, {Magdaleno Romeo}, {Marocco},
  {Mary}, {Nicolas}, {Ordenovic}, {Pailler}, {Palicio}, {Pallas-Quintela},
  {Panem}, {Pichon}, {Poggio}, {Recio-Blanco}, {Riclet}, {Rybizki},
  {Santovena}, {Sarro}, {Schultheis}, {Segol}, {Slezak}, {Smart}, {Sordo},
  {Soubiran}, {Suveges}, {Thevenin}, {Torralba Elipe}, {Ulla}, {Utrilla},
  {Vallenari}, {van Dillen}, {Zhao} and {Zorec}}]{Delchambre:2022}
\bibinfo{author}{{Delchambre}, L.}, \bibinfo{author}{{Bailer-Jones}, C.A.L.},
  \bibinfo{author}{{Bellas-Velidis}, I.}, \bibinfo{author}{{Drimmel}, R.},
  \bibinfo{author}{{Garabato}, D.}, \bibinfo{author}{{Carballo}, R.},
  \bibinfo{author}{{Hatzidimitriou}, D.}, \bibinfo{author}{{Marshall}, D.J.},
  \bibinfo{author}{{Andrae}, R.}, \bibinfo{author}{{Dafonte}, C.},
  \bibinfo{author}{{Livanou}, E.}, \bibinfo{author}{{Fouesneau}, M.},
  \bibinfo{author}{{Licata}, E.L.}, \bibinfo{author}{{Lindstrom}, H.E.P.},
  \bibinfo{author}{{Manteiga}, M.}, \bibinfo{author}{{Robin}, C.},
  \bibinfo{author}{{Silvelo}, A.}, \bibinfo{author}{{Abreu Aramburu}, A.},
  \bibinfo{author}{{Alvarez}, M.A.}, \bibinfo{author}{{Bakker}, J.},
  \bibinfo{author}{{Bijaoui}, A.}, \bibinfo{author}{{Brouillet}, N.},
  \bibinfo{author}{{Brugaletta}, E.}, \bibinfo{author}{{Burlacu}, A.},
  \bibinfo{author}{{Casamiquela}, L.}, \bibinfo{author}{{Chaoul}, L.},
  \bibinfo{author}{{Chiavassa}, A.}, \bibinfo{author}{{Contursi}, G.},
  \bibinfo{author}{{Cooper}, W.J.}, \bibinfo{author}{{Creevey}, O.L.},
  \bibinfo{author}{{Dapergolas}, A.}, \bibinfo{author}{{de Laverny}, P.},
  \bibinfo{author}{{Demouchy}, C.}, \bibinfo{author}{{Dharmawardena}, T.E.},
  \bibinfo{author}{{Edvardsson}, B.}, \bibinfo{author}{{Fremat}, Y.},
  \bibinfo{author}{{Garcia-Lario}, P.}, \bibinfo{author}{{Garcia-Torres}, M.},
  \bibinfo{author}{{Gavel}, A.}, \bibinfo{author}{{Gomez}, A.},
  \bibinfo{author}{{Gonzalez-Santamaria}, I.}, \bibinfo{author}{{Heiter}, U.},
  \bibinfo{author}{{Jean-Antoine Piccolo}, A.}, \bibinfo{author}{{Kontizas},
  M.}, \bibinfo{author}{{Kordopatis}, G.}, \bibinfo{author}{{Korn}, A.J.},
  \bibinfo{author}{{Lanzafame}, A.C.}, \bibinfo{author}{{Lebreton}, Y.},
  \bibinfo{author}{{Lobel}, A.}, \bibinfo{author}{{Lorca}, A.},
  \bibinfo{author}{{Magdaleno Romeo}, A.}, \bibinfo{author}{{Marocco}, F.},
  \bibinfo{author}{{Mary}, N.}, \bibinfo{author}{{Nicolas}, C.},
  \bibinfo{author}{{Ordenovic}, C.}, \bibinfo{author}{{Pailler}, F.},
  \bibinfo{author}{{Palicio}, P.A.}, \bibinfo{author}{{Pallas-Quintela}, L.},
  \bibinfo{author}{{Panem}, C.}, \bibinfo{author}{{Pichon}, B.},
  \bibinfo{author}{{Poggio}, E.}, \bibinfo{author}{{Recio-Blanco}, A.},
  \bibinfo{author}{{Riclet}, F.}, \bibinfo{author}{{Rybizki}, J.},
  \bibinfo{author}{{Santovena}, R.}, \bibinfo{author}{{Sarro}, L.M.},
  \bibinfo{author}{{Schultheis}, M.S.}, \bibinfo{author}{{Segol}, M.},
  \bibinfo{author}{{Slezak}, I.}, \bibinfo{author}{{Smart}, R.L.},
  \bibinfo{author}{{Sordo}, R.}, \bibinfo{author}{{Soubiran}, C.},
  \bibinfo{author}{{Suveges}, M.}, \bibinfo{author}{{Thevenin}, F.},
  \bibinfo{author}{{Torralba Elipe}, G.}, \bibinfo{author}{{Ulla}, A.},
  \bibinfo{author}{{Utrilla}, E.}, \bibinfo{author}{{Vallenari}, A.},
  \bibinfo{author}{{van Dillen}, E.}, \bibinfo{author}{{Zhao}, H.},
  \bibinfo{author}{{Zorec}, J.}, \bibinfo{year}{2022}.
\newblock \bibinfo{title}{{Gaia DR3: Apsis III -- Non-stellar content and
  source classification}}.
\newblock \bibinfo{journal}{arXiv e-prints} ,
  \bibinfo{pages}{arXiv:2206.06710}\href{http://arxiv.org/abs/2206.06710}{\tt
  arXiv:2206.06710}.
\bibitem[{Dempster et~al.(1977)Dempster, Laird and Rubin}]{Dempster:1977}
\bibinfo{author}{Dempster, A.P.}, \bibinfo{author}{Laird, N.M.},
  \bibinfo{author}{Rubin, D.B.}, \bibinfo{year}{1977}.
\newblock \bibinfo{title}{Maximum likelihood from incomplete data via the em
  algorithm}.
\newblock \bibinfo{journal}{Journal of the Royal Statistical Society: Series B
  (Methodological)} \bibinfo{volume}{39}, \bibinfo{pages}{1--22}.
\newblock \DOIprefix\doi{10.1111/j.2517-6161.1977.tb01600.x}.
\bibitem[{{Dewdney} et~al.(2009){Dewdney}, {Hall}, {Schilizzi} and
  {Lazio}}]{Dewdney:2009}
\bibinfo{author}{{Dewdney}, P.E.}, \bibinfo{author}{{Hall}, P.J.},
  \bibinfo{author}{{Schilizzi}, R.T.}, \bibinfo{author}{{Lazio}, T.J.L.W.},
  \bibinfo{year}{2009}.
\newblock \bibinfo{title}{{The Square Kilometre Array}}.
\newblock \bibinfo{journal}{IEEE Proceedings} \bibinfo{volume}{97},
  \bibinfo{pages}{1482--1496}.
\newblock \DOIprefix\doi{10.1109/JPROC.2009.2021005}.
\bibitem[{{Dey} et~al.(2019){Dey}, {Schlegel}, {Lang}, {Blum}, {Burleigh},
  {Fan}, {Findlay}, {Finkbeiner}, {Herrera}, {Juneau}, {Landriau}, {Levi},
  {McGreer}, {Meisner}, {Myers}, {Moustakas}, {Nugent}, {Patej}, {Schlafly},
  {Walker}, {Valdes}, {Weaver}, {Y{\`e}che}, {Zou}, {Zhou}, {Abareshi},
  {Abbott}, {Abolfathi}, {Aguilera}, {Alam}, {Allen}, {Alvarez}, {Annis},
  {Ansarinejad}, {Aubert}, {Beechert}, {Bell}, {BenZvi}, {Beutler}, {Bielby},
  {Bolton}, {Brice{\~n}o}, {Buckley-Geer}, {Butler}, {Calamida}, {Carlberg},
  {Carter}, {Casas}, {Castander}, {Choi}, {Comparat}, {Cukanovaite}, {Delubac},
  {DeVries}, {Dey}, {Dhungana}, {Dickinson}, {Ding}, {Donaldson}, {Duan},
  {Duckworth}, {Eftekharzadeh}, {Eisenstein}, {Etourneau}, {Fagrelius},
  {Farihi}, {Fitzpatrick}, {Font-Ribera}, {Fulmer}, {G{\"a}nsicke},
  {Gaztanaga}, {George}, {Gerdes}, {Gontcho}, {Gorgoni}, {Green}, {Guy},
  {Harmer}, {Hernandez}, {Honscheid}, {Huang}, {James}, {Jannuzi}, {Jiang},
  {Joyce}, {Karcher}, {Karkar}, {Kehoe}, {Kneib}, {Kueter-Young}, {Lan},
  {Lauer}, {Le Guillou}, {Le Van Suu}, {Lee}, {Lesser}, {Perreault Levasseur},
  {Li}, {Mann}, {Marshall}, {Mart{\'\i}nez-V{\'a}zquez}, {Martini}, {du Mas des
  Bourboux}, {McManus}, {Meier}, {M{\'e}nard}, {Metcalfe},
  {Mu{\~n}oz-Guti{\'e}rrez}, {Najita}, {Napier}, {Narayan}, {Newman}, {Nie},
  {Nord}, {Norman}, {Olsen}, {Paat}, {Palanque-Delabrouille}, {Peng},
  {Poppett}, {Poremba}, {Prakash}, {Rabinowitz}, {Raichoor}, {Rezaie},
  {Robertson}, {Roe}, {Ross}, {Ross}, {Rudnick}, {Safonova}, {Saha},
  {S{\'a}nchez}, {Savary}, {Schweiker}, {Scott}, {Seo}, {Shan}, {Silva},
  {Slepian}, {Soto}, {Sprayberry}, {Staten}, {Stillman}, {Stupak}, {Summers},
  {Sien Tie}, {Tirado}, {Vargas-Maga{\~n}a}, {Vivas}, {Wechsler}, {Williams},
  {Yang}, {Yang}, {Yapici}, {Zaritsky}, {Zenteno}, {Zhang}, {Zhang}, {Zhou} and
  {Zhou}}]{Dey:2019}
\bibinfo{author}{{Dey}, A.}, \bibinfo{author}{{Schlegel}, D.J.},
  \bibinfo{author}{{Lang}, D.}, \bibinfo{author}{{Blum}, R.},
  \bibinfo{author}{{Burleigh}, K.}, \bibinfo{author}{{Fan}, X.},
  \bibinfo{author}{{Findlay}, J.R.}, \bibinfo{author}{{Finkbeiner}, D.},
  \bibinfo{author}{{Herrera}, D.}, \bibinfo{author}{{Juneau}, S.},
  \bibinfo{author}{{Landriau}, M.}, \bibinfo{author}{{Levi}, M.},
  \bibinfo{author}{{McGreer}, I.}, \bibinfo{author}{{Meisner}, A.},
  \bibinfo{author}{{Myers}, A.D.}, \bibinfo{author}{{Moustakas}, J.},
  \bibinfo{author}{{Nugent}, P.}, \bibinfo{author}{{Patej}, A.},
  \bibinfo{author}{{Schlafly}, E.F.}, \bibinfo{author}{{Walker}, A.R.},
  \bibinfo{author}{{Valdes}, F.}, \bibinfo{author}{{Weaver}, B.A.},
  \bibinfo{author}{{Y{\`e}che}, C.}, \bibinfo{author}{{Zou}, H.},
  \bibinfo{author}{{Zhou}, X.}, \bibinfo{author}{{Abareshi}, B.},
  \bibinfo{author}{{Abbott}, T.M.C.}, \bibinfo{author}{{Abolfathi}, B.},
  \bibinfo{author}{{Aguilera}, C.}, \bibinfo{author}{{Alam}, S.},
  \bibinfo{author}{{Allen}, L.}, \bibinfo{author}{{Alvarez}, A.},
  \bibinfo{author}{{Annis}, J.}, \bibinfo{author}{{Ansarinejad}, B.},
  \bibinfo{author}{{Aubert}, M.}, \bibinfo{author}{{Beechert}, J.},
  \bibinfo{author}{{Bell}, E.F.}, \bibinfo{author}{{BenZvi}, S.Y.},
  \bibinfo{author}{{Beutler}, F.}, \bibinfo{author}{{Bielby}, R.M.},
  \bibinfo{author}{{Bolton}, A.S.}, \bibinfo{author}{{Brice{\~n}o}, C.},
  \bibinfo{author}{{Buckley-Geer}, E.J.}, \bibinfo{author}{{Butler}, K.},
  \bibinfo{author}{{Calamida}, A.}, \bibinfo{author}{{Carlberg}, R.G.},
  \bibinfo{author}{{Carter}, P.}, \bibinfo{author}{{Casas}, R.},
  \bibinfo{author}{{Castander}, F.J.}, \bibinfo{author}{{Choi}, Y.},
  \bibinfo{author}{{Comparat}, J.}, \bibinfo{author}{{Cukanovaite}, E.},
  \bibinfo{author}{{Delubac}, T.}, \bibinfo{author}{{DeVries}, K.},
  \bibinfo{author}{{Dey}, S.}, \bibinfo{author}{{Dhungana}, G.},
  \bibinfo{author}{{Dickinson}, M.}, \bibinfo{author}{{Ding}, Z.},
  \bibinfo{author}{{Donaldson}, J.B.}, \bibinfo{author}{{Duan}, Y.},
  \bibinfo{author}{{Duckworth}, C.J.}, \bibinfo{author}{{Eftekharzadeh}, S.},
  \bibinfo{author}{{Eisenstein}, D.J.}, \bibinfo{author}{{Etourneau}, T.},
  \bibinfo{author}{{Fagrelius}, P.A.}, \bibinfo{author}{{Farihi}, J.},
  \bibinfo{author}{{Fitzpatrick}, M.}, \bibinfo{author}{{Font-Ribera}, A.},
  \bibinfo{author}{{Fulmer}, L.}, \bibinfo{author}{{G{\"a}nsicke}, B.T.},
  \bibinfo{author}{{Gaztanaga}, E.}, \bibinfo{author}{{George}, K.},
  \bibinfo{author}{{Gerdes}, D.W.}, \bibinfo{author}{{Gontcho}, S.G.A.},
  \bibinfo{author}{{Gorgoni}, C.}, \bibinfo{author}{{Green}, G.},
  \bibinfo{author}{{Guy}, J.}, \bibinfo{author}{{Harmer}, D.},
  \bibinfo{author}{{Hernandez}, M.}, \bibinfo{author}{{Honscheid}, K.},
  \bibinfo{author}{{Huang}, L.W.}, \bibinfo{author}{{James}, D.J.},
  \bibinfo{author}{{Jannuzi}, B.T.}, \bibinfo{author}{{Jiang}, L.},
  \bibinfo{author}{{Joyce}, R.}, \bibinfo{author}{{Karcher}, A.},
  \bibinfo{author}{{Karkar}, S.}, \bibinfo{author}{{Kehoe}, R.},
  \bibinfo{author}{{Kneib}, J.P.}, \bibinfo{author}{{Kueter-Young}, A.},
  \bibinfo{author}{{Lan}, T.W.}, \bibinfo{author}{{Lauer}, T.R.},
  \bibinfo{author}{{Le Guillou}, L.}, \bibinfo{author}{{Le Van Suu}, A.},
  \bibinfo{author}{{Lee}, J.H.}, \bibinfo{author}{{Lesser}, M.},
  \bibinfo{author}{{Perreault Levasseur}, L.}, \bibinfo{author}{{Li}, T.S.},
  \bibinfo{author}{{Mann}, J.L.}, \bibinfo{author}{{Marshall}, R.},
  \bibinfo{author}{{Mart{\'\i}nez-V{\'a}zquez}, C.E.},
  \bibinfo{author}{{Martini}, P.}, \bibinfo{author}{{du Mas des Bourboux}, H.},
  \bibinfo{author}{{McManus}, S.}, \bibinfo{author}{{Meier}, T.G.},
  \bibinfo{author}{{M{\'e}nard}, B.}, \bibinfo{author}{{Metcalfe}, N.},
  \bibinfo{author}{{Mu{\~n}oz-Guti{\'e}rrez}, A.}, \bibinfo{author}{{Najita},
  J.}, \bibinfo{author}{{Napier}, K.}, \bibinfo{author}{{Narayan}, G.},
  \bibinfo{author}{{Newman}, J.A.}, \bibinfo{author}{{Nie}, J.},
  \bibinfo{author}{{Nord}, B.}, \bibinfo{author}{{Norman}, D.J.},
  \bibinfo{author}{{Olsen}, K.A.G.}, \bibinfo{author}{{Paat}, A.},
  \bibinfo{author}{{Palanque-Delabrouille}, N.}, \bibinfo{author}{{Peng}, X.},
  \bibinfo{author}{{Poppett}, C.L.}, \bibinfo{author}{{Poremba}, M.R.},
  \bibinfo{author}{{Prakash}, A.}, \bibinfo{author}{{Rabinowitz}, D.},
  \bibinfo{author}{{Raichoor}, A.}, \bibinfo{author}{{Rezaie}, M.},
  \bibinfo{author}{{Robertson}, A.N.}, \bibinfo{author}{{Roe}, N.A.},
  \bibinfo{author}{{Ross}, A.J.}, \bibinfo{author}{{Ross}, N.P.},
  \bibinfo{author}{{Rudnick}, G.}, \bibinfo{author}{{Safonova}, S.},
  \bibinfo{author}{{Saha}, A.}, \bibinfo{author}{{S{\'a}nchez}, F.J.},
  \bibinfo{author}{{Savary}, E.}, \bibinfo{author}{{Schweiker}, H.},
  \bibinfo{author}{{Scott}, A.}, \bibinfo{author}{{Seo}, H.J.},
  \bibinfo{author}{{Shan}, H.}, \bibinfo{author}{{Silva}, D.R.},
  \bibinfo{author}{{Slepian}, Z.}, \bibinfo{author}{{Soto}, C.},
  \bibinfo{author}{{Sprayberry}, D.}, \bibinfo{author}{{Staten}, R.},
  \bibinfo{author}{{Stillman}, C.M.}, \bibinfo{author}{{Stupak}, R.J.},
  \bibinfo{author}{{Summers}, D.L.}, \bibinfo{author}{{Sien Tie}, S.},
  \bibinfo{author}{{Tirado}, H.}, \bibinfo{author}{{Vargas-Maga{\~n}a}, M.},
  \bibinfo{author}{{Vivas}, A.K.}, \bibinfo{author}{{Wechsler}, R.H.},
  \bibinfo{author}{{Williams}, D.}, \bibinfo{author}{{Yang}, J.},
  \bibinfo{author}{{Yang}, Q.}, \bibinfo{author}{{Yapici}, T.},
  \bibinfo{author}{{Zaritsky}, D.}, \bibinfo{author}{{Zenteno}, A.},
  \bibinfo{author}{{Zhang}, K.}, \bibinfo{author}{{Zhang}, T.},
  \bibinfo{author}{{Zhou}, R.}, \bibinfo{author}{{Zhou}, Z.},
  \bibinfo{year}{2019}.
\newblock \bibinfo{title}{{Overview of the DESI Legacy Imaging Surveys}}.
\newblock \bibinfo{journal}{\aj} \bibinfo{volume}{157}, \bibinfo{pages}{168}.
\newblock \DOIprefix\doi{10.3847/1538-3881/ab089d},
  \href{http://arxiv.org/abs/1804.08657}{\tt arXiv:1804.08657}.
\bibitem[{{Doi} et~al.(2010){Doi}, {Tanaka}, {Fukugita}, {Gunn}, {Yasuda},
  {Ivezi{\'c}}, {Brinkmann}, {de Haars}, {Kleinman}, {Krzesinski} and {French
  Leger}}]{Doi:2010}
\bibinfo{author}{{Doi}, M.}, \bibinfo{author}{{Tanaka}, M.},
  \bibinfo{author}{{Fukugita}, M.}, \bibinfo{author}{{Gunn}, J.E.},
  \bibinfo{author}{{Yasuda}, N.}, \bibinfo{author}{{Ivezi{\'c}}, {\v Z}.},
  \bibinfo{author}{{Brinkmann}, J.}, \bibinfo{author}{{de Haars}, E.},
  \bibinfo{author}{{Kleinman}, S.J.}, \bibinfo{author}{{Krzesinski}, J.},
  \bibinfo{author}{{French Leger}, R.}, \bibinfo{year}{2010}.
\newblock \bibinfo{title}{{Photometric Response Functions of the Sloan Digital
  Sky Survey Imager}}.
\newblock \bibinfo{journal}{\aj} \bibinfo{volume}{139},
  \bibinfo{pages}{1628--1648}.
\newblock \DOIprefix\doi{10.1088/0004-6256/139/4/1628},
  \href{http://arxiv.org/abs/1002.3701}{\tt arXiv:1002.3701}.
\bibitem[{{Domingos} and {Pazzani}(1997)}]{Domingos:1997}
\bibinfo{author}{{Domingos}, P.}, \bibinfo{author}{{Pazzani}, M.},
  \bibinfo{year}{1997}.
\newblock \bibinfo{title}{On the optimality of the simple bayesian classifier
  under zero-one loss}.
\newblock \bibinfo{journal}{Machine Learning} \bibinfo{volume}{29},
  \bibinfo{pages}{103--130}.
\newblock \URLprefix \url{https://doi.org/10.1023/A:1007413511361},
  \DOIprefix\doi{10.1023/A:1007413511361}.
\bibitem[{{Eisenhardt} et~al.(2012){Eisenhardt}, {Wu}, {Tsai}, {Assef},
  {Benford}, {Blain}, {Bridge}, {Condon}, {Cushing}, {Cutri}, {Evans},
  {Gelino}, {Griffith}, {Grillmair}, {Jarrett}, {Lonsdale}, {Masci}, {Mason},
  {Petty}, {Sayers}, {Stanford}, {Stern}, {Wright} and {Yan}}]{Eisenhardt:2012}
\bibinfo{author}{{Eisenhardt}, P.R.M.}, \bibinfo{author}{{Wu}, J.},
  \bibinfo{author}{{Tsai}, C.W.}, \bibinfo{author}{{Assef}, R.},
  \bibinfo{author}{{Benford}, D.}, \bibinfo{author}{{Blain}, A.},
  \bibinfo{author}{{Bridge}, C.}, \bibinfo{author}{{Condon}, J.J.},
  \bibinfo{author}{{Cushing}, M.C.}, \bibinfo{author}{{Cutri}, R.},
  \bibinfo{author}{{Evans}, Neal~J., I.}, \bibinfo{author}{{Gelino}, C.},
  \bibinfo{author}{{Griffith}, R.L.}, \bibinfo{author}{{Grillmair}, C.J.},
  \bibinfo{author}{{Jarrett}, T.}, \bibinfo{author}{{Lonsdale}, C.J.},
  \bibinfo{author}{{Masci}, F.J.}, \bibinfo{author}{{Mason}, B.S.},
  \bibinfo{author}{{Petty}, S.}, \bibinfo{author}{{Sayers}, J.},
  \bibinfo{author}{{Stanford}, S.A.}, \bibinfo{author}{{Stern}, D.},
  \bibinfo{author}{{Wright}, E.L.}, \bibinfo{author}{{Yan}, L.},
  \bibinfo{year}{2012}.
\newblock \bibinfo{title}{{The First Hyper-luminous Infrared Galaxy Discovered
  by WISE}}.
\newblock \bibinfo{journal}{\apj} \bibinfo{volume}{755}, \bibinfo{pages}{173}.
\newblock \DOIprefix\doi{10.1088/0004-637X/755/2/173},
  \href{http://arxiv.org/abs/1208.5517}{\tt arXiv:1208.5517}.
\bibitem[{{Eisenstein} et~al.(2001){Eisenstein}, {Annis}, {Gunn}, {Szalay},
  {Connolly}, {Nichol}, {Bahcall}, {Bernardi}, {Burles}, {Castander},
  {Fukugita}, {Hogg}, {Ivezi{\'c}}, {Knapp}, {Lupton}, {Narayanan}, {Postman},
  {Reichart}, {Richmond}, {Schneider}, {Schlegel}, {Strauss}, {SubbaRao},
  {Tucker}, {Vanden Berk}, {Vogeley}, {Weinberg} and {Yanny}}]{Eisenstein:2001}
\bibinfo{author}{{Eisenstein}, D.J.}, \bibinfo{author}{{Annis}, J.},
  \bibinfo{author}{{Gunn}, J.E.}, \bibinfo{author}{{Szalay}, A.S.},
  \bibinfo{author}{{Connolly}, A.J.}, \bibinfo{author}{{Nichol}, R.C.},
  \bibinfo{author}{{Bahcall}, N.A.}, \bibinfo{author}{{Bernardi}, M.},
  \bibinfo{author}{{Burles}, S.}, \bibinfo{author}{{Castander}, F.J.},
  \bibinfo{author}{{Fukugita}, M.}, \bibinfo{author}{{Hogg}, D.W.},
  \bibinfo{author}{{Ivezi{\'c}}, {\v Z}.}, \bibinfo{author}{{Knapp}, G.R.},
  \bibinfo{author}{{Lupton}, R.H.}, \bibinfo{author}{{Narayanan}, V.},
  \bibinfo{author}{{Postman}, M.}, \bibinfo{author}{{Reichart}, D.E.},
  \bibinfo{author}{{Richmond}, M.}, \bibinfo{author}{{Schneider}, D.P.},
  \bibinfo{author}{{Schlegel}, D.J.}, \bibinfo{author}{{Strauss}, M.A.},
  \bibinfo{author}{{SubbaRao}, M.}, \bibinfo{author}{{Tucker}, D.L.},
  \bibinfo{author}{{Vanden Berk}, D.}, \bibinfo{author}{{Vogeley}, M.S.},
  \bibinfo{author}{{Weinberg}, D.H.}, \bibinfo{author}{{Yanny}, B.},
  \bibinfo{year}{2001}.
\newblock \bibinfo{title}{{Spectroscopic Target Selection for the Sloan Digital
  Sky Survey: The Luminous Red Galaxy Sample}}.
\newblock \bibinfo{journal}{\aj} \bibinfo{volume}{122},
  \bibinfo{pages}{2267--2280}.
\newblock \DOIprefix\doi{10.1086/323717},
  \href{http://arxiv.org/abs/astro-ph/0108153}{\tt arXiv:astro-ph/0108153}.
\bibitem[{{Gaia Collaboration} et~al.(2022a){Gaia Collaboration},
  {Bailer-Jones}, {Teyssier}, {Delchambre}, {Ducourant}, {Garabato},
  {Hatzidimitriou}, {Klioner}, {Rimoldini}, {Bellas-Velidis}, {Carballo},
  {Carnerero}, {Diener}, {Fouesneau}, {Galluccio}, {Gavras}, {Krone-Martins},
  {Raiteri}, {Teixeira}, {Brown}, {Vallenari}, {Prusti}, {de Bruijne},
  {Arenou}, {Babusiaux}, {Biermann}, {Creevey}, {Evans}, {Eyer}, {Guerra},
  {Hutton}, {Jordi}, {Lammers}, {Lindegren}, {Luri}, {Mignard}, {Panem},
  {Pourbaix}, {Randich}, {Sartoretti}, {Soubiran}, {Tanga}, {Walton},
  {Bastian}, {Drimmel}, {Jansen}, {Katz}, {Lattanzi}, {van Leeuwen}, {Bakker},
  {Cacciari}, {Casta{\~n}eda}, {De Angeli}, {Fabricius}, {Fr{\'e}mat},
  {Guerrier}, {Heiter}, {Masana}, {Messineo}, {Mowlavi}, {Nicolas},
  {Nienartowicz}, {Pailler}, {Panuzzo}, {Riclet}, {Roux}, {Seabroke}, {Sordo},
  {Th{\'e}venin}, {Gracia-Abril}, {Portell}, {Altmann}, {Andrae}, {Audard},
  {Benson}, {Berthier}, {Blomme}, {Burgess}, {Busonero}, {Busso},
  {C{\'a}novas}, {Carry}, {Cellino}, {Cheek}, {Clementini}, {Damerdji},
  {Davidson}, {de Teodoro}, {Nu{\~n}ez Campos}, {Dell'Oro}, {Esquej},
  {Fern{\'a}ndez-Hern{\'a}ndez}, {Fraile}, {Garc{\'\i}a-Lario}, {Gosset},
  {Haigron}, {Halbwachs}, {Hambly}, {Harrison}, {Hern{\'a}ndez}, {Hestroffer},
  {Hodgkin}, {Holl}, {Jan{\ss}en}, {Jevardat de Fombelle}, {Jordan},
  {Lanzafame}, {L{\"o}ffler}, {Marchal}, {Marrese}, {Moitinho}, {Muinonen},
  {Osborne}, {Pancino}, {Pauwels}, {Recio-Blanco}, {Reyl{\'e}}, {Riello},
  {Roegiers}, {Rybizki}, {Sarro}, {Siopis}, {Smith}, {Sozzetti}, {Utrilla},
  {van Leeuwen}, {Abbas}, {{\'A}brah{\'a}m}, {Abreu Aramburu}, {Aerts},
  {Aguado}, {Ajaj}, {Aldea-Montero}, {Altavilla}, {{\'A}lvarez}, {Alves},
  {Anderson}, {Anglada Varela}, {Antoja}, {Baines}, {Baker},
  {Balaguer-N{\'u}{\~n}ez}, {Balbinot}, {Balog}, {Barache}, {Barbato},
  {Barros}, {Barstow}, {Bartolom{\'e}}, {Bassilana}, {Bauchet}, {Becciani},
  {Bellazzini}, {Berihuete}, {Bernet}, {Bertone}, {Bianchi}, {Binnenfeld},
  {Blanco-Cuaresma}, {Boch}, {Bombrun}, {Bossini}, {Bouquillon}, {Bragaglia},
  {Bramante}, {Breedt}, {Bressan}, {Brouillet}, {Brugaletta}, {Bucciarelli},
  {Burlacu}, {Butkevich}, {Buzzi}, {Caffau}, {Cancelliere}, {Cantat-Gaudin},
  {Carlucci}, {Carrasco}, {Casamiquela}, {Castellani}, {Castro-Ginard},
  {Chaoul}, {Charlot}, {Chemin}, {Chiaramida}, {Chiavassa}, {Chornay},
  {Comoretto}, {Contursi}, {Cooper}, {Cornez}, {Cowell}, {Crifo}, {Cropper},
  {Crosta}, {Crowley}, {Dafonte}, {Dapergolas}, {David}, {de Laverny}, {De
  Luise}, {De March}, {De Ridder}, {de Souza}, {de Torres}, {del Peloso}, {del
  Pozo}, {Delbo}, {Delgado}, {Delisle}, {Demouchy}, {Dharmawardena}, {Diakite},
  {Distefano}, {Dolding}, {Enke}, {Fabre}, {Fabrizio}, {Faigler}, {Fedorets},
  {Fernique}, {Figueras}, {Fournier}, {Fouron}, {Fragkoudi}, {Gai},
  {Garcia-Gutierrez}, {Garcia-Reinaldos}, {Garc{\'\i}a-Torres}, {Garofalo},
  {Gavel}, {Gerlach}, {Geyer}, {Giacobbe}, {Gilmore}, {Girona}, {Giuffrida},
  {Gomel}, {Gomez}, {Gonz{\'a}lez-N{\'u}{\~n}ez},
  {Gonz{\'a}lez-Santamar{\'\i}a}, {Gonz{\'a}lez-Vidal}, {Granvik}, {Guillout},
  {Guiraud}, {Guti{\'e}rrez-S{\'a}nchez}, {Guy}, {Hauser}, {Haywood}, {Helmer},
  {Helmi}, {Sarmiento}, {Hidalgo}, {H{\l}adczuk}, {Hobbs}, {Holland}, {Huckle},
  {Jardine}, {Jasniewicz}, {Jean-Antoine Piccolo}, {Jim{\'e}nez-Arranz},
  {Juaristi Campillo}, {Julbe}, {Karbevska}, {Kervella}, {Khanna}, {Kontizas},
  {Kordopatis}, {Korn}, {K{\'o}sp{\'a}l}, {Kostrzewa-Rutkowska},
  {Kruszy{\'n}ska}, {Kun}, {Laizeau}, {Lambert}, {Lanza}, {Lasne}, {Le
  Campion}, {Lebreton}, {Lebzelter}, {Leccia}, {Leclerc}, {Lecoeur-Taibi},
  {Liao}, {Licata}, {Lindstr{\o}m}, {Lister}, {Livanou}, {Lobel}, {Lorca},
  {Loup}, {Madrero Pardo}, {Magdaleno Romeo}, {Managau}, {Mann}, {Manteiga},
  {Marchant}, {Marconi}, {Marcos}, {Marcos Santos}, {Mar{\'\i}n Pina},
  {Marinoni}, {Marocco}, {Marshall}, {Polo}, {Mart{\'\i}n-Fleitas}, {Marton},
  {Mary}, {Masip}, {Massari}, {Mastrobuono-Battisti}, {Mazeh}, {McMillan},
  {Messina}, {Michalik}, {Millar}, {Mints}, {Molina}, {Molinaro}, {Moln{\'a}r},
  {Monari}, {Mongui{\'o}}, {Montegriffo}, {Montero}, {Mor}, {Mora},
  {Morbidelli}, {Morel}, {Morris}, {Muraveva}, {Murphy}, {Musella}, {Nagy},
  {Noval}, {Oca{\~n}a}, {Ogden}, {Ordenovic}, {Osinde}, {Pagani}, {Pagano},
  {Palaversa}, {Palicio}, {Pallas-Quintela}, {Panahi}, {Payne-Wardenaar},
  {Pe{\~n}alosa Esteller}, {Penttil{\"a}}, {Pichon}, {Piersimoni}, {Pineau},
  {Plachy}, {Plum}, {Poggio}, {Pr{\v{s}}a}, {Pulone}, {Racero}, {Ragaini},
  {Rainer}, {Ramos}, {Ramos-Lerate}, {Re Fiorentin}, {Regibo}, {Richards},
  {Rios Diaz}, {Ripepi}, {Riva}, {Rix}, {Rixon}, {Robichon}, {Robin}, {Robin},
  {Roelens}, {Rogues}, {Rohrbasser}, {Romero-G{\'o}mez}, {Rowell}, {Royer},
  {Ruz Mieres}, {Rybicki}, {Sadowski}, {S{\'a}ez N{\'u}{\~n}ez}, {Sagrist{\`a}
  Sell{\'e}s}, {Sahlmann}, {Salguero}, {Samaras}, {Sanchez Gimenez}, {Sanna},
  {Santove{\~n}a}, {Sarasso}, {Schultheis}, {Sciacca}, {Segol}, {Segovia},
  {S{\'e}gransan}, {Semeux}, {Shahaf}, {Siddiqui}, {Siebert}, {Siltala},
  {Silvelo}, {Slezak}, {Slezak}, {Smart}, {Snaith}, {Solano}, {Solitro},
  {Souami}, {Souchay}, {Spagna}, {Spina}, {Spoto}, {Steele},
  {Steidelm{\"u}ller}, {Stephenson}, {S{\"u}veges}, {Surdej}, {Szabados},
  {Szegedi-Elek}, {Taris}, {Taylor}, {Tolomei}, {Tonello}, {Torra}, {Torra},
  {Torralba Elipe}, {Trabucchi}, {Tsounis}, {Turon}, {Ulla}, {Unger},
  {Vaillant}, {van Dillen}, {van Reeven}, {Vanel}, {Vecchiato}, {Viala},
  {Vicente}, {Voutsinas}, {Weiler}, {Wevers}, {Wyrzykowski}, {Yoldas}, {Yvard},
  {Zhao}, {Zorec}, {Zucker} and {Zwitter}}]{GaiaExgal:2022}
\bibinfo{author}{{Gaia Collaboration}}, \bibinfo{author}{{Bailer-Jones},
  C.A.L.}, \bibinfo{author}{{Teyssier}, D.}, \bibinfo{author}{{Delchambre},
  L.}, \bibinfo{author}{{Ducourant}, C.}, \bibinfo{author}{{Garabato}, D.},
  \bibinfo{author}{{Hatzidimitriou}, D.}, \bibinfo{author}{{Klioner}, S.A.},
  \bibinfo{author}{{Rimoldini}, L.}, \bibinfo{author}{{Bellas-Velidis}, I.},
  \bibinfo{author}{{Carballo}, R.}, \bibinfo{author}{{Carnerero}, M.I.},
  \bibinfo{author}{{Diener}, C.}, \bibinfo{author}{{Fouesneau}, M.},
  \bibinfo{author}{{Galluccio}, L.}, \bibinfo{author}{{Gavras}, P.},
  \bibinfo{author}{{Krone-Martins}, A.}, \bibinfo{author}{{Raiteri}, C.M.},
  \bibinfo{author}{{Teixeira}, R.}, \bibinfo{author}{{Brown}, A.G.A.},
  \bibinfo{author}{{Vallenari}, A.}, \bibinfo{author}{{Prusti}, T.},
  \bibinfo{author}{{de Bruijne}, J.H.J.}, \bibinfo{author}{{Arenou}, F.},
  \bibinfo{author}{{Babusiaux}, C.}, \bibinfo{author}{{Biermann}, M.},
  \bibinfo{author}{{Creevey}, O.L.}, \bibinfo{author}{{Evans}, D.W.},
  \bibinfo{author}{{Eyer}, L.}, \bibinfo{author}{{Guerra}, R.},
  \bibinfo{author}{{Hutton}, A.}, \bibinfo{author}{{Jordi}, C.},
  \bibinfo{author}{{Lammers}, U.L.}, \bibinfo{author}{{Lindegren}, L.},
  \bibinfo{author}{{Luri}, X.}, \bibinfo{author}{{Mignard}, F.},
  \bibinfo{author}{{Panem}, C.}, \bibinfo{author}{{Pourbaix}, D.},
  \bibinfo{author}{{Randich}, S.}, \bibinfo{author}{{Sartoretti}, P.},
  \bibinfo{author}{{Soubiran}, C.}, \bibinfo{author}{{Tanga}, P.},
  \bibinfo{author}{{Walton}, N.A.}, \bibinfo{author}{{Bastian}, U.},
  \bibinfo{author}{{Drimmel}, R.}, \bibinfo{author}{{Jansen}, F.},
  \bibinfo{author}{{Katz}, D.}, \bibinfo{author}{{Lattanzi}, M.G.},
  \bibinfo{author}{{van Leeuwen}, F.}, \bibinfo{author}{{Bakker}, J.},
  \bibinfo{author}{{Cacciari}, C.}, \bibinfo{author}{{Casta{\~n}eda}, J.},
  \bibinfo{author}{{De Angeli}, F.}, \bibinfo{author}{{Fabricius}, C.},
  \bibinfo{author}{{Fr{\'e}mat}, Y.}, \bibinfo{author}{{Guerrier}, A.},
  \bibinfo{author}{{Heiter}, U.}, \bibinfo{author}{{Masana}, E.},
  \bibinfo{author}{{Messineo}, R.}, \bibinfo{author}{{Mowlavi}, N.},
  \bibinfo{author}{{Nicolas}, C.}, \bibinfo{author}{{Nienartowicz}, K.},
  \bibinfo{author}{{Pailler}, F.}, \bibinfo{author}{{Panuzzo}, P.},
  \bibinfo{author}{{Riclet}, F.}, \bibinfo{author}{{Roux}, W.},
  \bibinfo{author}{{Seabroke}, G.M.}, \bibinfo{author}{{Sordo}, R.},
  \bibinfo{author}{{Th{\'e}venin}, F.}, \bibinfo{author}{{Gracia-Abril}, G.},
  \bibinfo{author}{{Portell}, J.}, \bibinfo{author}{{Altmann}, M.},
  \bibinfo{author}{{Andrae}, R.}, \bibinfo{author}{{Audard}, M.},
  \bibinfo{author}{{Benson}, K.}, \bibinfo{author}{{Berthier}, J.},
  \bibinfo{author}{{Blomme}, R.}, \bibinfo{author}{{Burgess}, P.W.},
  \bibinfo{author}{{Busonero}, D.}, \bibinfo{author}{{Busso}, G.},
  \bibinfo{author}{{C{\'a}novas}, H.}, \bibinfo{author}{{Carry}, B.},
  \bibinfo{author}{{Cellino}, A.}, \bibinfo{author}{{Cheek}, N.},
  \bibinfo{author}{{Clementini}, G.}, \bibinfo{author}{{Damerdji}, Y.},
  \bibinfo{author}{{Davidson}, M.}, \bibinfo{author}{{de Teodoro}, P.},
  \bibinfo{author}{{Nu{\~n}ez Campos}, M.}, \bibinfo{author}{{Dell'Oro}, A.},
  \bibinfo{author}{{Esquej}, P.},
  \bibinfo{author}{{Fern{\'a}ndez-Hern{\'a}ndez}, J.},
  \bibinfo{author}{{Fraile}, E.}, \bibinfo{author}{{Garc{\'\i}a-Lario}, P.},
  \bibinfo{author}{{Gosset}, E.}, \bibinfo{author}{{Haigron}, R.},
  \bibinfo{author}{{Halbwachs}, J.L.}, \bibinfo{author}{{Hambly}, N.C.},
  \bibinfo{author}{{Harrison}, D.L.}, \bibinfo{author}{{Hern{\'a}ndez}, J.},
  \bibinfo{author}{{Hestroffer}, D.}, \bibinfo{author}{{Hodgkin}, S.T.},
  \bibinfo{author}{{Holl}, B.}, \bibinfo{author}{{Jan{\ss}en}, K.},
  \bibinfo{author}{{Jevardat de Fombelle}, G.}, \bibinfo{author}{{Jordan}, S.},
  \bibinfo{author}{{Lanzafame}, A.C.}, \bibinfo{author}{{L{\"o}ffler}, W.},
  \bibinfo{author}{{Marchal}, O.}, \bibinfo{author}{{Marrese}, P.M.},
  \bibinfo{author}{{Moitinho}, A.}, \bibinfo{author}{{Muinonen}, K.},
  \bibinfo{author}{{Osborne}, P.}, \bibinfo{author}{{Pancino}, E.},
  \bibinfo{author}{{Pauwels}, T.}, \bibinfo{author}{{Recio-Blanco}, A.},
  \bibinfo{author}{{Reyl{\'e}}, C.}, \bibinfo{author}{{Riello}, M.},
  \bibinfo{author}{{Roegiers}, T.}, \bibinfo{author}{{Rybizki}, J.},
  \bibinfo{author}{{Sarro}, L.M.}, \bibinfo{author}{{Siopis}, C.},
  \bibinfo{author}{{Smith}, M.}, \bibinfo{author}{{Sozzetti}, A.},
  \bibinfo{author}{{Utrilla}, E.}, \bibinfo{author}{{van Leeuwen}, M.},
  \bibinfo{author}{{Abbas}, U.}, \bibinfo{author}{{{\'A}brah{\'a}m}, P.},
  \bibinfo{author}{{Abreu Aramburu}, A.}, \bibinfo{author}{{Aerts}, C.},
  \bibinfo{author}{{Aguado}, J.J.}, \bibinfo{author}{{Ajaj}, M.},
  \bibinfo{author}{{Aldea-Montero}, F.}, \bibinfo{author}{{Altavilla}, G.},
  \bibinfo{author}{{{\'A}lvarez}, M.A.}, \bibinfo{author}{{Alves}, J.},
  \bibinfo{author}{{Anderson}, R.I.}, \bibinfo{author}{{Anglada Varela}, E.},
  \bibinfo{author}{{Antoja}, T.}, \bibinfo{author}{{Baines}, D.},
  \bibinfo{author}{{Baker}, S.G.}, \bibinfo{author}{{Balaguer-N{\'u}{\~n}ez},
  L.}, \bibinfo{author}{{Balbinot}, E.}, \bibinfo{author}{{Balog}, Z.},
  \bibinfo{author}{{Barache}, C.}, \bibinfo{author}{{Barbato}, D.},
  \bibinfo{author}{{Barros}, M.}, \bibinfo{author}{{Barstow}, M.A.},
  \bibinfo{author}{{Bartolom{\'e}}, S.}, \bibinfo{author}{{Bassilana}, J.L.},
  \bibinfo{author}{{Bauchet}, N.}, \bibinfo{author}{{Becciani}, U.},
  \bibinfo{author}{{Bellazzini}, M.}, \bibinfo{author}{{Berihuete}, A.},
  \bibinfo{author}{{Bernet}, M.}, \bibinfo{author}{{Bertone}, S.},
  \bibinfo{author}{{Bianchi}, L.}, \bibinfo{author}{{Binnenfeld}, A.},
  \bibinfo{author}{{Blanco-Cuaresma}, S.}, \bibinfo{author}{{Boch}, T.},
  \bibinfo{author}{{Bombrun}, A.}, \bibinfo{author}{{Bossini}, D.},
  \bibinfo{author}{{Bouquillon}, S.}, \bibinfo{author}{{Bragaglia}, A.},
  \bibinfo{author}{{Bramante}, L.}, \bibinfo{author}{{Breedt}, E.},
  \bibinfo{author}{{Bressan}, A.}, \bibinfo{author}{{Brouillet}, N.},
  \bibinfo{author}{{Brugaletta}, E.}, \bibinfo{author}{{Bucciarelli}, B.},
  \bibinfo{author}{{Burlacu}, A.}, \bibinfo{author}{{Butkevich}, A.G.},
  \bibinfo{author}{{Buzzi}, R.}, \bibinfo{author}{{Caffau}, E.},
  \bibinfo{author}{{Cancelliere}, R.}, \bibinfo{author}{{Cantat-Gaudin}, T.},
  \bibinfo{author}{{Carlucci}, T.}, \bibinfo{author}{{Carrasco}, J.M.},
  \bibinfo{author}{{Casamiquela}, L.}, \bibinfo{author}{{Castellani}, M.},
  \bibinfo{author}{{Castro-Ginard}, A.}, \bibinfo{author}{{Chaoul}, L.},
  \bibinfo{author}{{Charlot}, P.}, \bibinfo{author}{{Chemin}, L.},
  \bibinfo{author}{{Chiaramida}, V.}, \bibinfo{author}{{Chiavassa}, A.},
  \bibinfo{author}{{Chornay}, N.}, \bibinfo{author}{{Comoretto}, G.},
  \bibinfo{author}{{Contursi}, G.}, \bibinfo{author}{{Cooper}, W.J.},
  \bibinfo{author}{{Cornez}, T.}, \bibinfo{author}{{Cowell}, S.},
  \bibinfo{author}{{Crifo}, F.}, \bibinfo{author}{{Cropper}, M.},
  \bibinfo{author}{{Crosta}, M.}, \bibinfo{author}{{Crowley}, C.},
  \bibinfo{author}{{Dafonte}, C.}, \bibinfo{author}{{Dapergolas}, A.},
  \bibinfo{author}{{David}, P.}, \bibinfo{author}{{de Laverny}, P.},
  \bibinfo{author}{{De Luise}, F.}, \bibinfo{author}{{De March}, R.},
  \bibinfo{author}{{De Ridder}, J.}, \bibinfo{author}{{de Souza}, R.},
  \bibinfo{author}{{de Torres}, A.}, \bibinfo{author}{{del Peloso}, E.F.},
  \bibinfo{author}{{del Pozo}, E.}, \bibinfo{author}{{Delbo}, M.},
  \bibinfo{author}{{Delgado}, A.}, \bibinfo{author}{{Delisle}, J.B.},
  \bibinfo{author}{{Demouchy}, C.}, \bibinfo{author}{{Dharmawardena}, T.E.},
  \bibinfo{author}{{Diakite}, S.}, \bibinfo{author}{{Distefano}, E.},
  \bibinfo{author}{{Dolding}, C.}, \bibinfo{author}{{Enke}, H.},
  \bibinfo{author}{{Fabre}, C.}, \bibinfo{author}{{Fabrizio}, M.},
  \bibinfo{author}{{Faigler}, S.}, \bibinfo{author}{{Fedorets}, G.},
  \bibinfo{author}{{Fernique}, P.}, \bibinfo{author}{{Figueras}, F.},
  \bibinfo{author}{{Fournier}, Y.}, \bibinfo{author}{{Fouron}, C.},
  \bibinfo{author}{{Fragkoudi}, F.}, \bibinfo{author}{{Gai}, M.},
  \bibinfo{author}{{Garcia-Gutierrez}, A.},
  \bibinfo{author}{{Garcia-Reinaldos}, M.},
  \bibinfo{author}{{Garc{\'\i}a-Torres}, M.}, \bibinfo{author}{{Garofalo}, A.},
  \bibinfo{author}{{Gavel}, A.}, \bibinfo{author}{{Gerlach}, E.},
  \bibinfo{author}{{Geyer}, R.}, \bibinfo{author}{{Giacobbe}, P.},
  \bibinfo{author}{{Gilmore}, G.}, \bibinfo{author}{{Girona}, S.},
  \bibinfo{author}{{Giuffrida}, G.}, \bibinfo{author}{{Gomel}, R.},
  \bibinfo{author}{{Gomez}, A.}, \bibinfo{author}{{Gonz{\'a}lez-N{\'u}{\~n}ez},
  J.}, \bibinfo{author}{{Gonz{\'a}lez-Santamar{\'\i}a}, I.},
  \bibinfo{author}{{Gonz{\'a}lez-Vidal}, J.J.}, \bibinfo{author}{{Granvik},
  M.}, \bibinfo{author}{{Guillout}, P.}, \bibinfo{author}{{Guiraud}, J.},
  \bibinfo{author}{{Guti{\'e}rrez-S{\'a}nchez}, R.}, \bibinfo{author}{{Guy},
  L.P.}, \bibinfo{author}{{Hauser}, M.}, \bibinfo{author}{{Haywood}, M.},
  \bibinfo{author}{{Helmer}, A.}, \bibinfo{author}{{Helmi}, A.},
  \bibinfo{author}{{Sarmiento}, M.H.}, \bibinfo{author}{{Hidalgo}, S.L.},
  \bibinfo{author}{{H{\l}adczuk}, N.}, \bibinfo{author}{{Hobbs}, D.},
  \bibinfo{author}{{Holland}, G.}, \bibinfo{author}{{Huckle}, H.E.},
  \bibinfo{author}{{Jardine}, K.}, \bibinfo{author}{{Jasniewicz}, G.},
  \bibinfo{author}{{Jean-Antoine Piccolo}, A.},
  \bibinfo{author}{{Jim{\'e}nez-Arranz}, {\'O}.}, \bibinfo{author}{{Juaristi
  Campillo}, J.}, \bibinfo{author}{{Julbe}, F.}, \bibinfo{author}{{Karbevska},
  L.}, \bibinfo{author}{{Kervella}, P.}, \bibinfo{author}{{Khanna}, S.},
  \bibinfo{author}{{Kontizas}, M.}, \bibinfo{author}{{Kordopatis}, G.},
  \bibinfo{author}{{Korn}, A.J.}, \bibinfo{author}{{K{\'o}sp{\'a}l}, {\'A}.},
  \bibinfo{author}{{Kostrzewa-Rutkowska}, Z.},
  \bibinfo{author}{{Kruszy{\'n}ska}, K.}, \bibinfo{author}{{Kun}, M.},
  \bibinfo{author}{{Laizeau}, P.}, \bibinfo{author}{{Lambert}, S.},
  \bibinfo{author}{{Lanza}, A.F.}, \bibinfo{author}{{Lasne}, Y.},
  \bibinfo{author}{{Le Campion}, J.F.}, \bibinfo{author}{{Lebreton}, Y.},
  \bibinfo{author}{{Lebzelter}, T.}, \bibinfo{author}{{Leccia}, S.},
  \bibinfo{author}{{Leclerc}, N.}, \bibinfo{author}{{Lecoeur-Taibi}, I.},
  \bibinfo{author}{{Liao}, S.}, \bibinfo{author}{{Licata}, E.L.},
  \bibinfo{author}{{Lindstr{\o}m}, H.E.P.}, \bibinfo{author}{{Lister}, T.A.},
  \bibinfo{author}{{Livanou}, E.}, \bibinfo{author}{{Lobel}, A.},
  \bibinfo{author}{{Lorca}, A.}, \bibinfo{author}{{Loup}, C.},
  \bibinfo{author}{{Madrero Pardo}, P.}, \bibinfo{author}{{Magdaleno Romeo},
  A.}, \bibinfo{author}{{Managau}, S.}, \bibinfo{author}{{Mann}, R.G.},
  \bibinfo{author}{{Manteiga}, M.}, \bibinfo{author}{{Marchant}, J.M.},
  \bibinfo{author}{{Marconi}, M.}, \bibinfo{author}{{Marcos}, J.},
  \bibinfo{author}{{Marcos Santos}, M.M.S.}, \bibinfo{author}{{Mar{\'\i}n
  Pina}, D.}, \bibinfo{author}{{Marinoni}, S.}, \bibinfo{author}{{Marocco},
  F.}, \bibinfo{author}{{Marshall}, D.J.}, \bibinfo{author}{{Polo}, L.M.},
  \bibinfo{author}{{Mart{\'\i}n-Fleitas}, J.M.}, \bibinfo{author}{{Marton},
  G.}, \bibinfo{author}{{Mary}, N.}, \bibinfo{author}{{Masip}, A.},
  \bibinfo{author}{{Massari}, D.}, \bibinfo{author}{{Mastrobuono-Battisti},
  A.}, \bibinfo{author}{{Mazeh}, T.}, \bibinfo{author}{{McMillan}, P.J.},
  \bibinfo{author}{{Messina}, S.}, \bibinfo{author}{{Michalik}, D.},
  \bibinfo{author}{{Millar}, N.R.}, \bibinfo{author}{{Mints}, A.},
  \bibinfo{author}{{Molina}, D.}, \bibinfo{author}{{Molinaro}, R.},
  \bibinfo{author}{{Moln{\'a}r}, L.}, \bibinfo{author}{{Monari}, G.},
  \bibinfo{author}{{Mongui{\'o}}, M.}, \bibinfo{author}{{Montegriffo}, P.},
  \bibinfo{author}{{Montero}, A.}, \bibinfo{author}{{Mor}, R.},
  \bibinfo{author}{{Mora}, A.}, \bibinfo{author}{{Morbidelli}, R.},
  \bibinfo{author}{{Morel}, T.}, \bibinfo{author}{{Morris}, D.},
  \bibinfo{author}{{Muraveva}, T.}, \bibinfo{author}{{Murphy}, C.P.},
  \bibinfo{author}{{Musella}, I.}, \bibinfo{author}{{Nagy}, Z.},
  \bibinfo{author}{{Noval}, L.}, \bibinfo{author}{{Oca{\~n}a}, F.},
  \bibinfo{author}{{Ogden}, A.}, \bibinfo{author}{{Ordenovic}, C.},
  \bibinfo{author}{{Osinde}, J.O.}, \bibinfo{author}{{Pagani}, C.},
  \bibinfo{author}{{Pagano}, I.}, \bibinfo{author}{{Palaversa}, L.},
  \bibinfo{author}{{Palicio}, P.A.}, \bibinfo{author}{{Pallas-Quintela}, L.},
  \bibinfo{author}{{Panahi}, A.}, \bibinfo{author}{{Payne-Wardenaar}, S.},
  \bibinfo{author}{{Pe{\~n}alosa Esteller}, X.},
  \bibinfo{author}{{Penttil{\"a}}, A.}, \bibinfo{author}{{Pichon}, B.},
  \bibinfo{author}{{Piersimoni}, A.M.}, \bibinfo{author}{{Pineau}, F.X.},
  \bibinfo{author}{{Plachy}, E.}, \bibinfo{author}{{Plum}, G.},
  \bibinfo{author}{{Poggio}, E.}, \bibinfo{author}{{Pr{\v{s}}a}, A.},
  \bibinfo{author}{{Pulone}, L.}, \bibinfo{author}{{Racero}, E.},
  \bibinfo{author}{{Ragaini}, S.}, \bibinfo{author}{{Rainer}, M.},
  \bibinfo{author}{{Ramos}, P.}, \bibinfo{author}{{Ramos-Lerate}, M.},
  \bibinfo{author}{{Re Fiorentin}, P.}, \bibinfo{author}{{Regibo}, S.},
  \bibinfo{author}{{Richards}, P.J.}, \bibinfo{author}{{Rios Diaz}, C.},
  \bibinfo{author}{{Ripepi}, V.}, \bibinfo{author}{{Riva}, A.},
  \bibinfo{author}{{Rix}, H.W.}, \bibinfo{author}{{Rixon}, G.},
  \bibinfo{author}{{Robichon}, N.}, \bibinfo{author}{{Robin}, A.C.},
  \bibinfo{author}{{Robin}, C.}, \bibinfo{author}{{Roelens}, M.},
  \bibinfo{author}{{Rogues}, H.R.O.}, \bibinfo{author}{{Rohrbasser}, L.},
  \bibinfo{author}{{Romero-G{\'o}mez}, M.}, \bibinfo{author}{{Rowell}, N.},
  \bibinfo{author}{{Royer}, F.}, \bibinfo{author}{{Ruz Mieres}, D.},
  \bibinfo{author}{{Rybicki}, K.A.}, \bibinfo{author}{{Sadowski}, G.},
  \bibinfo{author}{{S{\'a}ez N{\'u}{\~n}ez}, A.},
  \bibinfo{author}{{Sagrist{\`a} Sell{\'e}s}, A.}, \bibinfo{author}{{Sahlmann},
  J.}, \bibinfo{author}{{Salguero}, E.}, \bibinfo{author}{{Samaras}, N.},
  \bibinfo{author}{{Sanchez Gimenez}, V.}, \bibinfo{author}{{Sanna}, N.},
  \bibinfo{author}{{Santove{\~n}a}, R.}, \bibinfo{author}{{Sarasso}, M.},
  \bibinfo{author}{{Schultheis}, M.S.}, \bibinfo{author}{{Sciacca}, E.},
  \bibinfo{author}{{Segol}, M.}, \bibinfo{author}{{Segovia}, J.C.},
  \bibinfo{author}{{S{\'e}gransan}, D.}, \bibinfo{author}{{Semeux}, D.},
  \bibinfo{author}{{Shahaf}, S.}, \bibinfo{author}{{Siddiqui}, H.I.},
  \bibinfo{author}{{Siebert}, A.}, \bibinfo{author}{{Siltala}, L.},
  \bibinfo{author}{{Silvelo}, A.}, \bibinfo{author}{{Slezak}, E.},
  \bibinfo{author}{{Slezak}, I.}, \bibinfo{author}{{Smart}, R.L.},
  \bibinfo{author}{{Snaith}, O.N.}, \bibinfo{author}{{Solano}, E.},
  \bibinfo{author}{{Solitro}, F.}, \bibinfo{author}{{Souami}, D.},
  \bibinfo{author}{{Souchay}, J.}, \bibinfo{author}{{Spagna}, A.},
  \bibinfo{author}{{Spina}, L.}, \bibinfo{author}{{Spoto}, F.},
  \bibinfo{author}{{Steele}, I.A.}, \bibinfo{author}{{Steidelm{\"u}ller}, H.},
  \bibinfo{author}{{Stephenson}, C.A.}, \bibinfo{author}{{S{\"u}veges}, M.},
  \bibinfo{author}{{Surdej}, J.}, \bibinfo{author}{{Szabados}, L.},
  \bibinfo{author}{{Szegedi-Elek}, E.}, \bibinfo{author}{{Taris}, F.},
  \bibinfo{author}{{Taylor}, M.B.}, \bibinfo{author}{{Tolomei}, L.},
  \bibinfo{author}{{Tonello}, N.}, \bibinfo{author}{{Torra}, F.},
  \bibinfo{author}{{Torra}, J.}, \bibinfo{author}{{Torralba Elipe}, G.},
  \bibinfo{author}{{Trabucchi}, M.}, \bibinfo{author}{{Tsounis}, A.T.},
  \bibinfo{author}{{Turon}, C.}, \bibinfo{author}{{Ulla}, A.},
  \bibinfo{author}{{Unger}, N.}, \bibinfo{author}{{Vaillant}, M.V.},
  \bibinfo{author}{{van Dillen}, E.}, \bibinfo{author}{{van Reeven}, W.},
  \bibinfo{author}{{Vanel}, O.}, \bibinfo{author}{{Vecchiato}, A.},
  \bibinfo{author}{{Viala}, Y.}, \bibinfo{author}{{Vicente}, D.},
  \bibinfo{author}{{Voutsinas}, S.}, \bibinfo{author}{{Weiler}, M.},
  \bibinfo{author}{{Wevers}, T.}, \bibinfo{author}{{Wyrzykowski}, {\L}.},
  \bibinfo{author}{{Yoldas}, A.}, \bibinfo{author}{{Yvard}, P.},
  \bibinfo{author}{{Zhao}, H.}, \bibinfo{author}{{Zorec}, J.},
  \bibinfo{author}{{Zucker}, S.}, \bibinfo{author}{{Zwitter}, T.},
  \bibinfo{year}{2022}a.
\newblock \bibinfo{title}{{Gaia Data Release 3: The extragalactic content}}.
\newblock \bibinfo{journal}{arXiv e-prints} ,
  \bibinfo{pages}{arXiv:2206.05681}\href{http://arxiv.org/abs/2206.05681}{\tt
  arXiv:2206.05681}.
\bibitem[{{Gaia Collaboration} et~al.(2018){Gaia Collaboration}, {Brown},
  {Vallenari}, {Prusti}, {de Bruijne}, {Babusiaux}, {Bailer-Jones}, {Biermann},
  {Evans}, {Eyer}, {Jansen}, {Jordi}, {Klioner}, {Lammers}, {Lindegren},
  {Luri}, {Mignard}, {Panem}, {Pourbaix}, {Randich}, {Sartoretti}, {Siddiqui},
  {Soubiran}, {van Leeuwen}, {Walton}, {Arenou}, {Bastian}, {Cropper},
  {Drimmel}, {Katz}, {Lattanzi}, {Bakker}, {Cacciari}, {Casta{\~n}eda},
  {Chaoul}, {Cheek}, {De Angeli}, {Fabricius}, {Guerra}, {Holl}, {Masana},
  {Messineo}, {Mowlavi}, {Nienartowicz}, {Panuzzo}, {Portell}, {Riello},
  {Seabroke}, {Tanga}, {Th{\'e}venin}, {Gracia-Abril}, {Comoretto},
  {Garcia-Reinaldos}, {Teyssier}, {Altmann}, {Andrae}, {Audard},
  {Bellas-Velidis}, {Benson}, {Berthier}, {Blomme}, {Burgess}, {Busso},
  {Carry}, {Cellino}, {Clementini}, {Clotet}, {Creevey}, {Davidson}, {De
  Ridder}, {Delchambre}, {Dell'Oro}, {Ducourant},
  {Fern{\'a}ndez-Hern{\'a}ndez}, {Fouesneau}, {Fr{\'e}mat}, {Galluccio},
  {Garc{\'\i}a-Torres}, {Gonz{\'a}lez-N{\'u}{\~n}ez}, {Gonz{\'a}lez-Vidal},
  {Gosset}, {Guy}, {Halbwachs}, {Hambly}, {Harrison}, {Hern{\'a}ndez},
  {Hestroffer}, {Hodgkin}, {Hutton}, {Jasniewicz}, {Jean-Antoine-Piccolo},
  {Jordan}, {Korn}, {Krone-Martins}, {Lanzafame}, {Lebzelter}, {L{\"o}ffler},
  {Manteiga}, {Marrese}, {Mart{\'\i}n-Fleitas}, {Moitinho}, {Mora}, {Muinonen},
  {Osinde}, {Pancino}, {Pauwels}, {Petit}, {Recio-Blanco}, {Richards},
  {Rimoldini}, {Robin}, {Sarro}, {Siopis}, {Smith}, {Sozzetti}, {S{\"u}veges},
  {Torra}, {van Reeven}, {Abbas}, {Abreu Aramburu}, {Accart}, {Aerts},
  {Altavilla}, {{\'A}lvarez}, {Alvarez}, {Alves}, {Anderson}, {Andrei},
  {Anglada Varela}, {Antiche}, {Antoja}, {Arcay}, {Astraatmadja}, {Bach},
  {Baker}, {Balaguer-N{\'u}{\~n}ez}, {Balm}, {Barache}, {Barata}, {Barbato},
  {Barblan}, {Barklem}, {Barrado}, {Barros}, {Barstow}, {Bartholom{\'e}
  Mu{\~n}oz}, {Bassilana}, {Becciani}, {Bellazzini}, {Berihuete}, {Bertone},
  {Bianchi}, {Bienaym{\'e}}, {Blanco-Cuaresma}, {Boch}, {Boeche}, {Bombrun},
  {Borrachero}, {Bossini}, {Bouquillon}, {Bourda}, {Bragaglia}, {Bramante},
  {Breddels}, {Bressan}, {Brouillet}, {Br{\"u}semeister}, {Brugaletta},
  {Bucciarelli}, {Burlacu}, {Busonero}, {Butkevich}, {Buzzi}, {Caffau},
  {Cancelliere}, {Cannizzaro}, {Cantat-Gaudin}, {Carballo}, {Carlucci},
  {Carrasco}, {Casamiquela}, {Castellani}, {Castro-Ginard}, {Charlot},
  {Chemin}, {Chiavassa}, {Cocozza}, {Costigan}, {Cowell}, {Crifo}, {Crosta},
  {Crowley}, {Cuypers}, {Dafonte}, {Damerdji}, {Dapergolas}, {David}, {David},
  {de Laverny}, {De Luise}, {De March}, {de Martino}, {de Souza}, {de Torres},
  {Debosscher}, {del Pozo}, {Delbo}, {Delgado}, {Delgado}, {Di Matteo},
  {Diakite}, {Diener}, {Distefano}, {Dolding}, {Drazinos}, {Dur{\'a}n},
  {Edvardsson}, {Enke}, {Eriksson}, {Esquej}, {Eynard Bontemps}, {Fabre},
  {Fabrizio}, {Faigler}, {Falc{\~a}o}, {Farr{\`a}s Casas}, {Federici},
  {Fedorets}, {Fernique}, {Figueras}, {Filippi}, {Findeisen}, {Fonti},
  {Fraile}, {Fraser}, {Fr{\'e}zouls}, {Gai}, {Galleti}, {Garabato},
  {Garc{\'\i}a-Sedano}, {Garofalo}, {Garralda}, {Gavel}, {Gavras}, {Gerssen},
  {Geyer}, {Giacobbe}, {Gilmore}, {Girona}, {Giuffrida}, {Glass}, {Gomes},
  {Granvik}, {Gueguen}, {Guerrier}, {Guiraud}, {Guti{\'e}rrez-S{\'a}nchez},
  {Haigron}, {Hatzidimitriou}, {Hauser}, {Haywood}, {Heiter}, {Helmi}, {Heu},
  {Hilger}, {Hobbs}, {Hofmann}, {Holland}, {Huckle}, {Hypki}, {Icardi},
  {Jan{\ss}en}, {Jevardat de Fombelle}, {Jonker}, {Juh{\'a}sz}, {Julbe},
  {Karampelas}, {Kewley}, {Klar}, {Kochoska}, {Kohley}, {Kolenberg},
  {Kontizas}, {Kontizas}, {Koposov}, {Kordopatis}, {Kostrzewa-Rutkowska},
  {Koubsky}, {Lambert}, {Lanza}, {Lasne}, {Lavigne}, {Le Fustec}, {Le
  Poncin-Lafitte}, {Lebreton}, {Leccia}, {Leclerc}, {Lecoeur-Taibi},
  {Lenhardt}, {Leroux}, {Liao}, {Licata}, {Lindstr{\o}m}, {Lister}, {Livanou},
  {Lobel}, {L{\'o}pez}, {Managau}, {Mann}, {Mantelet}, {Marchal}, {Marchant},
  {Marconi}, {Marinoni}, {Marschalk{\'o}}, {Marshall}, {Martino}, {Marton},
  {Mary}, {Massari}, {Matijevi{\v{c}}}, {Mazeh}, {McMillan}, {Messina},
  {Michalik}, {Millar}, {Molina}, {Molinaro}, {Moln{\'a}r}, {Montegriffo},
  {Mor}, {Morbidelli}, {Morel}, {Morris}, {Mulone}, {Muraveva}, {Musella},
  {Nelemans}, {Nicastro}, {Noval}, {O'Mullane}, {Ord{\'e}novic},
  {Ord{\'o}{\~n}ez-Blanco}, {Osborne}, {Pagani}, {Pagano}, {Pailler},
  {Palacin}, {Palaversa}, {Panahi}, {Pawlak}, {Piersimoni}, {Pineau}, {Plachy},
  {Plum}, {Poggio}, {Poujoulet}, {Pr{\v{s}}a}, {Pulone}, {Racero}, {Ragaini},
  {Rambaux}, {Ramos-Lerate}, {Regibo}, {Reyl{\'e}}, {Riclet}, {Ripepi}, {Riva},
  {Rivard}, {Rixon}, {Roegiers}, {Roelens}, {Romero-G{\'o}mez}, {Rowell},
  {Royer}, {Ruiz-Dern}, {Sadowski}, {Sagrist{\`a} Sell{\'e}s}, {Sahlmann},
  {Salgado}, {Salguero}, {Sanna}, {Santana-Ros}, {Sarasso}, {Savietto},
  {Schultheis}, {Sciacca}, {Segol}, {Segovia}, {S{\'e}gransan}, {Shih},
  {Siltala}, {Silva}, {Smart}, {Smith}, {Solano}, {Solitro}, {Sordo}, {Soria
  Nieto}, {Souchay}, {Spagna}, {Spoto}, {Stampa}, {Steele},
  {Steidelm{\"u}ller}, {Stephenson}, {Stoev}, {Suess}, {Surdej}, {Szabados},
  {Szegedi-Elek}, {Tapiador}, {Taris}, {Tauran}, {Taylor}, {Teixeira},
  {Terrett}, {Teyssandier}, {Thuillot}, {Titarenko}, {Torra Clotet}, {Turon},
  {Ulla}, {Utrilla}, {Uzzi}, {Vaillant}, {Valentini}, {Valette}, {van Elteren},
  {Van Hemelryck}, {van Leeuwen}, {Vaschetto}, {Vecchiato}, {Veljanoski},
  {Viala}, {Vicente}, {Vogt}, {von Essen}, {Voss}, {Votruba}, {Voutsinas},
  {Walmsley}, {Weiler}, {Wertz}, {Wevers}, {Wyrzykowski}, {Yoldas},
  {{\v{Z}}erjal}, {Ziaeepour}, {Zorec}, {Zschocke}, {Zucker}, {Zurbach} and
  {Zwitter}}]{Gaia:2018}
\bibinfo{author}{{Gaia Collaboration}}, \bibinfo{author}{{Brown}, A.G.A.},
  \bibinfo{author}{{Vallenari}, A.}, \bibinfo{author}{{Prusti}, T.},
  \bibinfo{author}{{de Bruijne}, J.H.J.}, \bibinfo{author}{{Babusiaux}, C.},
  \bibinfo{author}{{Bailer-Jones}, C.A.L.}, \bibinfo{author}{{Biermann}, M.},
  \bibinfo{author}{{Evans}, D.W.}, \bibinfo{author}{{Eyer}, L.},
  \bibinfo{author}{{Jansen}, F.}, \bibinfo{author}{{Jordi}, C.},
  \bibinfo{author}{{Klioner}, S.A.}, \bibinfo{author}{{Lammers}, U.},
  \bibinfo{author}{{Lindegren}, L.}, \bibinfo{author}{{Luri}, X.},
  \bibinfo{author}{{Mignard}, F.}, \bibinfo{author}{{Panem}, C.},
  \bibinfo{author}{{Pourbaix}, D.}, \bibinfo{author}{{Randich}, S.},
  \bibinfo{author}{{Sartoretti}, P.}, \bibinfo{author}{{Siddiqui}, H.I.},
  \bibinfo{author}{{Soubiran}, C.}, \bibinfo{author}{{van Leeuwen}, F.},
  \bibinfo{author}{{Walton}, N.A.}, \bibinfo{author}{{Arenou}, F.},
  \bibinfo{author}{{Bastian}, U.}, \bibinfo{author}{{Cropper}, M.},
  \bibinfo{author}{{Drimmel}, R.}, \bibinfo{author}{{Katz}, D.},
  \bibinfo{author}{{Lattanzi}, M.G.}, \bibinfo{author}{{Bakker}, J.},
  \bibinfo{author}{{Cacciari}, C.}, \bibinfo{author}{{Casta{\~n}eda}, J.},
  \bibinfo{author}{{Chaoul}, L.}, \bibinfo{author}{{Cheek}, N.},
  \bibinfo{author}{{De Angeli}, F.}, \bibinfo{author}{{Fabricius}, C.},
  \bibinfo{author}{{Guerra}, R.}, \bibinfo{author}{{Holl}, B.},
  \bibinfo{author}{{Masana}, E.}, \bibinfo{author}{{Messineo}, R.},
  \bibinfo{author}{{Mowlavi}, N.}, \bibinfo{author}{{Nienartowicz}, K.},
  \bibinfo{author}{{Panuzzo}, P.}, \bibinfo{author}{{Portell}, J.},
  \bibinfo{author}{{Riello}, M.}, \bibinfo{author}{{Seabroke}, G.M.},
  \bibinfo{author}{{Tanga}, P.}, \bibinfo{author}{{Th{\'e}venin}, F.},
  \bibinfo{author}{{Gracia-Abril}, G.}, \bibinfo{author}{{Comoretto}, G.},
  \bibinfo{author}{{Garcia-Reinaldos}, M.}, \bibinfo{author}{{Teyssier}, D.},
  \bibinfo{author}{{Altmann}, M.}, \bibinfo{author}{{Andrae}, R.},
  \bibinfo{author}{{Audard}, M.}, \bibinfo{author}{{Bellas-Velidis}, I.},
  \bibinfo{author}{{Benson}, K.}, \bibinfo{author}{{Berthier}, J.},
  \bibinfo{author}{{Blomme}, R.}, \bibinfo{author}{{Burgess}, P.},
  \bibinfo{author}{{Busso}, G.}, \bibinfo{author}{{Carry}, B.},
  \bibinfo{author}{{Cellino}, A.}, \bibinfo{author}{{Clementini}, G.},
  \bibinfo{author}{{Clotet}, M.}, \bibinfo{author}{{Creevey}, O.},
  \bibinfo{author}{{Davidson}, M.}, \bibinfo{author}{{De Ridder}, J.},
  \bibinfo{author}{{Delchambre}, L.}, \bibinfo{author}{{Dell'Oro}, A.},
  \bibinfo{author}{{Ducourant}, C.},
  \bibinfo{author}{{Fern{\'a}ndez-Hern{\'a}ndez}, J.},
  \bibinfo{author}{{Fouesneau}, M.}, \bibinfo{author}{{Fr{\'e}mat}, Y.},
  \bibinfo{author}{{Galluccio}, L.}, \bibinfo{author}{{Garc{\'\i}a-Torres},
  M.}, \bibinfo{author}{{Gonz{\'a}lez-N{\'u}{\~n}ez}, J.},
  \bibinfo{author}{{Gonz{\'a}lez-Vidal}, J.J.}, \bibinfo{author}{{Gosset}, E.},
  \bibinfo{author}{{Guy}, L.P.}, \bibinfo{author}{{Halbwachs}, J.L.},
  \bibinfo{author}{{Hambly}, N.C.}, \bibinfo{author}{{Harrison}, D.L.},
  \bibinfo{author}{{Hern{\'a}ndez}, J.}, \bibinfo{author}{{Hestroffer}, D.},
  \bibinfo{author}{{Hodgkin}, S.T.}, \bibinfo{author}{{Hutton}, A.},
  \bibinfo{author}{{Jasniewicz}, G.}, \bibinfo{author}{{Jean-Antoine-Piccolo},
  A.}, \bibinfo{author}{{Jordan}, S.}, \bibinfo{author}{{Korn}, A.J.},
  \bibinfo{author}{{Krone-Martins}, A.}, \bibinfo{author}{{Lanzafame}, A.C.},
  \bibinfo{author}{{Lebzelter}, T.}, \bibinfo{author}{{L{\"o}ffler}, W.},
  \bibinfo{author}{{Manteiga}, M.}, \bibinfo{author}{{Marrese}, P.M.},
  \bibinfo{author}{{Mart{\'\i}n-Fleitas}, J.M.}, \bibinfo{author}{{Moitinho},
  A.}, \bibinfo{author}{{Mora}, A.}, \bibinfo{author}{{Muinonen}, K.},
  \bibinfo{author}{{Osinde}, J.}, \bibinfo{author}{{Pancino}, E.},
  \bibinfo{author}{{Pauwels}, T.}, \bibinfo{author}{{Petit}, J.M.},
  \bibinfo{author}{{Recio-Blanco}, A.}, \bibinfo{author}{{Richards}, P.J.},
  \bibinfo{author}{{Rimoldini}, L.}, \bibinfo{author}{{Robin}, A.C.},
  \bibinfo{author}{{Sarro}, L.M.}, \bibinfo{author}{{Siopis}, C.},
  \bibinfo{author}{{Smith}, M.}, \bibinfo{author}{{Sozzetti}, A.},
  \bibinfo{author}{{S{\"u}veges}, M.}, \bibinfo{author}{{Torra}, J.},
  \bibinfo{author}{{van Reeven}, W.}, \bibinfo{author}{{Abbas}, U.},
  \bibinfo{author}{{Abreu Aramburu}, A.}, \bibinfo{author}{{Accart}, S.},
  \bibinfo{author}{{Aerts}, C.}, \bibinfo{author}{{Altavilla}, G.},
  \bibinfo{author}{{{\'A}lvarez}, M.A.}, \bibinfo{author}{{Alvarez}, R.},
  \bibinfo{author}{{Alves}, J.}, \bibinfo{author}{{Anderson}, R.I.},
  \bibinfo{author}{{Andrei}, A.H.}, \bibinfo{author}{{Anglada Varela}, E.},
  \bibinfo{author}{{Antiche}, E.}, \bibinfo{author}{{Antoja}, T.},
  \bibinfo{author}{{Arcay}, B.}, \bibinfo{author}{{Astraatmadja}, T.L.},
  \bibinfo{author}{{Bach}, N.}, \bibinfo{author}{{Baker}, S.G.},
  \bibinfo{author}{{Balaguer-N{\'u}{\~n}ez}, L.}, \bibinfo{author}{{Balm}, P.},
  \bibinfo{author}{{Barache}, C.}, \bibinfo{author}{{Barata}, C.},
  \bibinfo{author}{{Barbato}, D.}, \bibinfo{author}{{Barblan}, F.},
  \bibinfo{author}{{Barklem}, P.S.}, \bibinfo{author}{{Barrado}, D.},
  \bibinfo{author}{{Barros}, M.}, \bibinfo{author}{{Barstow}, M.A.},
  \bibinfo{author}{{Bartholom{\'e} Mu{\~n}oz}, S.},
  \bibinfo{author}{{Bassilana}, J.L.}, \bibinfo{author}{{Becciani}, U.},
  \bibinfo{author}{{Bellazzini}, M.}, \bibinfo{author}{{Berihuete}, A.},
  \bibinfo{author}{{Bertone}, S.}, \bibinfo{author}{{Bianchi}, L.},
  \bibinfo{author}{{Bienaym{\'e}}, O.}, \bibinfo{author}{{Blanco-Cuaresma},
  S.}, \bibinfo{author}{{Boch}, T.}, \bibinfo{author}{{Boeche}, C.},
  \bibinfo{author}{{Bombrun}, A.}, \bibinfo{author}{{Borrachero}, R.},
  \bibinfo{author}{{Bossini}, D.}, \bibinfo{author}{{Bouquillon}, S.},
  \bibinfo{author}{{Bourda}, G.}, \bibinfo{author}{{Bragaglia}, A.},
  \bibinfo{author}{{Bramante}, L.}, \bibinfo{author}{{Breddels}, M.A.},
  \bibinfo{author}{{Bressan}, A.}, \bibinfo{author}{{Brouillet}, N.},
  \bibinfo{author}{{Br{\"u}semeister}, T.}, \bibinfo{author}{{Brugaletta}, E.},
  \bibinfo{author}{{Bucciarelli}, B.}, \bibinfo{author}{{Burlacu}, A.},
  \bibinfo{author}{{Busonero}, D.}, \bibinfo{author}{{Butkevich}, A.G.},
  \bibinfo{author}{{Buzzi}, R.}, \bibinfo{author}{{Caffau}, E.},
  \bibinfo{author}{{Cancelliere}, R.}, \bibinfo{author}{{Cannizzaro}, G.},
  \bibinfo{author}{{Cantat-Gaudin}, T.}, \bibinfo{author}{{Carballo}, R.},
  \bibinfo{author}{{Carlucci}, T.}, \bibinfo{author}{{Carrasco}, J.M.},
  \bibinfo{author}{{Casamiquela}, L.}, \bibinfo{author}{{Castellani}, M.},
  \bibinfo{author}{{Castro-Ginard}, A.}, \bibinfo{author}{{Charlot}, P.},
  \bibinfo{author}{{Chemin}, L.}, \bibinfo{author}{{Chiavassa}, A.},
  \bibinfo{author}{{Cocozza}, G.}, \bibinfo{author}{{Costigan}, G.},
  \bibinfo{author}{{Cowell}, S.}, \bibinfo{author}{{Crifo}, F.},
  \bibinfo{author}{{Crosta}, M.}, \bibinfo{author}{{Crowley}, C.},
  \bibinfo{author}{{Cuypers}, J.}, \bibinfo{author}{{Dafonte}, C.},
  \bibinfo{author}{{Damerdji}, Y.}, \bibinfo{author}{{Dapergolas}, A.},
  \bibinfo{author}{{David}, P.}, \bibinfo{author}{{David}, M.},
  \bibinfo{author}{{de Laverny}, P.}, \bibinfo{author}{{De Luise}, F.},
  \bibinfo{author}{{De March}, R.}, \bibinfo{author}{{de Martino}, D.},
  \bibinfo{author}{{de Souza}, R.}, \bibinfo{author}{{de Torres}, A.},
  \bibinfo{author}{{Debosscher}, J.}, \bibinfo{author}{{del Pozo}, E.},
  \bibinfo{author}{{Delbo}, M.}, \bibinfo{author}{{Delgado}, A.},
  \bibinfo{author}{{Delgado}, H.E.}, \bibinfo{author}{{Di Matteo}, P.},
  \bibinfo{author}{{Diakite}, S.}, \bibinfo{author}{{Diener}, C.},
  \bibinfo{author}{{Distefano}, E.}, \bibinfo{author}{{Dolding}, C.},
  \bibinfo{author}{{Drazinos}, P.}, \bibinfo{author}{{Dur{\'a}n}, J.},
  \bibinfo{author}{{Edvardsson}, B.}, \bibinfo{author}{{Enke}, H.},
  \bibinfo{author}{{Eriksson}, K.}, \bibinfo{author}{{Esquej}, P.},
  \bibinfo{author}{{Eynard Bontemps}, G.}, \bibinfo{author}{{Fabre}, C.},
  \bibinfo{author}{{Fabrizio}, M.}, \bibinfo{author}{{Faigler}, S.},
  \bibinfo{author}{{Falc{\~a}o}, A.J.}, \bibinfo{author}{{Farr{\`a}s Casas},
  M.}, \bibinfo{author}{{Federici}, L.}, \bibinfo{author}{{Fedorets}, G.},
  \bibinfo{author}{{Fernique}, P.}, \bibinfo{author}{{Figueras}, F.},
  \bibinfo{author}{{Filippi}, F.}, \bibinfo{author}{{Findeisen}, K.},
  \bibinfo{author}{{Fonti}, A.}, \bibinfo{author}{{Fraile}, E.},
  \bibinfo{author}{{Fraser}, M.}, \bibinfo{author}{{Fr{\'e}zouls}, B.},
  \bibinfo{author}{{Gai}, M.}, \bibinfo{author}{{Galleti}, S.},
  \bibinfo{author}{{Garabato}, D.}, \bibinfo{author}{{Garc{\'\i}a-Sedano}, F.},
  \bibinfo{author}{{Garofalo}, A.}, \bibinfo{author}{{Garralda}, N.},
  \bibinfo{author}{{Gavel}, A.}, \bibinfo{author}{{Gavras}, P.},
  \bibinfo{author}{{Gerssen}, J.}, \bibinfo{author}{{Geyer}, R.},
  \bibinfo{author}{{Giacobbe}, P.}, \bibinfo{author}{{Gilmore}, G.},
  \bibinfo{author}{{Girona}, S.}, \bibinfo{author}{{Giuffrida}, G.},
  \bibinfo{author}{{Glass}, F.}, \bibinfo{author}{{Gomes}, M.},
  \bibinfo{author}{{Granvik}, M.}, \bibinfo{author}{{Gueguen}, A.},
  \bibinfo{author}{{Guerrier}, A.}, \bibinfo{author}{{Guiraud}, J.},
  \bibinfo{author}{{Guti{\'e}rrez-S{\'a}nchez}, R.},
  \bibinfo{author}{{Haigron}, R.}, \bibinfo{author}{{Hatzidimitriou}, D.},
  \bibinfo{author}{{Hauser}, M.}, \bibinfo{author}{{Haywood}, M.},
  \bibinfo{author}{{Heiter}, U.}, \bibinfo{author}{{Helmi}, A.},
  \bibinfo{author}{{Heu}, J.}, \bibinfo{author}{{Hilger}, T.},
  \bibinfo{author}{{Hobbs}, D.}, \bibinfo{author}{{Hofmann}, W.},
  \bibinfo{author}{{Holland}, G.}, \bibinfo{author}{{Huckle}, H.E.},
  \bibinfo{author}{{Hypki}, A.}, \bibinfo{author}{{Icardi}, V.},
  \bibinfo{author}{{Jan{\ss}en}, K.}, \bibinfo{author}{{Jevardat de Fombelle},
  G.}, \bibinfo{author}{{Jonker}, P.G.}, \bibinfo{author}{{Juh{\'a}sz},
  {\'A}.L.}, \bibinfo{author}{{Julbe}, F.}, \bibinfo{author}{{Karampelas}, A.},
  \bibinfo{author}{{Kewley}, A.}, \bibinfo{author}{{Klar}, J.},
  \bibinfo{author}{{Kochoska}, A.}, \bibinfo{author}{{Kohley}, R.},
  \bibinfo{author}{{Kolenberg}, K.}, \bibinfo{author}{{Kontizas}, M.},
  \bibinfo{author}{{Kontizas}, E.}, \bibinfo{author}{{Koposov}, S.E.},
  \bibinfo{author}{{Kordopatis}, G.}, \bibinfo{author}{{Kostrzewa-Rutkowska},
  Z.}, \bibinfo{author}{{Koubsky}, P.}, \bibinfo{author}{{Lambert}, S.},
  \bibinfo{author}{{Lanza}, A.F.}, \bibinfo{author}{{Lasne}, Y.},
  \bibinfo{author}{{Lavigne}, J.B.}, \bibinfo{author}{{Le Fustec}, Y.},
  \bibinfo{author}{{Le Poncin-Lafitte}, C.}, \bibinfo{author}{{Lebreton}, Y.},
  \bibinfo{author}{{Leccia}, S.}, \bibinfo{author}{{Leclerc}, N.},
  \bibinfo{author}{{Lecoeur-Taibi}, I.}, \bibinfo{author}{{Lenhardt}, H.},
  \bibinfo{author}{{Leroux}, F.}, \bibinfo{author}{{Liao}, S.},
  \bibinfo{author}{{Licata}, E.}, \bibinfo{author}{{Lindstr{\o}m}, H.E.P.},
  \bibinfo{author}{{Lister}, T.A.}, \bibinfo{author}{{Livanou}, E.},
  \bibinfo{author}{{Lobel}, A.}, \bibinfo{author}{{L{\'o}pez}, M.},
  \bibinfo{author}{{Managau}, S.}, \bibinfo{author}{{Mann}, R.G.},
  \bibinfo{author}{{Mantelet}, G.}, \bibinfo{author}{{Marchal}, O.},
  \bibinfo{author}{{Marchant}, J.M.}, \bibinfo{author}{{Marconi}, M.},
  \bibinfo{author}{{Marinoni}, S.}, \bibinfo{author}{{Marschalk{\'o}}, G.},
  \bibinfo{author}{{Marshall}, D.J.}, \bibinfo{author}{{Martino}, M.},
  \bibinfo{author}{{Marton}, G.}, \bibinfo{author}{{Mary}, N.},
  \bibinfo{author}{{Massari}, D.}, \bibinfo{author}{{Matijevi{\v{c}}}, G.},
  \bibinfo{author}{{Mazeh}, T.}, \bibinfo{author}{{McMillan}, P.J.},
  \bibinfo{author}{{Messina}, S.}, \bibinfo{author}{{Michalik}, D.},
  \bibinfo{author}{{Millar}, N.R.}, \bibinfo{author}{{Molina}, D.},
  \bibinfo{author}{{Molinaro}, R.}, \bibinfo{author}{{Moln{\'a}r}, L.},
  \bibinfo{author}{{Montegriffo}, P.}, \bibinfo{author}{{Mor}, R.},
  \bibinfo{author}{{Morbidelli}, R.}, \bibinfo{author}{{Morel}, T.},
  \bibinfo{author}{{Morris}, D.}, \bibinfo{author}{{Mulone}, A.F.},
  \bibinfo{author}{{Muraveva}, T.}, \bibinfo{author}{{Musella}, I.},
  \bibinfo{author}{{Nelemans}, G.}, \bibinfo{author}{{Nicastro}, L.},
  \bibinfo{author}{{Noval}, L.}, \bibinfo{author}{{O'Mullane}, W.},
  \bibinfo{author}{{Ord{\'e}novic}, C.},
  \bibinfo{author}{{Ord{\'o}{\~n}ez-Blanco}, D.}, \bibinfo{author}{{Osborne},
  P.}, \bibinfo{author}{{Pagani}, C.}, \bibinfo{author}{{Pagano}, I.},
  \bibinfo{author}{{Pailler}, F.}, \bibinfo{author}{{Palacin}, H.},
  \bibinfo{author}{{Palaversa}, L.}, \bibinfo{author}{{Panahi}, A.},
  \bibinfo{author}{{Pawlak}, M.}, \bibinfo{author}{{Piersimoni}, A.M.},
  \bibinfo{author}{{Pineau}, F.X.}, \bibinfo{author}{{Plachy}, E.},
  \bibinfo{author}{{Plum}, G.}, \bibinfo{author}{{Poggio}, E.},
  \bibinfo{author}{{Poujoulet}, E.}, \bibinfo{author}{{Pr{\v{s}}a}, A.},
  \bibinfo{author}{{Pulone}, L.}, \bibinfo{author}{{Racero}, E.},
  \bibinfo{author}{{Ragaini}, S.}, \bibinfo{author}{{Rambaux}, N.},
  \bibinfo{author}{{Ramos-Lerate}, M.}, \bibinfo{author}{{Regibo}, S.},
  \bibinfo{author}{{Reyl{\'e}}, C.}, \bibinfo{author}{{Riclet}, F.},
  \bibinfo{author}{{Ripepi}, V.}, \bibinfo{author}{{Riva}, A.},
  \bibinfo{author}{{Rivard}, A.}, \bibinfo{author}{{Rixon}, G.},
  \bibinfo{author}{{Roegiers}, T.}, \bibinfo{author}{{Roelens}, M.},
  \bibinfo{author}{{Romero-G{\'o}mez}, M.}, \bibinfo{author}{{Rowell}, N.},
  \bibinfo{author}{{Royer}, F.}, \bibinfo{author}{{Ruiz-Dern}, L.},
  \bibinfo{author}{{Sadowski}, G.}, \bibinfo{author}{{Sagrist{\`a} Sell{\'e}s},
  T.}, \bibinfo{author}{{Sahlmann}, J.}, \bibinfo{author}{{Salgado}, J.},
  \bibinfo{author}{{Salguero}, E.}, \bibinfo{author}{{Sanna}, N.},
  \bibinfo{author}{{Santana-Ros}, T.}, \bibinfo{author}{{Sarasso}, M.},
  \bibinfo{author}{{Savietto}, H.}, \bibinfo{author}{{Schultheis}, M.},
  \bibinfo{author}{{Sciacca}, E.}, \bibinfo{author}{{Segol}, M.},
  \bibinfo{author}{{Segovia}, J.C.}, \bibinfo{author}{{S{\'e}gransan}, D.},
  \bibinfo{author}{{Shih}, I.C.}, \bibinfo{author}{{Siltala}, L.},
  \bibinfo{author}{{Silva}, A.F.}, \bibinfo{author}{{Smart}, R.L.},
  \bibinfo{author}{{Smith}, K.W.}, \bibinfo{author}{{Solano}, E.},
  \bibinfo{author}{{Solitro}, F.}, \bibinfo{author}{{Sordo}, R.},
  \bibinfo{author}{{Soria Nieto}, S.}, \bibinfo{author}{{Souchay}, J.},
  \bibinfo{author}{{Spagna}, A.}, \bibinfo{author}{{Spoto}, F.},
  \bibinfo{author}{{Stampa}, U.}, \bibinfo{author}{{Steele}, I.A.},
  \bibinfo{author}{{Steidelm{\"u}ller}, H.}, \bibinfo{author}{{Stephenson},
  C.A.}, \bibinfo{author}{{Stoev}, H.}, \bibinfo{author}{{Suess}, F.F.},
  \bibinfo{author}{{Surdej}, J.}, \bibinfo{author}{{Szabados}, L.},
  \bibinfo{author}{{Szegedi-Elek}, E.}, \bibinfo{author}{{Tapiador}, D.},
  \bibinfo{author}{{Taris}, F.}, \bibinfo{author}{{Tauran}, G.},
  \bibinfo{author}{{Taylor}, M.B.}, \bibinfo{author}{{Teixeira}, R.},
  \bibinfo{author}{{Terrett}, D.}, \bibinfo{author}{{Teyssandier}, P.},
  \bibinfo{author}{{Thuillot}, W.}, \bibinfo{author}{{Titarenko}, A.},
  \bibinfo{author}{{Torra Clotet}, F.}, \bibinfo{author}{{Turon}, C.},
  \bibinfo{author}{{Ulla}, A.}, \bibinfo{author}{{Utrilla}, E.},
  \bibinfo{author}{{Uzzi}, S.}, \bibinfo{author}{{Vaillant}, M.},
  \bibinfo{author}{{Valentini}, G.}, \bibinfo{author}{{Valette}, V.},
  \bibinfo{author}{{van Elteren}, A.}, \bibinfo{author}{{Van Hemelryck}, E.},
  \bibinfo{author}{{van Leeuwen}, M.}, \bibinfo{author}{{Vaschetto}, M.},
  \bibinfo{author}{{Vecchiato}, A.}, \bibinfo{author}{{Veljanoski}, J.},
  \bibinfo{author}{{Viala}, Y.}, \bibinfo{author}{{Vicente}, D.},
  \bibinfo{author}{{Vogt}, S.}, \bibinfo{author}{{von Essen}, C.},
  \bibinfo{author}{{Voss}, H.}, \bibinfo{author}{{Votruba}, V.},
  \bibinfo{author}{{Voutsinas}, S.}, \bibinfo{author}{{Walmsley}, G.},
  \bibinfo{author}{{Weiler}, M.}, \bibinfo{author}{{Wertz}, O.},
  \bibinfo{author}{{Wevers}, T.}, \bibinfo{author}{{Wyrzykowski}, {\L}.},
  \bibinfo{author}{{Yoldas}, A.}, \bibinfo{author}{{{\v{Z}}erjal}, M.},
  \bibinfo{author}{{Ziaeepour}, H.}, \bibinfo{author}{{Zorec}, J.},
  \bibinfo{author}{{Zschocke}, S.}, \bibinfo{author}{{Zucker}, S.},
  \bibinfo{author}{{Zurbach}, C.}, \bibinfo{author}{{Zwitter}, T.},
  \bibinfo{year}{2018}.
\newblock \bibinfo{title}{{Gaia Data Release 2. Summary of the contents and
  survey properties}}.
\newblock \bibinfo{journal}{\aap} \bibinfo{volume}{616}, \bibinfo{pages}{A1}.
\newblock \DOIprefix\doi{10.1051/0004-6361/201833051},
  \href{http://arxiv.org/abs/1804.09365}{\tt arXiv:1804.09365}.
\bibitem[{{Gaia Collaboration} et~al.(2022b){Gaia Collaboration}, Davidson,
  Hambly, Mann, Morris, Rowell and Voutsinas}]{Gaia:2022}
\bibinfo{author}{{Gaia Collaboration}}, \bibinfo{author}{Davidson, M.},
  \bibinfo{author}{Hambly, N.}, \bibinfo{author}{Mann, R.},
  \bibinfo{author}{Morris, D.}, \bibinfo{author}{Rowell, N.},
  \bibinfo{author}{Voutsinas, S.}, \bibinfo{year}{2022}b.
\newblock \bibinfo{title}{Gaia data release 3. summary of the content and
  survey properties}.
\newblock \bibinfo{journal}{\aap} \DOIprefix\doi{10.1051/0004-6361/202243940}.
  \bibinfo{note}{papers submitted to A\&A from 04.04.22 to be made OA.}
\bibitem[{{Gaia Collaboration} et~al.(2016){Gaia Collaboration}, {Prusti}, {de
  Bruijne}, {Brown}, {Vallenari}, {Babusiaux}, {Bailer-Jones}, {Bastian},
  {Biermann}, {Evans}, {Eyer}, {Jansen}, {Jordi}, {Klioner}, {Lammers},
  {Lindegren}, {Luri}, {Mignard}, {Milligan}, {Panem}, {Poinsignon},
  {Pourbaix}, {Randich}, {Sarri}, {Sartoretti}, {Siddiqui}, {Soubiran},
  {Valette}, {van Leeuwen}, {Walton}, {Aerts}, {Arenou}, {Cropper}, {Drimmel},
  {H{\o}g}, {Katz}, {Lattanzi}, {O'Mullane}, {Grebel}, {Holland}, {Huc},
  {Passot}, {Bramante}, {Cacciari}, {Casta{\~n}eda}, {Chaoul}, {Cheek}, {De
  Angeli}, {Fabricius}, {Guerra}, {Hern{\'a}ndez}, {Jean-Antoine-Piccolo},
  {Masana}, {Messineo}, {Mowlavi}, {Nienartowicz}, {Ord{\'o}{\~n}ez-Blanco},
  {Panuzzo}, {Portell}, {Richards}, {Riello}, {Seabroke}, {Tanga},
  {Th{\'e}venin}, {Torra}, {Els}, {Gracia-Abril}, {Comoretto},
  {Garcia-Reinaldos}, {Lock}, {Mercier}, {Altmann}, {Andrae}, {Astraatmadja},
  {Bellas-Velidis}, {Benson}, {Berthier}, {Blomme}, {Busso}, {Carry},
  {Cellino}, {Clementini}, {Cowell}, {Creevey}, {Cuypers}, {Davidson}, {De
  Ridder}, {de Torres}, {Delchambre}, {Dell'Oro}, {Ducourant}, {Fr{\'e}mat},
  {Garc{\'\i}a-Torres}, {Gosset}, {Halbwachs}, {Hambly}, {Harrison}, {Hauser},
  {Hestroffer}, {Hodgkin}, {Huckle}, {Hutton}, {Jasniewicz}, {Jordan},
  {Kontizas}, {Korn}, {Lanzafame}, {Manteiga}, {Moitinho}, {Muinonen},
  {Osinde}, {Pancino}, {Pauwels}, {Petit}, {Recio-Blanco}, {Robin}, {Sarro},
  {Siopis}, {Smith}, {Smith}, {Sozzetti}, {Thuillot}, {van Reeven}, {Viala},
  {Abbas}, {Abreu Aramburu}, {Accart}, {Aguado}, {Allan}, {Allasia},
  {Altavilla}, {{\'A}lvarez}, {Alves}, {Anderson}, {Andrei}, {Anglada Varela},
  {Antiche}, {Antoja}, {Ant{\'o}n}, {Arcay}, {Atzei}, {Ayache}, {Bach},
  {Baker}, {Balaguer-N{\'u}{\~n}ez}, {Barache}, {Barata}, {Barbier}, {Barblan},
  {Baroni}, {Barrado y Navascu{\'e}s}, {Barros}, {Barstow}, {Becciani},
  {Bellazzini}, {Bellei}, {Bello Garc{\'\i}a}, {Belokurov}, {Bendjoya},
  {Berihuete}, {Bianchi}, {Bienaym{\'e}}, {Billebaud}, {Blagorodnova},
  {Blanco-Cuaresma}, {Boch}, {Bombrun}, {Borrachero}, {Bouquillon}, {Bourda},
  {Bouy}, {Bragaglia}, {Breddels}, {Brouillet}, {Br{\"u}semeister},
  {Bucciarelli}, {Budnik}, {Burgess}, {Burgon}, {Burlacu}, {Busonero}, {Buzzi},
  {Caffau}, {Cambras}, {Campbell}, {Cancelliere}, {Cantat-Gaudin}, {Carlucci},
  {Carrasco}, {Castellani}, {Charlot}, {Charnas}, {Charvet}, {Chassat},
  {Chiavassa}, {Clotet}, {Cocozza}, {Collins}, {Collins}, {Costigan}, {Crifo},
  {Cross}, {Crosta}, {Crowley}, {Dafonte}, {Damerdji}, {Dapergolas}, {David},
  {David}, {De Cat}, {de Felice}, {de Laverny}, {De Luise}, {De March}, {de
  Martino}, {de Souza}, {Debosscher}, {del Pozo}, {Delbo}, {Delgado},
  {Delgado}, {di Marco}, {Di Matteo}, {Diakite}, {Distefano}, {Dolding}, {Dos
  Anjos}, {Drazinos}, {Dur{\'a}n}, {Dzigan}, {Ecale}, {Edvardsson}, {Enke},
  {Erdmann}, {Escolar}, {Espina}, {Evans}, {Eynard Bontemps}, {Fabre},
  {Fabrizio}, {Faigler}, {Falc{\~a}o}, {Farr{\`a}s Casas}, {Faye}, {Federici},
  {Fedorets}, {Fern{\'a}ndez-Hern{\'a}ndez}, {Fernique}, {Fienga}, {Figueras},
  {Filippi}, {Findeisen}, {Fonti}, {Fouesneau}, {Fraile}, {Fraser}, {Fuchs},
  {Furnell}, {Gai}, {Galleti}, {Galluccio}, {Garabato}, {Garc{\'\i}a-Sedano},
  {Gar{\'e}}, {Garofalo}, {Garralda}, {Gavras}, {Gerssen}, {Geyer}, {Gilmore},
  {Girona}, {Giuffrida}, {Gomes}, {Gonz{\'a}lez-Marcos},
  {Gonz{\'a}lez-N{\'u}{\~n}ez}, {Gonz{\'a}lez-Vidal}, {Granvik}, {Guerrier},
  {Guillout}, {Guiraud}, {G{\'u}rpide}, {Guti{\'e}rrez-S{\'a}nchez}, {Guy},
  {Haigron}, {Hatzidimitriou}, {Haywood}, {Heiter}, {Helmi}, {Hobbs},
  {Hofmann}, {Holl}, {Holland }, {Hunt}, {Hypki}, {Icardi}, {Irwin}, {Jevardat
  de Fombelle}, {Jofr{\'e}}, {Jonker}, {Jorissen}, {Julbe}, {Karampelas},
  {Kochoska}, {Kohley}, {Kolenberg}, {Kontizas}, {Koposov}, {Kordopatis},
  {Koubsky}, {Kowalczyk}, {Krone-Martins}, {Kudryashova}, {Kull}, {Bachchan},
  {Lacoste-Seris}, {Lanza}, {Lavigne}, {Le Poncin-Lafitte}, {Lebreton},
  {Lebzelter}, {Leccia}, {Leclerc}, {Lecoeur-Taibi}, {Lemaitre}, {Lenhardt},
  {Leroux}, {Liao}, {Licata}, {Lindstr{\o}m}, {Lister}, {Livanou}, {Lobel},
  {L{\"o}ffler}, {L{\'o}pez}, {Lopez-Lozano}, {Lorenz}, {Loureiro},
  {MacDonald}, {Magalh{\~a}es Fernandes}, {Managau}, {Mann}, {Mantelet},
  {Marchal}, {Marchant}, {Marconi}, {Marie}, {Marinoni}, {Marrese},
  {Marschalk{\'o}}, {Marshall}, {Mart{\'\i}n-Fleitas}, {Martino}, {Mary},
  {Matijevi{\v{c}}}, {Mazeh}, {McMillan}, {Messina}, {Mestre}, {Michalik},
  {Millar}, {Miranda}, {Molina}, {Molinaro}, {Molinaro}, {Moln{\'a}r},
  {Moniez}, {Montegriffo}, {Monteiro}, {Mor}, {Mora}, {Morbidelli}, {Morel},
  {Morgenthaler}, {Morley}, {Morris}, {Mulone}, {Muraveva}, {Musella},
  {Narbonne}, {Nelemans}, {Nicastro}, {Noval}, {Ord{\'e}novic},
  {Ordieres-Mer{\'e}}, {Osborne}, {Pagani}, {Pagano}, {Pailler}, {Palacin},
  {Palaversa}, {Parsons}, {Paulsen}, {Pecoraro}, {Pedrosa}, {Pentik{\"a}inen},
  {Pereira}, {Pichon}, {Piersimoni}, {Pineau}, {Plachy}, {Plum}, {Poujoulet},
  {Pr{\v{s}}a}, {Pulone}, {Ragaini}, {Rago}, {Rambaux}, {Ramos-Lerate},
  {Ranalli}, {Rauw}, {Read}, {Regibo}, {Renk}, {Reyl{\'e}}, {Ribeiro},
  {Rimoldini}, {Ripepi}, {Riva}, {Rixon}, {Roelens}, {Romero-G{\'o}mez},
  {Rowell}, {Royer}, {Rudolph}, {Ruiz-Dern}, {Sadowski}, {Sagrist{\`a}
  Sell{\'e}s}, {Sahlmann}, {Salgado}, {Salguero}, {Sarasso}, {Savietto},
  {Schnorhk}, {Schultheis}, {Sciacca}, {Segol}, {Segovia}, {Segransan},
  {Serpell}, {Shih}, {Smareglia}, {Smart}, {Smith}, {Solano}, {Solitro},
  {Sordo}, {Soria Nieto}, {Souchay}, {Spagna}, {Spoto}, {Stampa}, {Steele},
  {Steidelm{\"u}ller}, {Stephenson}, {Stoev}, {Suess}, {S{\"u}veges}, {Surdej},
  {Szabados}, {Szegedi-Elek}, {Tapiador}, {Taris}, {Tauran}, {Taylor},
  {Teixeira}, {Terrett}, {Tingley}, {Trager}, {Turon}, {Ulla}, {Utrilla},
  {Valentini}, {van Elteren}, {Van Hemelryck}, {van Leeuwen}, {Varadi},
  {Vecchiato}, {Veljanoski}, {Via}, {Vicente}, {Vogt}, {Voss}, {Votruba},
  {Voutsinas}, {Walmsley}, {Weiler}, {Weingrill}, {Werner}, {Wevers},
  {Whitehead}, {Wyrzykowski}, {Yoldas}, {{\v{Z}}erjal}, {Zucker}, {Zurbach},
  {Zwitter}, {Alecu}, {Allen}, {Allende Prieto}, {Amorim},
  {Anglada-Escud{\'e}}, {Arsenijevic}, {Azaz}, {Balm}, {Beck}, {Bernstein},
  {Bigot}, {Bijaoui}, {Blasco}, {Bonfigli}, {Bono}, {Boudreault}, {Bressan},
  {Brown}, {Brunet}, {Bunclark}, {Buonanno}, {Butkevich}, {Carret}, {Carrion},
  {Chemin}, {Ch{\'e}reau}, {Corcione}, {Darmigny}, {de Boer}, {de Teodoro}, {de
  Zeeuw}, {Delle Luche}, {Domingues}, {Dubath}, {Fodor}, {Fr{\'e}zouls},
  {Fries}, {Fustes}, {Fyfe}, {Gallardo}, {Gallegos}, {Gardiol}, {Gebran},
  {Gomboc}, {G{\'o}mez}, {Grux}, {Gueguen}, {Heyrovsky}, {Hoar}, {Iannicola},
  {Isasi Parache}, {Janotto}, {Joliet}, {Jonckheere}, {Keil}, {Kim},
  {Klagyivik}, {Klar}, {Knude}, {Kochukhov}, {Kolka}, {Kos}, {Kutka}, {Lainey},
  {LeBouquin}, {Liu}, {Loreggia}, {Makarov}, {Marseille}, {Martayan},
  {Martinez-Rubi}, {Massart}, {Meynadier}, {Mignot}, {Munari}, {Nguyen},
  {Nordlander}, {Ocvirk}, {O'Flaherty}, {Olias Sanz}, {Ortiz}, {Osorio},
  {Oszkiewicz}, {Ouzounis}, {Palmer}, {Park}, {Pasquato}, {Peltzer}, {Peralta},
  {P{\'e}turaud}, {Pieniluoma}, {Pigozzi}, {Poels}, {Prat}, {Prod'homme},
  {Raison}, {Rebordao}, {Risquez}, {Rocca-Volmerange}, {Rosen}, {Ruiz-Fuertes},
  {Russo}, {Sembay}, {Serraller Vizcaino}, {Short}, {Siebert}, {Silva},
  {Sinachopoulos}, {Slezak}, {Soffel}, {Sosnowska}, {Strai{\v{z}}ys}, {ter
  Linden}, {Terrell}, {Theil}, {Tiede}, {Troisi}, {Tsalmantza}, {Tur},
  {Vaccari}, {Vachier}, {Valles}, {Van Hamme}, {Veltz}, {Virtanen}, {Wallut},
  {Wichmann}, {Wilkinson}, {Ziaeepour} and {Zschocke}}]{Gaia:2016}
\bibinfo{author}{{Gaia Collaboration}}, \bibinfo{author}{{Prusti}, T.},
  \bibinfo{author}{{de Bruijne}, J.H.J.}, \bibinfo{author}{{Brown}, A.G.A.},
  \bibinfo{author}{{Vallenari}, A.}, \bibinfo{author}{{Babusiaux}, C.},
  \bibinfo{author}{{Bailer-Jones}, C.A.L.}, \bibinfo{author}{{Bastian}, U.},
  \bibinfo{author}{{Biermann}, M.}, \bibinfo{author}{{Evans}, D.W.},
  \bibinfo{author}{{Eyer}, L.}, \bibinfo{author}{{Jansen}, F.},
  \bibinfo{author}{{Jordi}, C.}, \bibinfo{author}{{Klioner}, S.A.},
  \bibinfo{author}{{Lammers}, U.}, \bibinfo{author}{{Lindegren}, L.},
  \bibinfo{author}{{Luri}, X.}, \bibinfo{author}{{Mignard}, F.},
  \bibinfo{author}{{Milligan}, D.J.}, \bibinfo{author}{{Panem}, C.},
  \bibinfo{author}{{Poinsignon}, V.}, \bibinfo{author}{{Pourbaix}, D.},
  \bibinfo{author}{{Randich}, S.}, \bibinfo{author}{{Sarri}, G.},
  \bibinfo{author}{{Sartoretti}, P.}, \bibinfo{author}{{Siddiqui}, H.I.},
  \bibinfo{author}{{Soubiran}, C.}, \bibinfo{author}{{Valette}, V.},
  \bibinfo{author}{{van Leeuwen}, F.}, \bibinfo{author}{{Walton}, N.A.},
  \bibinfo{author}{{Aerts}, C.}, \bibinfo{author}{{Arenou}, F.},
  \bibinfo{author}{{Cropper}, M.}, \bibinfo{author}{{Drimmel}, R.},
  \bibinfo{author}{{H{\o}g}, E.}, \bibinfo{author}{{Katz}, D.},
  \bibinfo{author}{{Lattanzi}, M.G.}, \bibinfo{author}{{O'Mullane}, W.},
  \bibinfo{author}{{Grebel}, E.K.}, \bibinfo{author}{{Holland}, A.D.},
  \bibinfo{author}{{Huc}, C.}, \bibinfo{author}{{Passot}, X.},
  \bibinfo{author}{{Bramante}, L.}, \bibinfo{author}{{Cacciari}, C.},
  \bibinfo{author}{{Casta{\~n}eda}, J.}, \bibinfo{author}{{Chaoul}, L.},
  \bibinfo{author}{{Cheek}, N.}, \bibinfo{author}{{De Angeli}, F.},
  \bibinfo{author}{{Fabricius}, C.}, \bibinfo{author}{{Guerra}, R.},
  \bibinfo{author}{{Hern{\'a}ndez}, J.},
  \bibinfo{author}{{Jean-Antoine-Piccolo}, A.}, \bibinfo{author}{{Masana}, E.},
  \bibinfo{author}{{Messineo}, R.}, \bibinfo{author}{{Mowlavi}, N.},
  \bibinfo{author}{{Nienartowicz}, K.},
  \bibinfo{author}{{Ord{\'o}{\~n}ez-Blanco}, D.}, \bibinfo{author}{{Panuzzo},
  P.}, \bibinfo{author}{{Portell}, J.}, \bibinfo{author}{{Richards}, P.J.},
  \bibinfo{author}{{Riello}, M.}, \bibinfo{author}{{Seabroke}, G.M.},
  \bibinfo{author}{{Tanga}, P.}, \bibinfo{author}{{Th{\'e}venin}, F.},
  \bibinfo{author}{{Torra}, J.}, \bibinfo{author}{{Els}, S.G.},
  \bibinfo{author}{{Gracia-Abril}, G.}, \bibinfo{author}{{Comoretto}, G.},
  \bibinfo{author}{{Garcia-Reinaldos}, M.}, \bibinfo{author}{{Lock}, T.},
  \bibinfo{author}{{Mercier}, E.}, \bibinfo{author}{{Altmann}, M.},
  \bibinfo{author}{{Andrae}, R.}, \bibinfo{author}{{Astraatmadja}, T.L.},
  \bibinfo{author}{{Bellas-Velidis}, I.}, \bibinfo{author}{{Benson}, K.},
  \bibinfo{author}{{Berthier}, J.}, \bibinfo{author}{{Blomme}, R.},
  \bibinfo{author}{{Busso}, G.}, \bibinfo{author}{{Carry}, B.},
  \bibinfo{author}{{Cellino}, A.}, \bibinfo{author}{{Clementini}, G.},
  \bibinfo{author}{{Cowell}, S.}, \bibinfo{author}{{Creevey}, O.},
  \bibinfo{author}{{Cuypers}, J.}, \bibinfo{author}{{Davidson}, M.},
  \bibinfo{author}{{De Ridder}, J.}, \bibinfo{author}{{de Torres}, A.},
  \bibinfo{author}{{Delchambre}, L.}, \bibinfo{author}{{Dell'Oro}, A.},
  \bibinfo{author}{{Ducourant}, C.}, \bibinfo{author}{{Fr{\'e}mat}, Y.},
  \bibinfo{author}{{Garc{\'\i}a-Torres}, M.}, \bibinfo{author}{{Gosset}, E.},
  \bibinfo{author}{{Halbwachs}, J.L.}, \bibinfo{author}{{Hambly}, N.C.},
  \bibinfo{author}{{Harrison}, D.L.}, \bibinfo{author}{{Hauser}, M.},
  \bibinfo{author}{{Hestroffer}, D.}, \bibinfo{author}{{Hodgkin}, S.T.},
  \bibinfo{author}{{Huckle}, H.E.}, \bibinfo{author}{{Hutton}, A.},
  \bibinfo{author}{{Jasniewicz}, G.}, \bibinfo{author}{{Jordan}, S.},
  \bibinfo{author}{{Kontizas}, M.}, \bibinfo{author}{{Korn}, A.J.},
  \bibinfo{author}{{Lanzafame}, A.C.}, \bibinfo{author}{{Manteiga}, M.},
  \bibinfo{author}{{Moitinho}, A.}, \bibinfo{author}{{Muinonen}, K.},
  \bibinfo{author}{{Osinde}, J.}, \bibinfo{author}{{Pancino}, E.},
  \bibinfo{author}{{Pauwels}, T.}, \bibinfo{author}{{Petit}, J.M.},
  \bibinfo{author}{{Recio-Blanco}, A.}, \bibinfo{author}{{Robin}, A.C.},
  \bibinfo{author}{{Sarro}, L.M.}, \bibinfo{author}{{Siopis}, C.},
  \bibinfo{author}{{Smith}, M.}, \bibinfo{author}{{Smith}, K.W.},
  \bibinfo{author}{{Sozzetti}, A.}, \bibinfo{author}{{Thuillot}, W.},
  \bibinfo{author}{{van Reeven}, W.}, \bibinfo{author}{{Viala}, Y.},
  \bibinfo{author}{{Abbas}, U.}, \bibinfo{author}{{Abreu Aramburu}, A.},
  \bibinfo{author}{{Accart}, S.}, \bibinfo{author}{{Aguado}, J.J.},
  \bibinfo{author}{{Allan}, P.M.}, \bibinfo{author}{{Allasia}, W.},
  \bibinfo{author}{{Altavilla}, G.}, \bibinfo{author}{{{\'A}lvarez}, M.A.},
  \bibinfo{author}{{Alves}, J.}, \bibinfo{author}{{Anderson}, R.I.},
  \bibinfo{author}{{Andrei}, A.H.}, \bibinfo{author}{{Anglada Varela}, E.},
  \bibinfo{author}{{Antiche}, E.}, \bibinfo{author}{{Antoja}, T.},
  \bibinfo{author}{{Ant{\'o}n}, S.}, \bibinfo{author}{{Arcay}, B.},
  \bibinfo{author}{{Atzei}, A.}, \bibinfo{author}{{Ayache}, L.},
  \bibinfo{author}{{Bach}, N.}, \bibinfo{author}{{Baker}, S.G.},
  \bibinfo{author}{{Balaguer-N{\'u}{\~n}ez}, L.}, \bibinfo{author}{{Barache},
  C.}, \bibinfo{author}{{Barata}, C.}, \bibinfo{author}{{Barbier}, A.},
  \bibinfo{author}{{Barblan}, F.}, \bibinfo{author}{{Baroni}, M.},
  \bibinfo{author}{{Barrado y Navascu{\'e}s}, D.}, \bibinfo{author}{{Barros},
  M.}, \bibinfo{author}{{Barstow}, M.A.}, \bibinfo{author}{{Becciani}, U.},
  \bibinfo{author}{{Bellazzini}, M.}, \bibinfo{author}{{Bellei}, G.},
  \bibinfo{author}{{Bello Garc{\'\i}a}, A.}, \bibinfo{author}{{Belokurov}, V.},
  \bibinfo{author}{{Bendjoya}, P.}, \bibinfo{author}{{Berihuete}, A.},
  \bibinfo{author}{{Bianchi}, L.}, \bibinfo{author}{{Bienaym{\'e}}, O.},
  \bibinfo{author}{{Billebaud}, F.}, \bibinfo{author}{{Blagorodnova}, N.},
  \bibinfo{author}{{Blanco-Cuaresma}, S.}, \bibinfo{author}{{Boch}, T.},
  \bibinfo{author}{{Bombrun}, A.}, \bibinfo{author}{{Borrachero}, R.},
  \bibinfo{author}{{Bouquillon}, S.}, \bibinfo{author}{{Bourda}, G.},
  \bibinfo{author}{{Bouy}, H.}, \bibinfo{author}{{Bragaglia}, A.},
  \bibinfo{author}{{Breddels}, M.A.}, \bibinfo{author}{{Brouillet}, N.},
  \bibinfo{author}{{Br{\"u}semeister}, T.}, \bibinfo{author}{{Bucciarelli},
  B.}, \bibinfo{author}{{Budnik}, F.}, \bibinfo{author}{{Burgess}, P.},
  \bibinfo{author}{{Burgon}, R.}, \bibinfo{author}{{Burlacu}, A.},
  \bibinfo{author}{{Busonero}, D.}, \bibinfo{author}{{Buzzi}, R.},
  \bibinfo{author}{{Caffau}, E.}, \bibinfo{author}{{Cambras}, J.},
  \bibinfo{author}{{Campbell}, H.}, \bibinfo{author}{{Cancelliere}, R.},
  \bibinfo{author}{{Cantat-Gaudin}, T.}, \bibinfo{author}{{Carlucci}, T.},
  \bibinfo{author}{{Carrasco}, J.M.}, \bibinfo{author}{{Castellani}, M.},
  \bibinfo{author}{{Charlot}, P.}, \bibinfo{author}{{Charnas}, J.},
  \bibinfo{author}{{Charvet}, P.}, \bibinfo{author}{{Chassat}, F.},
  \bibinfo{author}{{Chiavassa}, A.}, \bibinfo{author}{{Clotet}, M.},
  \bibinfo{author}{{Cocozza}, G.}, \bibinfo{author}{{Collins}, R.S.},
  \bibinfo{author}{{Collins}, P.}, \bibinfo{author}{{Costigan}, G.},
  \bibinfo{author}{{Crifo}, F.}, \bibinfo{author}{{Cross}, N.J.G.},
  \bibinfo{author}{{Crosta}, M.}, \bibinfo{author}{{Crowley}, C.},
  \bibinfo{author}{{Dafonte}, C.}, \bibinfo{author}{{Damerdji}, Y.},
  \bibinfo{author}{{Dapergolas}, A.}, \bibinfo{author}{{David}, P.},
  \bibinfo{author}{{David}, M.}, \bibinfo{author}{{De Cat}, P.},
  \bibinfo{author}{{de Felice}, F.}, \bibinfo{author}{{de Laverny}, P.},
  \bibinfo{author}{{De Luise}, F.}, \bibinfo{author}{{De March}, R.},
  \bibinfo{author}{{de Martino}, D.}, \bibinfo{author}{{de Souza}, R.},
  \bibinfo{author}{{Debosscher}, J.}, \bibinfo{author}{{del Pozo}, E.},
  \bibinfo{author}{{Delbo}, M.}, \bibinfo{author}{{Delgado}, A.},
  \bibinfo{author}{{Delgado}, H.E.}, \bibinfo{author}{{di Marco}, F.},
  \bibinfo{author}{{Di Matteo}, P.}, \bibinfo{author}{{Diakite}, S.},
  \bibinfo{author}{{Distefano}, E.}, \bibinfo{author}{{Dolding}, C.},
  \bibinfo{author}{{Dos Anjos}, S.}, \bibinfo{author}{{Drazinos}, P.},
  \bibinfo{author}{{Dur{\'a}n}, J.}, \bibinfo{author}{{Dzigan}, Y.},
  \bibinfo{author}{{Ecale}, E.}, \bibinfo{author}{{Edvardsson}, B.},
  \bibinfo{author}{{Enke}, H.}, \bibinfo{author}{{Erdmann}, M.},
  \bibinfo{author}{{Escolar}, D.}, \bibinfo{author}{{Espina}, M.},
  \bibinfo{author}{{Evans}, N.W.}, \bibinfo{author}{{Eynard Bontemps}, G.},
  \bibinfo{author}{{Fabre}, C.}, \bibinfo{author}{{Fabrizio}, M.},
  \bibinfo{author}{{Faigler}, S.}, \bibinfo{author}{{Falc{\~a}o}, A.J.},
  \bibinfo{author}{{Farr{\`a}s Casas}, M.}, \bibinfo{author}{{Faye}, F.},
  \bibinfo{author}{{Federici}, L.}, \bibinfo{author}{{Fedorets}, G.},
  \bibinfo{author}{{Fern{\'a}ndez-Hern{\'a}ndez}, J.},
  \bibinfo{author}{{Fernique}, P.}, \bibinfo{author}{{Fienga}, A.},
  \bibinfo{author}{{Figueras}, F.}, \bibinfo{author}{{Filippi}, F.},
  \bibinfo{author}{{Findeisen}, K.}, \bibinfo{author}{{Fonti}, A.},
  \bibinfo{author}{{Fouesneau}, M.}, \bibinfo{author}{{Fraile}, E.},
  \bibinfo{author}{{Fraser}, M.}, \bibinfo{author}{{Fuchs}, J.},
  \bibinfo{author}{{Furnell}, R.}, \bibinfo{author}{{Gai}, M.},
  \bibinfo{author}{{Galleti}, S.}, \bibinfo{author}{{Galluccio}, L.},
  \bibinfo{author}{{Garabato}, D.}, \bibinfo{author}{{Garc{\'\i}a-Sedano}, F.},
  \bibinfo{author}{{Gar{\'e}}, P.}, \bibinfo{author}{{Garofalo}, A.},
  \bibinfo{author}{{Garralda}, N.}, \bibinfo{author}{{Gavras}, P.},
  \bibinfo{author}{{Gerssen}, J.}, \bibinfo{author}{{Geyer}, R.},
  \bibinfo{author}{{Gilmore}, G.}, \bibinfo{author}{{Girona}, S.},
  \bibinfo{author}{{Giuffrida}, G.}, \bibinfo{author}{{Gomes}, M.},
  \bibinfo{author}{{Gonz{\'a}lez-Marcos}, A.},
  \bibinfo{author}{{Gonz{\'a}lez-N{\'u}{\~n}ez}, J.},
  \bibinfo{author}{{Gonz{\'a}lez-Vidal}, J.J.}, \bibinfo{author}{{Granvik},
  M.}, \bibinfo{author}{{Guerrier}, A.}, \bibinfo{author}{{Guillout}, P.},
  \bibinfo{author}{{Guiraud}, J.}, \bibinfo{author}{{G{\'u}rpide}, A.},
  \bibinfo{author}{{Guti{\'e}rrez-S{\'a}nchez}, R.}, \bibinfo{author}{{Guy},
  L.P.}, \bibinfo{author}{{Haigron}, R.}, \bibinfo{author}{{Hatzidimitriou},
  D.}, \bibinfo{author}{{Haywood}, M.}, \bibinfo{author}{{Heiter}, U.},
  \bibinfo{author}{{Helmi}, A.}, \bibinfo{author}{{Hobbs}, D.},
  \bibinfo{author}{{Hofmann}, W.}, \bibinfo{author}{{Holl}, B.},
  \bibinfo{author}{{Holland }, G.}, \bibinfo{author}{{Hunt}, J.A.S.},
  \bibinfo{author}{{Hypki}, A.}, \bibinfo{author}{{Icardi}, V.},
  \bibinfo{author}{{Irwin}, M.}, \bibinfo{author}{{Jevardat de Fombelle}, G.},
  \bibinfo{author}{{Jofr{\'e}}, P.}, \bibinfo{author}{{Jonker}, P.G.},
  \bibinfo{author}{{Jorissen}, A.}, \bibinfo{author}{{Julbe}, F.},
  \bibinfo{author}{{Karampelas}, A.}, \bibinfo{author}{{Kochoska}, A.},
  \bibinfo{author}{{Kohley}, R.}, \bibinfo{author}{{Kolenberg}, K.},
  \bibinfo{author}{{Kontizas}, E.}, \bibinfo{author}{{Koposov}, S.E.},
  \bibinfo{author}{{Kordopatis}, G.}, \bibinfo{author}{{Koubsky}, P.},
  \bibinfo{author}{{Kowalczyk}, A.}, \bibinfo{author}{{Krone-Martins}, A.},
  \bibinfo{author}{{Kudryashova}, M.}, \bibinfo{author}{{Kull}, I.},
  \bibinfo{author}{{Bachchan}, R.K.}, \bibinfo{author}{{Lacoste-Seris}, F.},
  \bibinfo{author}{{Lanza}, A.F.}, \bibinfo{author}{{Lavigne}, J.B.},
  \bibinfo{author}{{Le Poncin-Lafitte}, C.}, \bibinfo{author}{{Lebreton}, Y.},
  \bibinfo{author}{{Lebzelter}, T.}, \bibinfo{author}{{Leccia}, S.},
  \bibinfo{author}{{Leclerc}, N.}, \bibinfo{author}{{Lecoeur-Taibi}, I.},
  \bibinfo{author}{{Lemaitre}, V.}, \bibinfo{author}{{Lenhardt}, H.},
  \bibinfo{author}{{Leroux}, F.}, \bibinfo{author}{{Liao}, S.},
  \bibinfo{author}{{Licata}, E.}, \bibinfo{author}{{Lindstr{\o}m}, H.E.P.},
  \bibinfo{author}{{Lister}, T.A.}, \bibinfo{author}{{Livanou}, E.},
  \bibinfo{author}{{Lobel}, A.}, \bibinfo{author}{{L{\"o}ffler}, W.},
  \bibinfo{author}{{L{\'o}pez}, M.}, \bibinfo{author}{{Lopez-Lozano}, A.},
  \bibinfo{author}{{Lorenz}, D.}, \bibinfo{author}{{Loureiro}, T.},
  \bibinfo{author}{{MacDonald}, I.}, \bibinfo{author}{{Magalh{\~a}es
  Fernandes}, T.}, \bibinfo{author}{{Managau}, S.}, \bibinfo{author}{{Mann},
  R.G.}, \bibinfo{author}{{Mantelet}, G.}, \bibinfo{author}{{Marchal}, O.},
  \bibinfo{author}{{Marchant}, J.M.}, \bibinfo{author}{{Marconi}, M.},
  \bibinfo{author}{{Marie}, J.}, \bibinfo{author}{{Marinoni}, S.},
  \bibinfo{author}{{Marrese}, P.M.}, \bibinfo{author}{{Marschalk{\'o}}, G.},
  \bibinfo{author}{{Marshall}, D.J.}, \bibinfo{author}{{Mart{\'\i}n-Fleitas},
  J.M.}, \bibinfo{author}{{Martino}, M.}, \bibinfo{author}{{Mary}, N.},
  \bibinfo{author}{{Matijevi{\v{c}}}, G.}, \bibinfo{author}{{Mazeh}, T.},
  \bibinfo{author}{{McMillan}, P.J.}, \bibinfo{author}{{Messina}, S.},
  \bibinfo{author}{{Mestre}, A.}, \bibinfo{author}{{Michalik}, D.},
  \bibinfo{author}{{Millar}, N.R.}, \bibinfo{author}{{Miranda}, B.M.H.},
  \bibinfo{author}{{Molina}, D.}, \bibinfo{author}{{Molinaro}, R.},
  \bibinfo{author}{{Molinaro}, M.}, \bibinfo{author}{{Moln{\'a}r}, L.},
  \bibinfo{author}{{Moniez}, M.}, \bibinfo{author}{{Montegriffo}, P.},
  \bibinfo{author}{{Monteiro}, D.}, \bibinfo{author}{{Mor}, R.},
  \bibinfo{author}{{Mora}, A.}, \bibinfo{author}{{Morbidelli}, R.},
  \bibinfo{author}{{Morel}, T.}, \bibinfo{author}{{Morgenthaler}, S.},
  \bibinfo{author}{{Morley}, T.}, \bibinfo{author}{{Morris}, D.},
  \bibinfo{author}{{Mulone}, A.F.}, \bibinfo{author}{{Muraveva}, T.},
  \bibinfo{author}{{Musella}, I.}, \bibinfo{author}{{Narbonne}, J.},
  \bibinfo{author}{{Nelemans}, G.}, \bibinfo{author}{{Nicastro}, L.},
  \bibinfo{author}{{Noval}, L.}, \bibinfo{author}{{Ord{\'e}novic}, C.},
  \bibinfo{author}{{Ordieres-Mer{\'e}}, J.}, \bibinfo{author}{{Osborne}, P.},
  \bibinfo{author}{{Pagani}, C.}, \bibinfo{author}{{Pagano}, I.},
  \bibinfo{author}{{Pailler}, F.}, \bibinfo{author}{{Palacin}, H.},
  \bibinfo{author}{{Palaversa}, L.}, \bibinfo{author}{{Parsons}, P.},
  \bibinfo{author}{{Paulsen}, T.}, \bibinfo{author}{{Pecoraro}, M.},
  \bibinfo{author}{{Pedrosa}, R.}, \bibinfo{author}{{Pentik{\"a}inen}, H.},
  \bibinfo{author}{{Pereira}, J.}, \bibinfo{author}{{Pichon}, B.},
  \bibinfo{author}{{Piersimoni}, A.M.}, \bibinfo{author}{{Pineau}, F.X.},
  \bibinfo{author}{{Plachy}, E.}, \bibinfo{author}{{Plum}, G.},
  \bibinfo{author}{{Poujoulet}, E.}, \bibinfo{author}{{Pr{\v{s}}a}, A.},
  \bibinfo{author}{{Pulone}, L.}, \bibinfo{author}{{Ragaini}, S.},
  \bibinfo{author}{{Rago}, S.}, \bibinfo{author}{{Rambaux}, N.},
  \bibinfo{author}{{Ramos-Lerate}, M.}, \bibinfo{author}{{Ranalli}, P.},
  \bibinfo{author}{{Rauw}, G.}, \bibinfo{author}{{Read}, A.},
  \bibinfo{author}{{Regibo}, S.}, \bibinfo{author}{{Renk}, F.},
  \bibinfo{author}{{Reyl{\'e}}, C.}, \bibinfo{author}{{Ribeiro}, R.A.},
  \bibinfo{author}{{Rimoldini}, L.}, \bibinfo{author}{{Ripepi}, V.},
  \bibinfo{author}{{Riva}, A.}, \bibinfo{author}{{Rixon}, G.},
  \bibinfo{author}{{Roelens}, M.}, \bibinfo{author}{{Romero-G{\'o}mez}, M.},
  \bibinfo{author}{{Rowell}, N.}, \bibinfo{author}{{Royer}, F.},
  \bibinfo{author}{{Rudolph}, A.}, \bibinfo{author}{{Ruiz-Dern}, L.},
  \bibinfo{author}{{Sadowski}, G.}, \bibinfo{author}{{Sagrist{\`a} Sell{\'e}s},
  T.}, \bibinfo{author}{{Sahlmann}, J.}, \bibinfo{author}{{Salgado}, J.},
  \bibinfo{author}{{Salguero}, E.}, \bibinfo{author}{{Sarasso}, M.},
  \bibinfo{author}{{Savietto}, H.}, \bibinfo{author}{{Schnorhk}, A.},
  \bibinfo{author}{{Schultheis}, M.}, \bibinfo{author}{{Sciacca}, E.},
  \bibinfo{author}{{Segol}, M.}, \bibinfo{author}{{Segovia}, J.C.},
  \bibinfo{author}{{Segransan}, D.}, \bibinfo{author}{{Serpell}, E.},
  \bibinfo{author}{{Shih}, I.C.}, \bibinfo{author}{{Smareglia}, R.},
  \bibinfo{author}{{Smart}, R.L.}, \bibinfo{author}{{Smith}, C.},
  \bibinfo{author}{{Solano}, E.}, \bibinfo{author}{{Solitro}, F.},
  \bibinfo{author}{{Sordo}, R.}, \bibinfo{author}{{Soria Nieto}, S.},
  \bibinfo{author}{{Souchay}, J.}, \bibinfo{author}{{Spagna}, A.},
  \bibinfo{author}{{Spoto}, F.}, \bibinfo{author}{{Stampa}, U.},
  \bibinfo{author}{{Steele}, I.A.}, \bibinfo{author}{{Steidelm{\"u}ller}, H.},
  \bibinfo{author}{{Stephenson}, C.A.}, \bibinfo{author}{{Stoev}, H.},
  \bibinfo{author}{{Suess}, F.F.}, \bibinfo{author}{{S{\"u}veges}, M.},
  \bibinfo{author}{{Surdej}, J.}, \bibinfo{author}{{Szabados}, L.},
  \bibinfo{author}{{Szegedi-Elek}, E.}, \bibinfo{author}{{Tapiador}, D.},
  \bibinfo{author}{{Taris}, F.}, \bibinfo{author}{{Tauran}, G.},
  \bibinfo{author}{{Taylor}, M.B.}, \bibinfo{author}{{Teixeira}, R.},
  \bibinfo{author}{{Terrett}, D.}, \bibinfo{author}{{Tingley}, B.},
  \bibinfo{author}{{Trager}, S.C.}, \bibinfo{author}{{Turon}, C.},
  \bibinfo{author}{{Ulla}, A.}, \bibinfo{author}{{Utrilla}, E.},
  \bibinfo{author}{{Valentini}, G.}, \bibinfo{author}{{van Elteren}, A.},
  \bibinfo{author}{{Van Hemelryck}, E.}, \bibinfo{author}{{van Leeuwen}, M.},
  \bibinfo{author}{{Varadi}, M.}, \bibinfo{author}{{Vecchiato}, A.},
  \bibinfo{author}{{Veljanoski}, J.}, \bibinfo{author}{{Via}, T.},
  \bibinfo{author}{{Vicente}, D.}, \bibinfo{author}{{Vogt}, S.},
  \bibinfo{author}{{Voss}, H.}, \bibinfo{author}{{Votruba}, V.},
  \bibinfo{author}{{Voutsinas}, S.}, \bibinfo{author}{{Walmsley}, G.},
  \bibinfo{author}{{Weiler}, M.}, \bibinfo{author}{{Weingrill}, K.},
  \bibinfo{author}{{Werner}, D.}, \bibinfo{author}{{Wevers}, T.},
  \bibinfo{author}{{Whitehead}, G.}, \bibinfo{author}{{Wyrzykowski}, {\L}.},
  \bibinfo{author}{{Yoldas}, A.}, \bibinfo{author}{{{\v{Z}}erjal}, M.},
  \bibinfo{author}{{Zucker}, S.}, \bibinfo{author}{{Zurbach}, C.},
  \bibinfo{author}{{Zwitter}, T.}, \bibinfo{author}{{Alecu}, A.},
  \bibinfo{author}{{Allen}, M.}, \bibinfo{author}{{Allende Prieto}, C.},
  \bibinfo{author}{{Amorim}, A.}, \bibinfo{author}{{Anglada-Escud{\'e}}, G.},
  \bibinfo{author}{{Arsenijevic}, V.}, \bibinfo{author}{{Azaz}, S.},
  \bibinfo{author}{{Balm}, P.}, \bibinfo{author}{{Beck}, M.},
  \bibinfo{author}{{Bernstein}, H.H.}, \bibinfo{author}{{Bigot}, L.},
  \bibinfo{author}{{Bijaoui}, A.}, \bibinfo{author}{{Blasco}, C.},
  \bibinfo{author}{{Bonfigli}, M.}, \bibinfo{author}{{Bono}, G.},
  \bibinfo{author}{{Boudreault}, S.}, \bibinfo{author}{{Bressan}, A.},
  \bibinfo{author}{{Brown}, S.}, \bibinfo{author}{{Brunet}, P.M.},
  \bibinfo{author}{{Bunclark}, P.}, \bibinfo{author}{{Buonanno}, R.},
  \bibinfo{author}{{Butkevich}, A.G.}, \bibinfo{author}{{Carret}, C.},
  \bibinfo{author}{{Carrion}, C.}, \bibinfo{author}{{Chemin}, L.},
  \bibinfo{author}{{Ch{\'e}reau}, F.}, \bibinfo{author}{{Corcione}, L.},
  \bibinfo{author}{{Darmigny}, E.}, \bibinfo{author}{{de Boer}, K.S.},
  \bibinfo{author}{{de Teodoro}, P.}, \bibinfo{author}{{de Zeeuw}, P.T.},
  \bibinfo{author}{{Delle Luche}, C.}, \bibinfo{author}{{Domingues}, C.D.},
  \bibinfo{author}{{Dubath}, P.}, \bibinfo{author}{{Fodor}, F.},
  \bibinfo{author}{{Fr{\'e}zouls}, B.}, \bibinfo{author}{{Fries}, A.},
  \bibinfo{author}{{Fustes}, D.}, \bibinfo{author}{{Fyfe}, D.},
  \bibinfo{author}{{Gallardo}, E.}, \bibinfo{author}{{Gallegos}, J.},
  \bibinfo{author}{{Gardiol}, D.}, \bibinfo{author}{{Gebran}, M.},
  \bibinfo{author}{{Gomboc}, A.}, \bibinfo{author}{{G{\'o}mez}, A.},
  \bibinfo{author}{{Grux}, E.}, \bibinfo{author}{{Gueguen}, A.},
  \bibinfo{author}{{Heyrovsky}, A.}, \bibinfo{author}{{Hoar}, J.},
  \bibinfo{author}{{Iannicola}, G.}, \bibinfo{author}{{Isasi Parache}, Y.},
  \bibinfo{author}{{Janotto}, A.M.}, \bibinfo{author}{{Joliet}, E.},
  \bibinfo{author}{{Jonckheere}, A.}, \bibinfo{author}{{Keil}, R.},
  \bibinfo{author}{{Kim}, D.W.}, \bibinfo{author}{{Klagyivik}, P.},
  \bibinfo{author}{{Klar}, J.}, \bibinfo{author}{{Knude}, J.},
  \bibinfo{author}{{Kochukhov}, O.}, \bibinfo{author}{{Kolka}, I.},
  \bibinfo{author}{{Kos}, J.}, \bibinfo{author}{{Kutka}, A.},
  \bibinfo{author}{{Lainey}, V.}, \bibinfo{author}{{LeBouquin}, D.},
  \bibinfo{author}{{Liu}, C.}, \bibinfo{author}{{Loreggia}, D.},
  \bibinfo{author}{{Makarov}, V.V.}, \bibinfo{author}{{Marseille}, M.G.},
  \bibinfo{author}{{Martayan}, C.}, \bibinfo{author}{{Martinez-Rubi}, O.},
  \bibinfo{author}{{Massart}, B.}, \bibinfo{author}{{Meynadier}, F.},
  \bibinfo{author}{{Mignot}, S.}, \bibinfo{author}{{Munari}, U.},
  \bibinfo{author}{{Nguyen}, A.T.}, \bibinfo{author}{{Nordlander}, T.},
  \bibinfo{author}{{Ocvirk}, P.}, \bibinfo{author}{{O'Flaherty}, K.S.},
  \bibinfo{author}{{Olias Sanz}, A.}, \bibinfo{author}{{Ortiz}, P.},
  \bibinfo{author}{{Osorio}, J.}, \bibinfo{author}{{Oszkiewicz}, D.},
  \bibinfo{author}{{Ouzounis}, A.}, \bibinfo{author}{{Palmer}, M.},
  \bibinfo{author}{{Park}, P.}, \bibinfo{author}{{Pasquato}, E.},
  \bibinfo{author}{{Peltzer}, C.}, \bibinfo{author}{{Peralta}, J.},
  \bibinfo{author}{{P{\'e}turaud}, F.}, \bibinfo{author}{{Pieniluoma}, T.},
  \bibinfo{author}{{Pigozzi}, E.}, \bibinfo{author}{{Poels}, J.},
  \bibinfo{author}{{Prat}, G.}, \bibinfo{author}{{Prod'homme}, T.},
  \bibinfo{author}{{Raison}, F.}, \bibinfo{author}{{Rebordao}, J.M.},
  \bibinfo{author}{{Risquez}, D.}, \bibinfo{author}{{Rocca-Volmerange}, B.},
  \bibinfo{author}{{Rosen}, S.}, \bibinfo{author}{{Ruiz-Fuertes}, M.I.},
  \bibinfo{author}{{Russo}, F.}, \bibinfo{author}{{Sembay}, S.},
  \bibinfo{author}{{Serraller Vizcaino}, I.}, \bibinfo{author}{{Short}, A.},
  \bibinfo{author}{{Siebert}, A.}, \bibinfo{author}{{Silva}, H.},
  \bibinfo{author}{{Sinachopoulos}, D.}, \bibinfo{author}{{Slezak}, E.},
  \bibinfo{author}{{Soffel}, M.}, \bibinfo{author}{{Sosnowska}, D.},
  \bibinfo{author}{{Strai{\v{z}}ys}, V.}, \bibinfo{author}{{ter Linden}, M.},
  \bibinfo{author}{{Terrell}, D.}, \bibinfo{author}{{Theil}, S.},
  \bibinfo{author}{{Tiede}, C.}, \bibinfo{author}{{Troisi}, L.},
  \bibinfo{author}{{Tsalmantza}, P.}, \bibinfo{author}{{Tur}, D.},
  \bibinfo{author}{{Vaccari}, M.}, \bibinfo{author}{{Vachier}, F.},
  \bibinfo{author}{{Valles}, P.}, \bibinfo{author}{{Van Hamme}, W.},
  \bibinfo{author}{{Veltz}, L.}, \bibinfo{author}{{Virtanen}, J.},
  \bibinfo{author}{{Wallut}, J.M.}, \bibinfo{author}{{Wichmann}, R.},
  \bibinfo{author}{{Wilkinson}, M.I.}, \bibinfo{author}{{Ziaeepour}, H.},
  \bibinfo{author}{{Zschocke}, S.}, \bibinfo{year}{2016}.
\newblock \bibinfo{title}{{The Gaia mission}}.
\newblock \bibinfo{journal}{\aap} \bibinfo{volume}{595}, \bibinfo{pages}{A1}.
\newblock \DOIprefix\doi{10.1051/0004-6361/201629272},
  \href{http://arxiv.org/abs/1609.04153}{\tt arXiv:1609.04153}.
\bibitem[{{He} and {Garcia}(2009)}]{He:2009}
\bibinfo{author}{{He}, H.}, \bibinfo{author}{{Garcia}, E.A.},
  \bibinfo{year}{2009}.
\newblock \bibinfo{title}{Learning from imbalanced data}.
\newblock \bibinfo{journal}{IEEE Transactions on knowledge and data
  engineering} \bibinfo{volume}{21}, \bibinfo{pages}{1263--1284}.
\bibitem[{Hornik et~al.(1989)Hornik, Stinchcombe and White}]{Hornik:1989}
\bibinfo{author}{Hornik, K.}, \bibinfo{author}{Stinchcombe, M.},
  \bibinfo{author}{White, H.}, \bibinfo{year}{1989}.
\newblock \bibinfo{title}{Multilayer feedforward networks are universal
  approximators}.
\newblock \bibinfo{journal}{Neural networks} \bibinfo{volume}{2},
  \bibinfo{pages}{359--366}.
\newblock \DOIprefix\doi{10.1016/0893-6080(89)90020-8}.
\bibitem[{Hunter(2007)}]{matplotlib}
\bibinfo{author}{Hunter, J.D.}, \bibinfo{year}{2007}.
\newblock \bibinfo{title}{Matplotlib: A 2d graphics environment}.
\newblock \bibinfo{journal}{Computing in Science \& Engineering}
  \bibinfo{volume}{9}, \bibinfo{pages}{90--95}.
\newblock \DOIprefix\doi{10.1109/MCSE.2007.55}.
\bibitem[{Ivezi{\'c} et~al.(2019)Ivezi{\'c}, Connolly, VanderPlas and
  Gray}]{Ivezic:2019book}
\bibinfo{author}{Ivezi{\'c}, {\v{Z}}.}, \bibinfo{author}{Connolly, A.J.},
  \bibinfo{author}{VanderPlas, J.T.}, \bibinfo{author}{Gray, A.},
  \bibinfo{year}{2019}.
\newblock \bibinfo{title}{Statistics, data mining, and machine learning in
  astronomy: A practical python guide for the analysis of survey data}.
\newblock \bibinfo{publisher}{Princeton University Press}.
\bibitem[{{Ivezi\'{c}} et~al.(2019){Ivezi\'{c}}, {Kahn}, {Tyson}, {Abel},
  {Acosta}, {Allsman}, {Alonso}, {AlSayyad}, {Anderson}, {Andrew}, {Angel},
  {Angeli}, {Ansari}, {Antilogus}, {Araujo}, {Armstrong}, {Arndt}, {Astier},
  {Aubourg}, {Auza}, {Axelrod}, {Bard}, {Barr}, {Barrau}, {Bartlett}, {Bauer},
  {Bauman}, {Baumont}, {Bechtol}, {Bechtol}, {Becker}, {Becla}, {Beldica},
  {Bellavia}, {Bianco}, {Biswas}, {Blanc}, {Blazek}, {Bland ford}, {Bloom},
  {Bogart}, {Bond}, {Booth}, {Borgland}, {Borne}, {Bosch}, {Boutigny},
  {Brackett}, {Bradshaw}, {Brand t}, {Brown}, {Bullock}, {Burchat}, {Burke},
  {Cagnoli}, {Calabrese}, {Callahan}, {Callen}, {Carlin}, {Carlson}, {Chand
  rasekharan}, {Charles-Emerson}, {Chesley}, {Cheu}, {Chiang}, {Chiang},
  {Chirino}, {Chow}, {Ciardi}, {Claver}, {Cohen-Tanugi}, {Cockrum}, {Coles},
  {Connolly}, {Cook}, {Cooray}, {Covey}, {Cribbs}, {Cui}, {Cutri}, {Daly},
  {Daniel}, {Daruich}, {Daubard}, {Daues}, {Dawson}, {Delgado}, {Dellapenna},
  {de Peyster}, {de Val-Borro}, {Digel}, {Doherty}, {Dubois},
  {Dubois-Felsmann}, {Durech}, {Economou}, {Eifler}, {Eracleous}, {Emmons},
  {Fausti Neto}, {Ferguson}, {Figueroa}, {Fisher-Levine}, {Focke}, {Foss},
  {Frank}, {Freemon}, {Gangler}, {Gawiser}, {Geary}, {Gee}, {Geha}, {Gessner},
  {Gibson}, {Gilmore}, {Glanzman}, {Glick}, {Goldina}, {Goldstein}, {Goodenow},
  {Graham}, {Gressler}, {Gris}, {Guy}, {Guyonnet}, {Haller}, {Harris},
  {Hascall}, {Haupt}, {Hernand ez}, {Herrmann}, {Hileman}, {Hoblitt},
  {Hodgson}, {Hogan}, {Howard}, {Huang}, {Huffer}, {Ingraham}, {Innes},
  {Jacoby}, {Jain}, {Jammes}, {Jee}, {Jenness}, {Jernigan}, {Jevremovi{\'c}},
  {Johns}, {Johnson}, {Johnson}, {Jones}, {Juramy-Gilles}, {Juri{\'c}},
  {Kalirai}, {Kallivayalil}, {Kalmbach}, {Kantor}, {Karst}, {Kasliwal},
  {Kelly}, {Kessler}, {Kinnison}, {Kirkby}, {Knox}, {Kotov}, {Krabbendam},
  {Krughoff}, {Kub{\'a}nek}, {Kuczewski}, {Kulkarni}, {Ku}, {Kurita}, {Lage},
  {Lambert}, {Lange}, {Langton}, {Le Guillou}, {Levine}, {Liang}, {Lim},
  {Lintott}, {Long}, {Lopez}, {Lotz}, {Lupton}, {Lust}, {MacArthur}, {Mahabal},
  {Mand elbaum}, {Markiewicz}, {Marsh}, {Marshall}, {Marshall}, {May},
  {McKercher}, {McQueen}, {Meyers}, {Migliore}, {Miller}, {Mills}, {Miraval},
  {Moeyens}, {Moolekamp}, {Monet}, {Moniez}, {Monkewitz}, {Montgomery},
  {Morrison}, {Mueller}, {Muller}, {Mu{\~n}oz Arancibia}, {Neill}, {Newbry},
  {Nief}, {Nomerotski}, {Nordby}, {O'Connor}, {Oliver}, {Olivier}, {Olsen},
  {O'Mullane}, {Ortiz}, {Osier}, {Owen}, {Pain}, {Palecek}, {Parejko},
  {Parsons}, {Pease}, {Peterson}, {Peterson}, {Petravick}, {Libby Petrick},
  {Petry}, {Pierfederici}, {Pietrowicz}, {Pike}, {Pinto}, {Plante}, {Plate},
  {Plutchak}, {Price}, {Prouza}, {Radeka}, {Rajagopal}, {Rasmussen},
  {Regnault}, {Reil}, {Reiss}, {Reuter}, {Ridgway}, {Riot}, {Ritz}, {Robinson},
  {Roby}, {Roodman}, {Rosing}, {Roucelle}, {Rumore}, {Russo}, {Saha},
  {Sassolas}, {Schalk}, {Schellart}, {Schindler}, {Schmidt}, {Schneider},
  {Schneider}, {Schoening}, {Schumacher}, {Schwamb}, {Sebag}, {Selvy},
  {Sembroski}, {Seppala}, {Serio}, {Serrano}, {Shaw}, {Shipsey}, {Sick},
  {Silvestri}, {Slater}, {Smith}, {Smith}, {Sobhani}, {Soldahl},
  {Storrie-Lombardi}, {Stover}, {Strauss}, {Street}, {Stubbs}, {Sullivan},
  {Sweeney}, {Swinbank}, {Szalay}, {Takacs}, {Tether}, {Thaler}, {Thayer},
  {Thomas}, {Thornton}, {Thukral}, {Tice}, {Trilling}, {Turri}, {Van Berg},
  {Vanden Berk}, {Vetter}, {Virieux}, {Vucina}, {Wahl}, {Walkowicz}, {Walsh},
  {Walter}, {Wang}, {Wang}, {Warner}, {Wiecha}, {Willman}, {Winters},
  {Wittman}, {Wolff}, {Wood-Vasey}, {Wu}, {Xin}, {Yoachim} and
  {Zhan}}]{Ivezic:2019}
\bibinfo{author}{{Ivezi\'{c}}, {\v{Z}}.}, \bibinfo{author}{{Kahn}, S.M.},
  \bibinfo{author}{{Tyson}, J.A.}, \bibinfo{author}{{Abel}, B.},
  \bibinfo{author}{{Acosta}, E.}, \bibinfo{author}{{Allsman}, R.},
  \bibinfo{author}{{Alonso}, D.}, \bibinfo{author}{{AlSayyad}, Y.},
  \bibinfo{author}{{Anderson}, S.F.}, \bibinfo{author}{{Andrew}, J.},
  \bibinfo{author}{{Angel}, J.R.P.}, \bibinfo{author}{{Angeli}, G.Z.},
  \bibinfo{author}{{Ansari}, R.}, \bibinfo{author}{{Antilogus}, P.},
  \bibinfo{author}{{Araujo}, C.}, \bibinfo{author}{{Armstrong}, R.},
  \bibinfo{author}{{Arndt}, K.T.}, \bibinfo{author}{{Astier}, P.},
  \bibinfo{author}{{Aubourg}, E.}, \bibinfo{author}{{Auza}, N.},
  \bibinfo{author}{{Axelrod}, T.S.}, \bibinfo{author}{{Bard}, D.J.},
  \bibinfo{author}{{Barr}, J.D.}, \bibinfo{author}{{Barrau}, A.},
  \bibinfo{author}{{Bartlett}, J.G.}, \bibinfo{author}{{Bauer}, A.E.},
  \bibinfo{author}{{Bauman}, B.J.}, \bibinfo{author}{{Baumont}, S.},
  \bibinfo{author}{{Bechtol}, E.}, \bibinfo{author}{{Bechtol}, K.},
  \bibinfo{author}{{Becker}, A.C.}, \bibinfo{author}{{Becla}, J.},
  \bibinfo{author}{{Beldica}, C.}, \bibinfo{author}{{Bellavia}, S.},
  \bibinfo{author}{{Bianco}, F.B.}, \bibinfo{author}{{Biswas}, R.},
  \bibinfo{author}{{Blanc}, G.}, \bibinfo{author}{{Blazek}, J.},
  \bibinfo{author}{{Bland ford}, R.D.}, \bibinfo{author}{{Bloom}, J.S.},
  \bibinfo{author}{{Bogart}, J.}, \bibinfo{author}{{Bond}, T.W.},
  \bibinfo{author}{{Booth}, M.T.}, \bibinfo{author}{{Borgland}, A.W.},
  \bibinfo{author}{{Borne}, K.}, \bibinfo{author}{{Bosch}, J.F.},
  \bibinfo{author}{{Boutigny}, D.}, \bibinfo{author}{{Brackett}, C.A.},
  \bibinfo{author}{{Bradshaw}, A.}, \bibinfo{author}{{Brand t}, W.N.},
  \bibinfo{author}{{Brown}, M.E.}, \bibinfo{author}{{Bullock}, J.S.},
  \bibinfo{author}{{Burchat}, P.}, \bibinfo{author}{{Burke}, D.L.},
  \bibinfo{author}{{Cagnoli}, G.}, \bibinfo{author}{{Calabrese}, D.},
  \bibinfo{author}{{Callahan}, S.}, \bibinfo{author}{{Callen}, A.L.},
  \bibinfo{author}{{Carlin}, J.L.}, \bibinfo{author}{{Carlson}, E.L.},
  \bibinfo{author}{{Chand rasekharan}, S.}, \bibinfo{author}{{Charles-Emerson},
  G.}, \bibinfo{author}{{Chesley}, S.}, \bibinfo{author}{{Cheu}, E.C.},
  \bibinfo{author}{{Chiang}, H.F.}, \bibinfo{author}{{Chiang}, J.},
  \bibinfo{author}{{Chirino}, C.}, \bibinfo{author}{{Chow}, D.},
  \bibinfo{author}{{Ciardi}, D.R.}, \bibinfo{author}{{Claver}, C.F.},
  \bibinfo{author}{{Cohen-Tanugi}, J.}, \bibinfo{author}{{Cockrum}, J.J.},
  \bibinfo{author}{{Coles}, R.}, \bibinfo{author}{{Connolly}, A.J.},
  \bibinfo{author}{{Cook}, K.H.}, \bibinfo{author}{{Cooray}, A.},
  \bibinfo{author}{{Covey}, K.R.}, \bibinfo{author}{{Cribbs}, C.},
  \bibinfo{author}{{Cui}, W.}, \bibinfo{author}{{Cutri}, R.},
  \bibinfo{author}{{Daly}, P.N.}, \bibinfo{author}{{Daniel}, S.F.},
  \bibinfo{author}{{Daruich}, F.}, \bibinfo{author}{{Daubard}, G.},
  \bibinfo{author}{{Daues}, G.}, \bibinfo{author}{{Dawson}, W.},
  \bibinfo{author}{{Delgado}, F.}, \bibinfo{author}{{Dellapenna}, A.},
  \bibinfo{author}{{de Peyster}, R.}, \bibinfo{author}{{de Val-Borro}, M.},
  \bibinfo{author}{{Digel}, S.W.}, \bibinfo{author}{{Doherty}, P.},
  \bibinfo{author}{{Dubois}, R.}, \bibinfo{author}{{Dubois-Felsmann}, G.P.},
  \bibinfo{author}{{Durech}, J.}, \bibinfo{author}{{Economou}, F.},
  \bibinfo{author}{{Eifler}, T.}, \bibinfo{author}{{Eracleous}, M.},
  \bibinfo{author}{{Emmons}, B.L.}, \bibinfo{author}{{Fausti Neto}, A.},
  \bibinfo{author}{{Ferguson}, H.}, \bibinfo{author}{{Figueroa}, E.},
  \bibinfo{author}{{Fisher-Levine}, M.}, \bibinfo{author}{{Focke}, W.},
  \bibinfo{author}{{Foss}, M.D.}, \bibinfo{author}{{Frank}, J.},
  \bibinfo{author}{{Freemon}, M.D.}, \bibinfo{author}{{Gangler}, E.},
  \bibinfo{author}{{Gawiser}, E.}, \bibinfo{author}{{Geary}, J.C.},
  \bibinfo{author}{{Gee}, P.}, \bibinfo{author}{{Geha}, M.},
  \bibinfo{author}{{Gessner}, C.J.B.}, \bibinfo{author}{{Gibson}, R.R.},
  \bibinfo{author}{{Gilmore}, D.K.}, \bibinfo{author}{{Glanzman}, T.},
  \bibinfo{author}{{Glick}, W.}, \bibinfo{author}{{Goldina}, T.},
  \bibinfo{author}{{Goldstein}, D.A.}, \bibinfo{author}{{Goodenow}, I.},
  \bibinfo{author}{{Graham}, M.L.}, \bibinfo{author}{{Gressler}, W.J.},
  \bibinfo{author}{{Gris}, P.}, \bibinfo{author}{{Guy}, L.P.},
  \bibinfo{author}{{Guyonnet}, A.}, \bibinfo{author}{{Haller}, G.},
  \bibinfo{author}{{Harris}, R.}, \bibinfo{author}{{Hascall}, P.A.},
  \bibinfo{author}{{Haupt}, J.}, \bibinfo{author}{{Hernand ez}, F.},
  \bibinfo{author}{{Herrmann}, S.}, \bibinfo{author}{{Hileman}, E.},
  \bibinfo{author}{{Hoblitt}, J.}, \bibinfo{author}{{Hodgson}, J.A.},
  \bibinfo{author}{{Hogan}, C.}, \bibinfo{author}{{Howard}, J.D.},
  \bibinfo{author}{{Huang}, D.}, \bibinfo{author}{{Huffer}, M.E.},
  \bibinfo{author}{{Ingraham}, P.}, \bibinfo{author}{{Innes}, W.R.},
  \bibinfo{author}{{Jacoby}, S.H.}, \bibinfo{author}{{Jain}, B.},
  \bibinfo{author}{{Jammes}, F.}, \bibinfo{author}{{Jee}, M.J.},
  \bibinfo{author}{{Jenness}, T.}, \bibinfo{author}{{Jernigan}, G.},
  \bibinfo{author}{{Jevremovi{\'c}}, D.}, \bibinfo{author}{{Johns}, K.},
  \bibinfo{author}{{Johnson}, A.S.}, \bibinfo{author}{{Johnson}, M.W.G.},
  \bibinfo{author}{{Jones}, R.L.}, \bibinfo{author}{{Juramy-Gilles}, C.},
  \bibinfo{author}{{Juri{\'c}}, M.}, \bibinfo{author}{{Kalirai}, J.S.},
  \bibinfo{author}{{Kallivayalil}, N.J.}, \bibinfo{author}{{Kalmbach}, B.},
  \bibinfo{author}{{Kantor}, J.P.}, \bibinfo{author}{{Karst}, P.},
  \bibinfo{author}{{Kasliwal}, M.M.}, \bibinfo{author}{{Kelly}, H.},
  \bibinfo{author}{{Kessler}, R.}, \bibinfo{author}{{Kinnison}, V.},
  \bibinfo{author}{{Kirkby}, D.}, \bibinfo{author}{{Knox}, L.},
  \bibinfo{author}{{Kotov}, I.V.}, \bibinfo{author}{{Krabbendam}, V.L.},
  \bibinfo{author}{{Krughoff}, K.S.}, \bibinfo{author}{{Kub{\'a}nek}, P.},
  \bibinfo{author}{{Kuczewski}, J.}, \bibinfo{author}{{Kulkarni}, S.},
  \bibinfo{author}{{Ku}, J.}, \bibinfo{author}{{Kurita}, N.R.},
  \bibinfo{author}{{Lage}, C.S.}, \bibinfo{author}{{Lambert}, R.},
  \bibinfo{author}{{Lange}, T.}, \bibinfo{author}{{Langton}, J.B.},
  \bibinfo{author}{{Le Guillou}, L.}, \bibinfo{author}{{Levine}, D.},
  \bibinfo{author}{{Liang}, M.}, \bibinfo{author}{{Lim}, K.T.},
  \bibinfo{author}{{Lintott}, C.J.}, \bibinfo{author}{{Long}, K.E.},
  \bibinfo{author}{{Lopez}, M.}, \bibinfo{author}{{Lotz}, P.J.},
  \bibinfo{author}{{Lupton}, R.H.}, \bibinfo{author}{{Lust}, N.B.},
  \bibinfo{author}{{MacArthur}, L.A.}, \bibinfo{author}{{Mahabal}, A.},
  \bibinfo{author}{{Mand elbaum}, R.}, \bibinfo{author}{{Markiewicz}, T.W.},
  \bibinfo{author}{{Marsh}, D.S.}, \bibinfo{author}{{Marshall}, P.J.},
  \bibinfo{author}{{Marshall}, S.}, \bibinfo{author}{{May}, M.},
  \bibinfo{author}{{McKercher}, R.}, \bibinfo{author}{{McQueen}, M.},
  \bibinfo{author}{{Meyers}, J.}, \bibinfo{author}{{Migliore}, M.},
  \bibinfo{author}{{Miller}, M.}, \bibinfo{author}{{Mills}, D.J.},
  \bibinfo{author}{{Miraval}, C.}, \bibinfo{author}{{Moeyens}, J.},
  \bibinfo{author}{{Moolekamp}, F.E.}, \bibinfo{author}{{Monet}, D.G.},
  \bibinfo{author}{{Moniez}, M.}, \bibinfo{author}{{Monkewitz}, S.},
  \bibinfo{author}{{Montgomery}, C.}, \bibinfo{author}{{Morrison}, C.B.},
  \bibinfo{author}{{Mueller}, F.}, \bibinfo{author}{{Muller}, G.P.},
  \bibinfo{author}{{Mu{\~n}oz Arancibia}, F.}, \bibinfo{author}{{Neill}, D.R.},
  \bibinfo{author}{{Newbry}, S.P.}, \bibinfo{author}{{Nief}, J.Y.},
  \bibinfo{author}{{Nomerotski}, A.}, \bibinfo{author}{{Nordby}, M.},
  \bibinfo{author}{{O'Connor}, P.}, \bibinfo{author}{{Oliver}, J.},
  \bibinfo{author}{{Olivier}, S.S.}, \bibinfo{author}{{Olsen}, K.},
  \bibinfo{author}{{O'Mullane}, W.}, \bibinfo{author}{{Ortiz}, S.},
  \bibinfo{author}{{Osier}, S.}, \bibinfo{author}{{Owen}, R.E.},
  \bibinfo{author}{{Pain}, R.}, \bibinfo{author}{{Palecek}, P.E.},
  \bibinfo{author}{{Parejko}, J.K.}, \bibinfo{author}{{Parsons}, J.B.},
  \bibinfo{author}{{Pease}, N.M.}, \bibinfo{author}{{Peterson}, J.M.},
  \bibinfo{author}{{Peterson}, J.R.}, \bibinfo{author}{{Petravick}, D.L.},
  \bibinfo{author}{{Libby Petrick}, M.E.}, \bibinfo{author}{{Petry}, C.E.},
  \bibinfo{author}{{Pierfederici}, F.}, \bibinfo{author}{{Pietrowicz}, S.},
  \bibinfo{author}{{Pike}, R.}, \bibinfo{author}{{Pinto}, P.A.},
  \bibinfo{author}{{Plante}, R.}, \bibinfo{author}{{Plate}, S.},
  \bibinfo{author}{{Plutchak}, J.P.}, \bibinfo{author}{{Price}, P.A.},
  \bibinfo{author}{{Prouza}, M.}, \bibinfo{author}{{Radeka}, V.},
  \bibinfo{author}{{Rajagopal}, J.}, \bibinfo{author}{{Rasmussen}, A.P.},
  \bibinfo{author}{{Regnault}, N.}, \bibinfo{author}{{Reil}, K.A.},
  \bibinfo{author}{{Reiss}, D.J.}, \bibinfo{author}{{Reuter}, M.A.},
  \bibinfo{author}{{Ridgway}, S.T.}, \bibinfo{author}{{Riot}, V.J.},
  \bibinfo{author}{{Ritz}, S.}, \bibinfo{author}{{Robinson}, S.},
  \bibinfo{author}{{Roby}, W.}, \bibinfo{author}{{Roodman}, A.},
  \bibinfo{author}{{Rosing}, W.}, \bibinfo{author}{{Roucelle}, C.},
  \bibinfo{author}{{Rumore}, M.R.}, \bibinfo{author}{{Russo}, S.},
  \bibinfo{author}{{Saha}, A.}, \bibinfo{author}{{Sassolas}, B.},
  \bibinfo{author}{{Schalk}, T.L.}, \bibinfo{author}{{Schellart}, P.},
  \bibinfo{author}{{Schindler}, R.H.}, \bibinfo{author}{{Schmidt}, S.},
  \bibinfo{author}{{Schneider}, D.P.}, \bibinfo{author}{{Schneider}, M.D.},
  \bibinfo{author}{{Schoening}, W.}, \bibinfo{author}{{Schumacher}, G.},
  \bibinfo{author}{{Schwamb}, M.E.}, \bibinfo{author}{{Sebag}, J.},
  \bibinfo{author}{{Selvy}, B.}, \bibinfo{author}{{Sembroski}, G.H.},
  \bibinfo{author}{{Seppala}, L.G.}, \bibinfo{author}{{Serio}, A.},
  \bibinfo{author}{{Serrano}, E.}, \bibinfo{author}{{Shaw}, R.A.},
  \bibinfo{author}{{Shipsey}, I.}, \bibinfo{author}{{Sick}, J.},
  \bibinfo{author}{{Silvestri}, N.}, \bibinfo{author}{{Slater}, C.T.},
  \bibinfo{author}{{Smith}, J.A.}, \bibinfo{author}{{Smith}, R.C.},
  \bibinfo{author}{{Sobhani}, S.}, \bibinfo{author}{{Soldahl}, C.},
  \bibinfo{author}{{Storrie-Lombardi}, L.}, \bibinfo{author}{{Stover}, E.},
  \bibinfo{author}{{Strauss}, M.A.}, \bibinfo{author}{{Street}, R.A.},
  \bibinfo{author}{{Stubbs}, C.W.}, \bibinfo{author}{{Sullivan}, I.S.},
  \bibinfo{author}{{Sweeney}, D.}, \bibinfo{author}{{Swinbank}, J.D.},
  \bibinfo{author}{{Szalay}, A.}, \bibinfo{author}{{Takacs}, P.},
  \bibinfo{author}{{Tether}, S.A.}, \bibinfo{author}{{Thaler}, J.J.},
  \bibinfo{author}{{Thayer}, J.G.}, \bibinfo{author}{{Thomas}, S.},
  \bibinfo{author}{{Thornton}, A.J.}, \bibinfo{author}{{Thukral}, V.},
  \bibinfo{author}{{Tice}, J.}, \bibinfo{author}{{Trilling}, D.E.},
  \bibinfo{author}{{Turri}, M.}, \bibinfo{author}{{Van Berg}, R.},
  \bibinfo{author}{{Vanden Berk}, D.}, \bibinfo{author}{{Vetter}, K.},
  \bibinfo{author}{{Virieux}, F.}, \bibinfo{author}{{Vucina}, T.},
  \bibinfo{author}{{Wahl}, W.}, \bibinfo{author}{{Walkowicz}, L.},
  \bibinfo{author}{{Walsh}, B.}, \bibinfo{author}{{Walter}, C.W.},
  \bibinfo{author}{{Wang}, D.L.}, \bibinfo{author}{{Wang}, S.Y.},
  \bibinfo{author}{{Warner}, M.}, \bibinfo{author}{{Wiecha}, O.},
  \bibinfo{author}{{Willman}, B.}, \bibinfo{author}{{Winters}, S.E.},
  \bibinfo{author}{{Wittman}, D.}, \bibinfo{author}{{Wolff}, S.C.},
  \bibinfo{author}{{Wood-Vasey}, W.M.}, \bibinfo{author}{{Wu}, X.},
  \bibinfo{author}{{Xin}, B.}, \bibinfo{author}{{Yoachim}, P.},
  \bibinfo{author}{{Zhan}, H.}, \bibinfo{year}{2019}.
\newblock \bibinfo{title}{{LSST: From Science Drivers to Reference Design and
  Anticipated Data Products}}.
\newblock \bibinfo{journal}{\apj} \bibinfo{volume}{873}, \bibinfo{pages}{111}.
\newblock \DOIprefix\doi{10.3847/1538-4357/ab042c},
  \href{http://arxiv.org/abs/0805.2366}{\tt arXiv:0805.2366}.
\bibitem[{{King} and {Zeng}(2001)}]{King:2001}
\bibinfo{author}{{King}, G.}, \bibinfo{author}{{Zeng}, L.},
  \bibinfo{year}{2001}.
\newblock \bibinfo{title}{Logistic regression in rare events data}.
\newblock \bibinfo{journal}{Political Analysis} \bibinfo{volume}{9},
  \bibinfo{pages}{137--163}.
\newblock \DOIprefix\doi{10.1093/oxfordjournals.pan.a004868}.
\bibitem[{{Kingma} and {Ba}(2014)}]{adamoptimizer}
\bibinfo{author}{{Kingma}, D.P.}, \bibinfo{author}{{Ba}, J.},
  \bibinfo{year}{2014}.
\newblock \bibinfo{title}{{Adam: A Method for Stochastic Optimization}}.
\newblock \bibinfo{journal}{arXiv e-prints} ,
  \bibinfo{pages}{arXiv:1412.6980}\href{http://arxiv.org/abs/1412.6980}{\tt
  arXiv:1412.6980}.
\bibitem[{{Kov{\'a}cs} and {Szapudi}(2015)}]{Kovacs:2015}
\bibinfo{author}{{Kov{\'a}cs}, A.}, \bibinfo{author}{{Szapudi}, I.},
  \bibinfo{year}{2015}.
\newblock \bibinfo{title}{{Star-galaxy separation strategies for WISE-2MASS
  all-sky infrared galaxy catalogues}}.
\newblock \bibinfo{journal}{\mnras} \bibinfo{volume}{448},
  \bibinfo{pages}{1305--1313}.
\newblock \DOIprefix\doi{10.1093/mnras/stv063},
  \href{http://arxiv.org/abs/1401.0156}{\tt arXiv:1401.0156}.
\bibitem[{{Kubat} et~al.(1998){Kubat}, {Holte} and {Matwin}}]{Kubat:1998}
\bibinfo{author}{{Kubat}, M.}, \bibinfo{author}{{Holte}, R.C.},
  \bibinfo{author}{{Matwin}, S.}, \bibinfo{year}{1998}.
\newblock \bibinfo{title}{Machine learning for the detection of oil spills in
  satellite radar images}.
\newblock \bibinfo{journal}{Machine Learning} \bibinfo{volume}{30},
  \bibinfo{pages}{195--215}.
\newblock \URLprefix \url{https://doi.org/10.1023/A:1007452223027},
  \DOIprefix\doi{10.1023/A:1007452223027}.
\bibitem[{{Lake} and {Wright}(2016)}]{Lake:2016}
\bibinfo{author}{{Lake}, S.E.}, \bibinfo{author}{{Wright}, E.L.},
  \bibinfo{year}{2016}.
\newblock \bibinfo{title}{{K-Corrections: An Examination of Their Contribution
  to the Uncertainty of Luminosity Measurements}}.
\newblock \bibinfo{journal}{The Open Journal of Astrophysics}
  \bibinfo{volume}{1}, \bibinfo{pages}{3}.
\newblock \DOIprefix\doi{10.21105/astro.1603.07299},
  \href{http://arxiv.org/abs/1603.07299}{\tt arXiv:1603.07299}.
\bibitem[{{Lake} et~al.(2018){Lake}, {Wright}, {Assef}, {Jarrett}, {Petty},
  {Stanford}, {Stern} and {Tsai}}]{Lake:2018}
\bibinfo{author}{{Lake}, S.E.}, \bibinfo{author}{{Wright}, E.L.},
  \bibinfo{author}{{Assef}, R.J.}, \bibinfo{author}{{Jarrett}, T.H.},
  \bibinfo{author}{{Petty}, S.}, \bibinfo{author}{{Stanford}, S.A.},
  \bibinfo{author}{{Stern}, D.}, \bibinfo{author}{{Tsai}, C.W.},
  \bibinfo{year}{2018}.
\newblock \bibinfo{title}{{The 2.4 {$\mu$}m Galaxy Luminosity Function as
  Measured Using WISE. II. Sample Selection}}.
\newblock \bibinfo{journal}{\apj} \bibinfo{volume}{866}, \bibinfo{pages}{44}.
\newblock \DOIprefix\doi{10.3847/1538-4357/aae096},
  \href{http://arxiv.org/abs/1702.07828}{\tt arXiv:1702.07828}.
\bibitem[{{Lake} et~al.(2019){Lake}, {Wright}, {Assef}, {Jarrett}, {Petty},
  {Stanford} and {Tsai}}]{Lake:2019}
\bibinfo{author}{{Lake}, S.E.}, \bibinfo{author}{{Wright}, E.L.},
  \bibinfo{author}{{Assef}, R.J.}, \bibinfo{author}{{Jarrett}, T.H.},
  \bibinfo{author}{{Petty}, S.}, \bibinfo{author}{{Stanford}, S.A.},
  \bibinfo{author}{{Tsai}, C.W.}, \bibinfo{year}{2019}.
\newblock \bibinfo{title}{{The Contribution of Galaxies to the 3.4
  {\ensuremath{\mu}}m Cosmic Infrared Background as Measured Using WISE}}.
\newblock \bibinfo{journal}{\apj} \bibinfo{volume}{887}, \bibinfo{pages}{207}.
\newblock \DOIprefix\doi{10.3847/1538-4357/ab5289},
  \href{http://arxiv.org/abs/1908.00731}{\tt arXiv:1908.00731}.
\bibitem[{{Lake} et~al.(2012){Lake}, {Wright}, {Petty}, {Assef}, {Jarrett},
  {Stanford}, {Stern} and {Tsai}}]{Lake:2012}
\bibinfo{author}{{Lake}, S.E.}, \bibinfo{author}{{Wright}, E.L.},
  \bibinfo{author}{{Petty}, S.}, \bibinfo{author}{{Assef}, R.J.},
  \bibinfo{author}{{Jarrett}, T.H.}, \bibinfo{author}{{Stanford}, S.A.},
  \bibinfo{author}{{Stern}, D.}, \bibinfo{author}{{Tsai}, C.W.},
  \bibinfo{year}{2012}.
\newblock \bibinfo{title}{{Optical Spectroscopic Survey of High-latitude
  WISE-selected Sources}}.
\newblock \bibinfo{journal}{\aj} \bibinfo{volume}{143}, \bibinfo{pages}{7}.
\newblock \DOIprefix\doi{10.1088/0004-6256/143/1/7},
  \href{http://arxiv.org/abs/1111.0341}{\tt arXiv:1111.0341}.
\bibitem[{{Lawrence} et~al.(2007){Lawrence}, {Warren}, {Almaini}, {Edge},
  {Hambly}, {Jameson}, {Lucas}, {Casali}, {Adamson}, {Dye}, {Emerson},
  {Foucaud}, {Hewett}, {Hirst}, {Hodgkin}, {Irwin}, {Lodieu}, {McMahon},
  {Simpson}, {Smail}, {Mortlock} and {Folger}}]{Lawrence:2007}
\bibinfo{author}{{Lawrence}, A.}, \bibinfo{author}{{Warren}, S.J.},
  \bibinfo{author}{{Almaini}, O.}, \bibinfo{author}{{Edge}, A.C.},
  \bibinfo{author}{{Hambly}, N.C.}, \bibinfo{author}{{Jameson}, R.F.},
  \bibinfo{author}{{Lucas}, P.}, \bibinfo{author}{{Casali}, M.},
  \bibinfo{author}{{Adamson}, A.}, \bibinfo{author}{{Dye}, S.},
  \bibinfo{author}{{Emerson}, J.P.}, \bibinfo{author}{{Foucaud}, S.},
  \bibinfo{author}{{Hewett}, P.}, \bibinfo{author}{{Hirst}, P.},
  \bibinfo{author}{{Hodgkin}, S.T.}, \bibinfo{author}{{Irwin}, M.J.},
  \bibinfo{author}{{Lodieu}, N.}, \bibinfo{author}{{McMahon}, R.G.},
  \bibinfo{author}{{Simpson}, C.}, \bibinfo{author}{{Smail}, I.},
  \bibinfo{author}{{Mortlock}, D.}, \bibinfo{author}{{Folger}, M.},
  \bibinfo{year}{2007}.
\newblock \bibinfo{title}{{The UKIRT Infrared Deep Sky Survey (UKIDSS)}}.
\newblock \bibinfo{journal}{\mnras} \bibinfo{volume}{379},
  \bibinfo{pages}{1599--1617}.
\newblock \DOIprefix\doi{10.1111/j.1365-2966.2007.12040.x},
  \href{http://arxiv.org/abs/astro-ph/0604426}{\tt arXiv:astro-ph/0604426}.
\bibitem[{{le Cun}(1988)}]{leCun:1988}
\bibinfo{author}{{le Cun}, Y.}, \bibinfo{year}{1988}.
\newblock \bibinfo{title}{A theoretical framework for back-propagation}, in:
  \bibinfo{editor}{{Touresky}, D.}, \bibinfo{editor}{{Hinton}, G.},
  \bibinfo{editor}{{Sejnowski}, T.} (Eds.), \bibinfo{booktitle}{Proceedings of
  the 1988 connectionist models summer school}, pp. \bibinfo{pages}{21--28}.
\bibitem[{{Louppe}(2014)}]{Louppe:2014}
\bibinfo{author}{{Louppe}, G.}, \bibinfo{year}{2014}.
\newblock \bibinfo{title}{{Understanding Random Forests: From Theory to
  Practice}}.
\newblock \bibinfo{journal}{arXiv e-prints} ,
  \bibinfo{pages}{arXiv:1407.7502}\href{http://arxiv.org/abs/1407.7502}{\tt
  arXiv:1407.7502}.
\bibitem[{{Maalouf} et~al.(2018){Maalouf}, {Homouz} and
  {Trafalis}}]{Maalouf:2018}
\bibinfo{author}{{Maalouf}, M.}, \bibinfo{author}{{Homouz}, D.},
  \bibinfo{author}{{Trafalis}, T.B.}, \bibinfo{year}{2018}.
\newblock \bibinfo{title}{Logistic regression in large rare events and
  imbalanced data: A performance comparison of prior correction and weighting
  methods}.
\newblock \bibinfo{journal}{Computational Intelligence} \bibinfo{volume}{34},
  \bibinfo{pages}{161--174}.
\newblock \DOIprefix\doi{10.1111/coin.12123}.
\bibitem[{{Martin} et~al.(2005){Martin}, {Fanson}, {Schiminovich}, {Morrissey},
  {Friedman}, {Barlow}, {Conrow}, {Grange}, {Jelinsky}, {Milliard}, {Siegmund},
  {Bianchi}, {Byun}, {Donas}, {Forster}, {Heckman}, {Lee}, {Madore}, {Malina},
  {Neff}, {Rich}, {Small}, {Surber}, {Szalay}, {Welsh} and
  {Wyder}}]{Martin:2005}
\bibinfo{author}{{Martin}, D.C.}, \bibinfo{author}{{Fanson}, J.},
  \bibinfo{author}{{Schiminovich}, D.}, \bibinfo{author}{{Morrissey}, P.},
  \bibinfo{author}{{Friedman}, P.G.}, \bibinfo{author}{{Barlow}, T.A.},
  \bibinfo{author}{{Conrow}, T.}, \bibinfo{author}{{Grange}, R.},
  \bibinfo{author}{{Jelinsky}, P.N.}, \bibinfo{author}{{Milliard}, B.},
  \bibinfo{author}{{Siegmund}, O.H.W.}, \bibinfo{author}{{Bianchi}, L.},
  \bibinfo{author}{{Byun}, Y.I.}, \bibinfo{author}{{Donas}, J.},
  \bibinfo{author}{{Forster}, K.}, \bibinfo{author}{{Heckman}, T.M.},
  \bibinfo{author}{{Lee}, Y.W.}, \bibinfo{author}{{Madore}, B.F.},
  \bibinfo{author}{{Malina}, R.F.}, \bibinfo{author}{{Neff}, S.G.},
  \bibinfo{author}{{Rich}, R.M.}, \bibinfo{author}{{Small}, T.},
  \bibinfo{author}{{Surber}, F.}, \bibinfo{author}{{Szalay}, A.S.},
  \bibinfo{author}{{Welsh}, B.}, \bibinfo{author}{{Wyder}, T.K.},
  \bibinfo{year}{2005}.
\newblock \bibinfo{title}{{The Galaxy Evolution Explorer: A Space Ultraviolet
  Survey Mission}}.
\newblock \bibinfo{journal}{\apjl} \bibinfo{volume}{619},
  \bibinfo{pages}{L1--L6}.
\newblock \DOIprefix\doi{10.1086/426387},
  \href{http://arxiv.org/abs/astro-ph/0411302}{\tt arXiv:astro-ph/0411302}.
\bibitem[{{Massaro} et~al.(2015){Massaro}, {Maselli}, {Leto}, {Marchegiani},
  {Perri}, {Giommi} and {Piranomonte}}]{Massaro:2015}
\bibinfo{author}{{Massaro}, E.}, \bibinfo{author}{{Maselli}, A.},
  \bibinfo{author}{{Leto}, C.}, \bibinfo{author}{{Marchegiani}, P.},
  \bibinfo{author}{{Perri}, M.}, \bibinfo{author}{{Giommi}, P.},
  \bibinfo{author}{{Piranomonte}, S.}, \bibinfo{year}{2015}.
\newblock \bibinfo{title}{{The 5th edition of the Roma-BZCAT. A short
  presentation}}.
\newblock \bibinfo{journal}{\apss} \bibinfo{volume}{357}, \bibinfo{pages}{75}.
\newblock \DOIprefix\doi{10.1007/s10509-015-2254-2},
  \href{http://arxiv.org/abs/1502.07755}{\tt arXiv:1502.07755}.
\bibitem[{Paszke et~al.(2017)Paszke, Gross, Chintala, Chanan, Yang, DeVito,
  Lin, Desmaison, Antiga and Lerer}]{pytorch}
\bibinfo{author}{Paszke, A.}, \bibinfo{author}{Gross, S.},
  \bibinfo{author}{Chintala, S.}, \bibinfo{author}{Chanan, G.},
  \bibinfo{author}{Yang, E.}, \bibinfo{author}{DeVito, Z.},
  \bibinfo{author}{Lin, Z.}, \bibinfo{author}{Desmaison, A.},
  \bibinfo{author}{Antiga, L.}, \bibinfo{author}{Lerer, A.},
  \bibinfo{year}{2017}.
\newblock \bibinfo{title}{Automatic differentiation in {PyTorch}}, in:
  \bibinfo{booktitle}{NeurIPS Autodiff Workshop}.
\bibitem[{{Pedregosa} et~al.(2012){Pedregosa}, {Varoquaux}, {Gramfort},
  {Michel}, {Thirion}, {Grisel}, {Blondel}, {M{\"u}ller}, {Nothman}, {Louppe},
  {Prettenhofer}, {Weiss}, {Dubourg}, {Vanderplas}, {Passos}, {Cournapeau},
  {Brucher}, {Perrot} and {Duchesnay}}]{Pedregosa:2021}
\bibinfo{author}{{Pedregosa}, F.}, \bibinfo{author}{{Varoquaux}, G.},
  \bibinfo{author}{{Gramfort}, A.}, \bibinfo{author}{{Michel}, V.},
  \bibinfo{author}{{Thirion}, B.}, \bibinfo{author}{{Grisel}, O.},
  \bibinfo{author}{{Blondel}, M.}, \bibinfo{author}{{M{\"u}ller}, A.},
  \bibinfo{author}{{Nothman}, J.}, \bibinfo{author}{{Louppe}, G.},
  \bibinfo{author}{{Prettenhofer}, P.}, \bibinfo{author}{{Weiss}, R.},
  \bibinfo{author}{{Dubourg}, V.}, \bibinfo{author}{{Vanderplas}, J.},
  \bibinfo{author}{{Passos}, A.}, \bibinfo{author}{{Cournapeau}, D.},
  \bibinfo{author}{{Brucher}, M.}, \bibinfo{author}{{Perrot}, M.},
  \bibinfo{author}{{Duchesnay}, {\'E}.}, \bibinfo{year}{2012}.
\newblock \bibinfo{title}{{Scikit-learn: Machine Learning in Python}}.
\newblock \bibinfo{journal}{arXiv e-prints} ,
  \bibinfo{pages}{arXiv:1201.0490}\href{http://arxiv.org/abs/1201.0490}{\tt
  arXiv:1201.0490}.
\bibitem[{{Petroff} et~al.(2015){Petroff}, {Keane}, {Barr}, {Reynolds},
  {Sarkissian}, {Edwards}, {Stevens}, {Brem}, {Jameson}, {Burke-Spolaor},
  {Johnston}, {Bhat}, {Kudale} and {Bhandari}}]{Petroff:2015}
\bibinfo{author}{{Petroff}, E.}, \bibinfo{author}{{Keane}, E.F.},
  \bibinfo{author}{{Barr}, E.D.}, \bibinfo{author}{{Reynolds}, J.E.},
  \bibinfo{author}{{Sarkissian}, J.}, \bibinfo{author}{{Edwards}, P.G.},
  \bibinfo{author}{{Stevens}, J.}, \bibinfo{author}{{Brem}, C.},
  \bibinfo{author}{{Jameson}, A.}, \bibinfo{author}{{Burke-Spolaor}, S.},
  \bibinfo{author}{{Johnston}, S.}, \bibinfo{author}{{Bhat}, N.D.R.},
  \bibinfo{author}{{Kudale}, P.C.S.}, \bibinfo{author}{{Bhandari}, S.},
  \bibinfo{year}{2015}.
\newblock \bibinfo{title}{{Identifying the source of perytons at the Parkes
  radio telescope}}.
\newblock \bibinfo{journal}{\mnras} \bibinfo{volume}{451},
  \bibinfo{pages}{3933--3940}.
\newblock \DOIprefix\doi{10.1093/mnras/stv1242},
  \href{http://arxiv.org/abs/1504.02165}{\tt arXiv:1504.02165}.
\bibitem[{{Roche}(2011)}]{Roche:2011}
\bibinfo{author}{{Roche}, A.}, \bibinfo{year}{2011}.
\newblock \bibinfo{title}{{EM algorithm and variants: an informal tutorial}}.
\newblock \bibinfo{journal}{arXiv e-prints} ,
  \bibinfo{pages}{arXiv:1105.1476}\href{http://arxiv.org/abs/1105.1476}{\tt
  arXiv:1105.1476}.
\bibitem[{{Skrutskie} et~al.(2006){Skrutskie}, {Cutri}, {Stiening}, {Weinberg},
  {Schneider}, {Carpenter}, {Beichman}, {Capps}, {Chester}, {Elias}, {Huchra},
  {Liebert}, {Lonsdale}, {Monet}, {Price}, {Seitzer}, {Jarrett}, {Kirkpatrick},
  {Gizis}, {Howard}, {Evans}, {Fowler}, {Fullmer}, {Hurt}, {Light}, {Kopan},
  {Marsh}, {McCallon}, {Tam}, {Van Dyk} and {Wheelock}}]{Skrutskie:2006}
\bibinfo{author}{{Skrutskie}, M.F.}, \bibinfo{author}{{Cutri}, R.M.},
  \bibinfo{author}{{Stiening}, R.}, \bibinfo{author}{{Weinberg}, M.D.},
  \bibinfo{author}{{Schneider}, S.}, \bibinfo{author}{{Carpenter}, J.M.},
  \bibinfo{author}{{Beichman}, C.}, \bibinfo{author}{{Capps}, R.},
  \bibinfo{author}{{Chester}, T.}, \bibinfo{author}{{Elias}, J.},
  \bibinfo{author}{{Huchra}, J.}, \bibinfo{author}{{Liebert}, J.},
  \bibinfo{author}{{Lonsdale}, C.}, \bibinfo{author}{{Monet}, D.G.},
  \bibinfo{author}{{Price}, S.}, \bibinfo{author}{{Seitzer}, P.},
  \bibinfo{author}{{Jarrett}, T.}, \bibinfo{author}{{Kirkpatrick}, J.D.},
  \bibinfo{author}{{Gizis}, J.E.}, \bibinfo{author}{{Howard}, E.},
  \bibinfo{author}{{Evans}, T.}, \bibinfo{author}{{Fowler}, J.},
  \bibinfo{author}{{Fullmer}, L.}, \bibinfo{author}{{Hurt}, R.},
  \bibinfo{author}{{Light}, R.}, \bibinfo{author}{{Kopan}, E.L.},
  \bibinfo{author}{{Marsh}, K.A.}, \bibinfo{author}{{McCallon}, H.L.},
  \bibinfo{author}{{Tam}, R.}, \bibinfo{author}{{Van Dyk}, S.},
  \bibinfo{author}{{Wheelock}, S.}, \bibinfo{year}{2006}.
\newblock \bibinfo{title}{{The Two Micron All Sky Survey (2MASS)}}.
\newblock \bibinfo{journal}{\aj} \bibinfo{volume}{131},
  \bibinfo{pages}{1163--1183}.
\newblock \DOIprefix\doi{10.1086/498708}.
\bibitem[{{Sola} and {Sevilla}(1997)}]{Sola:1997}
\bibinfo{author}{{Sola}, J.}, \bibinfo{author}{{Sevilla}, J.},
  \bibinfo{year}{1997}.
\newblock \bibinfo{title}{{Importance of input data normalization for the
  application of neural networks to complex industrial problems}}.
\newblock \bibinfo{journal}{IEEE Transactions on Nuclear Science}
  \bibinfo{volume}{44}, \bibinfo{pages}{1464--1468}.
\newblock \DOIprefix\doi{10.1109/23.589532}.
\bibitem[{{Stern} et~al.(2012){Stern}, {Assef}, {Benford}, {Blain}, {Cutri},
  {Dey}, {Eisenhardt}, {Griffith}, {Jarrett}, {Lake}, {Masci}, {Petty},
  {Stanford}, {Tsai}, {Wright}, {Yan}, {Harrison} and {Madsen}}]{Stern:2012}
\bibinfo{author}{{Stern}, D.}, \bibinfo{author}{{Assef}, R.J.},
  \bibinfo{author}{{Benford}, D.J.}, \bibinfo{author}{{Blain}, A.},
  \bibinfo{author}{{Cutri}, R.}, \bibinfo{author}{{Dey}, A.},
  \bibinfo{author}{{Eisenhardt}, P.}, \bibinfo{author}{{Griffith}, R.L.},
  \bibinfo{author}{{Jarrett}, T.H.}, \bibinfo{author}{{Lake}, S.},
  \bibinfo{author}{{Masci}, F.}, \bibinfo{author}{{Petty}, S.},
  \bibinfo{author}{{Stanford}, S.A.}, \bibinfo{author}{{Tsai}, C.W.},
  \bibinfo{author}{{Wright}, E.L.}, \bibinfo{author}{{Yan}, L.},
  \bibinfo{author}{{Harrison}, F.}, \bibinfo{author}{{Madsen}, K.},
  \bibinfo{year}{2012}.
\newblock \bibinfo{title}{{Mid-infrared Selection of Active Galactic Nuclei
  with the Wide-Field Infrared Survey Explorer. I. Characterizing WISE-selected
  Active Galactic Nuclei in COSMOS}}.
\newblock \bibinfo{journal}{\apj} \bibinfo{volume}{753}, \bibinfo{pages}{30}.
\newblock \DOIprefix\doi{10.1088/0004-637X/753/1/30},
  \href{http://arxiv.org/abs/1205.0811}{\tt arXiv:1205.0811}.
\bibitem[{{Strateva} et~al.(2001){Strateva}, {Ivezi{\'c}}, {Knapp},
  {Narayanan}, {Strauss}, {Gunn}, {Lupton}, {Schlegel}, {Bahcall}, {Brinkmann},
  {Brunner}, {Budav{\'a}ri}, {Csabai}, {Castander}, {Doi}, {Fukugita},
  {Gy{\H{o}}ry}, {Hamabe}, {Hennessy}, {Ichikawa}, {Kunszt}, {Lamb}, {McKay},
  {Okamura}, {Racusin}, {Sekiguchi}, {Schneider}, {Shimasaku} and
  {York}}]{Strateva:2001}
\bibinfo{author}{{Strateva}, I.}, \bibinfo{author}{{Ivezi{\'c}}, {\v{Z}}.},
  \bibinfo{author}{{Knapp}, G.R.}, \bibinfo{author}{{Narayanan}, V.K.},
  \bibinfo{author}{{Strauss}, M.A.}, \bibinfo{author}{{Gunn}, J.E.},
  \bibinfo{author}{{Lupton}, R.H.}, \bibinfo{author}{{Schlegel}, D.},
  \bibinfo{author}{{Bahcall}, N.A.}, \bibinfo{author}{{Brinkmann}, J.},
  \bibinfo{author}{{Brunner}, R.J.}, \bibinfo{author}{{Budav{\'a}ri}, T.},
  \bibinfo{author}{{Csabai}, I.}, \bibinfo{author}{{Castander}, F.J.},
  \bibinfo{author}{{Doi}, M.}, \bibinfo{author}{{Fukugita}, M.},
  \bibinfo{author}{{Gy{\H{o}}ry}, Z.}, \bibinfo{author}{{Hamabe}, M.},
  \bibinfo{author}{{Hennessy}, G.}, \bibinfo{author}{{Ichikawa}, T.},
  \bibinfo{author}{{Kunszt}, P.Z.}, \bibinfo{author}{{Lamb}, D.Q.},
  \bibinfo{author}{{McKay}, T.A.}, \bibinfo{author}{{Okamura}, S.},
  \bibinfo{author}{{Racusin}, J.}, \bibinfo{author}{{Sekiguchi}, M.},
  \bibinfo{author}{{Schneider}, D.P.}, \bibinfo{author}{{Shimasaku}, K.},
  \bibinfo{author}{{York}, D.}, \bibinfo{year}{2001}.
\newblock \bibinfo{title}{{Color Separation of Galaxy Types in the Sloan
  Digital Sky Survey Imaging Data}}.
\newblock \bibinfo{journal}{\aj} \bibinfo{volume}{122},
  \bibinfo{pages}{1861--1874}.
\newblock \DOIprefix\doi{10.1086/323301},
  \href{http://arxiv.org/abs/astro-ph/0107201}{\tt arXiv:astro-ph/0107201}.
\bibitem[{{The Fermi-LAT collaboration}(2019)}]{Fermi:2019_4lac}
\bibinfo{author}{{The Fermi-LAT collaboration}}, \bibinfo{year}{2019}.
\newblock \bibinfo{title}{{The Fourth Catalog of Active Galactic Nuclei
  Detected by the Fermi Large Area Telescope}}.
\newblock \bibinfo{journal}{arXiv e-prints} ,
  \bibinfo{pages}{arXiv:1905.10771}\href{http://arxiv.org/abs/1905.10771}{\tt
  arXiv:1905.10771}.
\bibitem[{{Urry} and {Padovani}(1995)}]{Urry:1995}
\bibinfo{author}{{Urry}, C.M.}, \bibinfo{author}{{Padovani}, P.},
  \bibinfo{year}{1995}.
\newblock \bibinfo{title}{{Unified Schemes for Radio-Loud Active Galactic
  Nuclei}}.
\newblock \bibinfo{journal}{\pasp} \bibinfo{volume}{107}, \bibinfo{pages}{803}.
\newblock \DOIprefix\doi{10.1086/133630},
  \href{http://arxiv.org/abs/astro-ph/9506063}{\tt arXiv:astro-ph/9506063}.
\bibitem[{{van der Walt} et~al.(2011){van der Walt}, {Colbert} and
  {Varoquaux}}]{numpy}
\bibinfo{author}{{van der Walt}, S.}, \bibinfo{author}{{Colbert}, S.C.},
  \bibinfo{author}{{Varoquaux}, G.}, \bibinfo{year}{2011}.
\newblock \bibinfo{title}{{The NumPy Array: A Structure for Efficient Numerical
  Computation}}.
\newblock \bibinfo{journal}{Computing in Science and Engineering}
  \bibinfo{volume}{13}, \bibinfo{pages}{22--30}.
\newblock \DOIprefix\doi{10.1109/MCSE.2011.37},
  \href{http://arxiv.org/abs/1102.1523}{\tt arXiv:1102.1523}.
\bibitem[{{Wiyatno} et~al.(2019){Wiyatno}, {Xu}, {Dia} and {de
  Berker}}]{Wiyatno:2019}
\bibinfo{author}{{Wiyatno}, R.R.}, \bibinfo{author}{{Xu}, A.},
  \bibinfo{author}{{Dia}, O.}, \bibinfo{author}{{de Berker}, A.},
  \bibinfo{year}{2019}.
\newblock \bibinfo{title}{{Adversarial Examples in Modern Machine Learning: A
  Review}}.
\newblock \bibinfo{journal}{arXiv e-prints} ,
  \bibinfo{pages}{arXiv:1911.05268}\href{http://arxiv.org/abs/1911.05268}{\tt
  arXiv:1911.05268}.
\bibitem[{{Wright} et~al.(2010){Wright}, {Eisenhardt}, {Mainzer}, {Ressler},
  {Cutri}, {Jarrett}, {Kirkpatrick}, {Padgett}, {McMillan}, {Skrutskie},
  {Stanford}, {Cohen}, {Walker}, {Mather}, {Leisawitz}, {Gautier}, {McLean},
  {Benford}, {Lonsdale}, {Blain}, {Mendez}, {Irace}, {Duval}, {Liu}, {Royer},
  {Heinrichsen}, {Howard}, {Shannon}, {Kendall}, {Walsh}, {Larsen}, {Cardon},
  {Schick}, {Schwalm}, {Abid}, {Fabinsky}, {Naes} and {Tsai}}]{Wright:2010}
\bibinfo{author}{{Wright}, E.L.}, \bibinfo{author}{{Eisenhardt}, P.R.M.},
  \bibinfo{author}{{Mainzer}, A.K.}, \bibinfo{author}{{Ressler}, M.E.},
  \bibinfo{author}{{Cutri}, R.M.}, \bibinfo{author}{{Jarrett}, T.},
  \bibinfo{author}{{Kirkpatrick}, J.D.}, \bibinfo{author}{{Padgett}, D.},
  \bibinfo{author}{{McMillan}, R.S.}, \bibinfo{author}{{Skrutskie}, M.},
  \bibinfo{author}{{Stanford}, S.A.}, \bibinfo{author}{{Cohen}, M.},
  \bibinfo{author}{{Walker}, R.G.}, \bibinfo{author}{{Mather}, J.C.},
  \bibinfo{author}{{Leisawitz}, D.}, \bibinfo{author}{{Gautier}, III, T.N.},
  \bibinfo{author}{{McLean}, I.}, \bibinfo{author}{{Benford}, D.},
  \bibinfo{author}{{Lonsdale}, C.J.}, \bibinfo{author}{{Blain}, A.},
  \bibinfo{author}{{Mendez}, B.}, \bibinfo{author}{{Irace}, W.R.},
  \bibinfo{author}{{Duval}, V.}, \bibinfo{author}{{Liu}, F.},
  \bibinfo{author}{{Royer}, D.}, \bibinfo{author}{{Heinrichsen}, I.},
  \bibinfo{author}{{Howard}, J.}, \bibinfo{author}{{Shannon}, M.},
  \bibinfo{author}{{Kendall}, M.}, \bibinfo{author}{{Walsh}, A.L.},
  \bibinfo{author}{{Larsen}, M.}, \bibinfo{author}{{Cardon}, J.G.},
  \bibinfo{author}{{Schick}, S.}, \bibinfo{author}{{Schwalm}, M.},
  \bibinfo{author}{{Abid}, M.}, \bibinfo{author}{{Fabinsky}, B.},
  \bibinfo{author}{{Naes}, L.}, \bibinfo{author}{{Tsai}, C.W.},
  \bibinfo{year}{2010}.
\newblock \bibinfo{title}{{The Wide-field Infrared Survey Explorer (WISE):
  Mission Description and Initial On-orbit Performance}}.
\newblock \bibinfo{journal}{\aj} \bibinfo{volume}{140},
  \bibinfo{pages}{1868--1881}.
\newblock \DOIprefix\doi{10.1088/0004-6256/140/6/1868},
  \href{http://arxiv.org/abs/1008.0031}{\tt arXiv:1008.0031}.
\bibitem[{Wu(1983)}]{Wu:1983}
\bibinfo{author}{Wu, C.F.J.}, \bibinfo{year}{1983}.
\newblock \bibinfo{title}{On the convergence properties of the em algorithm}.
\newblock \bibinfo{journal}{The Annals of Statistics} \bibinfo{volume}{11},
  \bibinfo{pages}{95--103}.
\newblock \URLprefix \url{http://www.jstor.org/stable/2240463}.
\bibitem[{{York} et~al.(2000){York}, {Adelman}, {Anderson}, {Anderson},
  {Annis}, {Bahcall}, {Bakken}, {Barkhouser}, {Bastian}, {Berman}, {Boroski},
  {Bracker}, {Briegel}, {Briggs}, {Brinkmann}, {Brunner}, {Burles}, {Carey},
  {Carr}, {Castander}, {Chen}, {Colestock}, {Connolly}, {Crocker}, {Csabai},
  {Czarapata}, {Davis}, {Doi}, {Dombeck}, {Eisenstein}, {Ellman}, {Elms},
  {Evans}, {Fan}, {Federwitz}, {Fiscelli}, {Friedman}, {Frieman}, {Fukugita},
  {Gillespie}, {Gunn}, {Gurbani}, {de Haas}, {Haldeman}, {Harris}, {Hayes},
  {Heckman}, {Hennessy}, {Hindsley}, {Holm}, {Holmgren}, {Huang}, {Hull},
  {Husby}, {Ichikawa}, {Ichikawa}, {Ivezi{\'c}}, {Kent}, {Kim}, {Kinney},
  {Klaene}, {Kleinman}, {Kleinman}, {Knapp}, {Korienek}, {Kron}, {Kunszt},
  {Lamb}, {Lee}, {Leger}, {Limmongkol}, {Lindenmeyer}, {Long}, {Loomis},
  {Loveday}, {Lucinio}, {Lupton}, {MacKinnon}, {Mannery}, {Mantsch}, {Margon},
  {McGehee}, {McKay}, {Meiksin}, {Merelli}, {Monet}, {Munn}, {Narayanan},
  {Nash}, {Neilsen}, {Neswold}, {Newberg}, {Nichol}, {Nicinski}, {Nonino},
  {Okada}, {Okamura}, {Ostriker}, {Owen}, {Pauls}, {Peoples}, {Peterson},
  {Petravick}, {Pier}, {Pope}, {Pordes}, {Prosapio}, {Rechenmacher}, {Quinn},
  {Richards}, {Richmond}, {Rivetta}, {Rockosi}, {Ruthmansdorfer}, {Sand ford},
  {Schlegel}, {Schneider}, {Sekiguchi}, {Sergey}, {Shimasaku}, {Siegmund},
  {Smee}, {Smith}, {Snedden}, {Stone}, {Stoughton}, {Strauss}, {Stubbs},
  {SubbaRao}, {Szalay}, {Szapudi}, {Szokoly}, {Thakar}, {Tremonti}, {Tucker},
  {Uomoto}, {Vanden Berk}, {Vogeley}, {Waddell}, {Wang}, {Watanabe},
  {Weinberg}, {Yanny}, {Yasuda} and {SDSS Collaboration}}]{York:2000}
\bibinfo{author}{{York}, D.G.}, \bibinfo{author}{{Adelman}, J.},
  \bibinfo{author}{{Anderson}, John~E., J.}, \bibinfo{author}{{Anderson},
  S.F.}, \bibinfo{author}{{Annis}, J.}, \bibinfo{author}{{Bahcall}, N.A.},
  \bibinfo{author}{{Bakken}, J.A.}, \bibinfo{author}{{Barkhouser}, R.},
  \bibinfo{author}{{Bastian}, S.}, \bibinfo{author}{{Berman}, E.},
  \bibinfo{author}{{Boroski}, W.N.}, \bibinfo{author}{{Bracker}, S.},
  \bibinfo{author}{{Briegel}, C.}, \bibinfo{author}{{Briggs}, J.W.},
  \bibinfo{author}{{Brinkmann}, J.}, \bibinfo{author}{{Brunner}, R.},
  \bibinfo{author}{{Burles}, S.}, \bibinfo{author}{{Carey}, L.},
  \bibinfo{author}{{Carr}, M.A.}, \bibinfo{author}{{Castander}, F.J.},
  \bibinfo{author}{{Chen}, B.}, \bibinfo{author}{{Colestock}, P.L.},
  \bibinfo{author}{{Connolly}, A.J.}, \bibinfo{author}{{Crocker}, J.H.},
  \bibinfo{author}{{Csabai}, I.}, \bibinfo{author}{{Czarapata}, P.C.},
  \bibinfo{author}{{Davis}, J.E.}, \bibinfo{author}{{Doi}, M.},
  \bibinfo{author}{{Dombeck}, T.}, \bibinfo{author}{{Eisenstein}, D.},
  \bibinfo{author}{{Ellman}, N.}, \bibinfo{author}{{Elms}, B.R.},
  \bibinfo{author}{{Evans}, M.L.}, \bibinfo{author}{{Fan}, X.},
  \bibinfo{author}{{Federwitz}, G.R.}, \bibinfo{author}{{Fiscelli}, L.},
  \bibinfo{author}{{Friedman}, S.}, \bibinfo{author}{{Frieman}, J.A.},
  \bibinfo{author}{{Fukugita}, M.}, \bibinfo{author}{{Gillespie}, B.},
  \bibinfo{author}{{Gunn}, J.E.}, \bibinfo{author}{{Gurbani}, V.K.},
  \bibinfo{author}{{de Haas}, E.}, \bibinfo{author}{{Haldeman}, M.},
  \bibinfo{author}{{Harris}, F.H.}, \bibinfo{author}{{Hayes}, J.},
  \bibinfo{author}{{Heckman}, T.M.}, \bibinfo{author}{{Hennessy}, G.S.},
  \bibinfo{author}{{Hindsley}, R.B.}, \bibinfo{author}{{Holm}, S.},
  \bibinfo{author}{{Holmgren}, D.J.}, \bibinfo{author}{{Huang}, C.h.},
  \bibinfo{author}{{Hull}, C.}, \bibinfo{author}{{Husby}, D.},
  \bibinfo{author}{{Ichikawa}, S.I.}, \bibinfo{author}{{Ichikawa}, T.},
  \bibinfo{author}{{Ivezi{\'c}}, {\v{Z}}.}, \bibinfo{author}{{Kent}, S.},
  \bibinfo{author}{{Kim}, R.S.J.}, \bibinfo{author}{{Kinney}, E.},
  \bibinfo{author}{{Klaene}, M.}, \bibinfo{author}{{Kleinman}, A.N.},
  \bibinfo{author}{{Kleinman}, S.}, \bibinfo{author}{{Knapp}, G.R.},
  \bibinfo{author}{{Korienek}, J.}, \bibinfo{author}{{Kron}, R.G.},
  \bibinfo{author}{{Kunszt}, P.Z.}, \bibinfo{author}{{Lamb}, D.Q.},
  \bibinfo{author}{{Lee}, B.}, \bibinfo{author}{{Leger}, R.F.},
  \bibinfo{author}{{Limmongkol}, S.}, \bibinfo{author}{{Lindenmeyer}, C.},
  \bibinfo{author}{{Long}, D.C.}, \bibinfo{author}{{Loomis}, C.},
  \bibinfo{author}{{Loveday}, J.}, \bibinfo{author}{{Lucinio}, R.},
  \bibinfo{author}{{Lupton}, R.H.}, \bibinfo{author}{{MacKinnon}, B.},
  \bibinfo{author}{{Mannery}, E.J.}, \bibinfo{author}{{Mantsch}, P.M.},
  \bibinfo{author}{{Margon}, B.}, \bibinfo{author}{{McGehee}, P.},
  \bibinfo{author}{{McKay}, T.A.}, \bibinfo{author}{{Meiksin}, A.},
  \bibinfo{author}{{Merelli}, A.}, \bibinfo{author}{{Monet}, D.G.},
  \bibinfo{author}{{Munn}, J.A.}, \bibinfo{author}{{Narayanan}, V.K.},
  \bibinfo{author}{{Nash}, T.}, \bibinfo{author}{{Neilsen}, E.},
  \bibinfo{author}{{Neswold}, R.}, \bibinfo{author}{{Newberg}, H.J.},
  \bibinfo{author}{{Nichol}, R.C.}, \bibinfo{author}{{Nicinski}, T.},
  \bibinfo{author}{{Nonino}, M.}, \bibinfo{author}{{Okada}, N.},
  \bibinfo{author}{{Okamura}, S.}, \bibinfo{author}{{Ostriker}, J.P.},
  \bibinfo{author}{{Owen}, R.}, \bibinfo{author}{{Pauls}, A.G.},
  \bibinfo{author}{{Peoples}, J.}, \bibinfo{author}{{Peterson}, R.L.},
  \bibinfo{author}{{Petravick}, D.}, \bibinfo{author}{{Pier}, J.R.},
  \bibinfo{author}{{Pope}, A.}, \bibinfo{author}{{Pordes}, R.},
  \bibinfo{author}{{Prosapio}, A.}, \bibinfo{author}{{Rechenmacher}, R.},
  \bibinfo{author}{{Quinn}, T.R.}, \bibinfo{author}{{Richards}, G.T.},
  \bibinfo{author}{{Richmond}, M.W.}, \bibinfo{author}{{Rivetta}, C.H.},
  \bibinfo{author}{{Rockosi}, C.M.}, \bibinfo{author}{{Ruthmansdorfer}, K.},
  \bibinfo{author}{{Sand ford}, D.}, \bibinfo{author}{{Schlegel}, D.J.},
  \bibinfo{author}{{Schneider}, D.P.}, \bibinfo{author}{{Sekiguchi}, M.},
  \bibinfo{author}{{Sergey}, G.}, \bibinfo{author}{{Shimasaku}, K.},
  \bibinfo{author}{{Siegmund}, W.A.}, \bibinfo{author}{{Smee}, S.},
  \bibinfo{author}{{Smith}, J.A.}, \bibinfo{author}{{Snedden}, S.},
  \bibinfo{author}{{Stone}, R.}, \bibinfo{author}{{Stoughton}, C.},
  \bibinfo{author}{{Strauss}, M.A.}, \bibinfo{author}{{Stubbs}, C.},
  \bibinfo{author}{{SubbaRao}, M.}, \bibinfo{author}{{Szalay}, A.S.},
  \bibinfo{author}{{Szapudi}, I.}, \bibinfo{author}{{Szokoly}, G.P.},
  \bibinfo{author}{{Thakar}, A.R.}, \bibinfo{author}{{Tremonti}, C.},
  \bibinfo{author}{{Tucker}, D.L.}, \bibinfo{author}{{Uomoto}, A.},
  \bibinfo{author}{{Vanden Berk}, D.}, \bibinfo{author}{{Vogeley}, M.S.},
  \bibinfo{author}{{Waddell}, P.}, \bibinfo{author}{{Wang}, S.i.},
  \bibinfo{author}{{Watanabe}, M.}, \bibinfo{author}{{Weinberg}, D.H.},
  \bibinfo{author}{{Yanny}, B.}, \bibinfo{author}{{Yasuda}, N.},
  \bibinfo{author}{{SDSS Collaboration}}, \bibinfo{year}{2000}.
\newblock \bibinfo{title}{{The Sloan Digital Sky Survey: Technical Summary}}.
\newblock \bibinfo{journal}{\aj} \bibinfo{volume}{120},
  \bibinfo{pages}{1579--1587}.
\newblock \DOIprefix\doi{10.1086/301513},
  \href{http://arxiv.org/abs/astro-ph/0006396}{\tt arXiv:astro-ph/0006396}.

\end{thebibliography}

\end{document}